\documentclass[onecolumn,12pt]{article} % For LaTeX2e

\def\bbE{\mathbb{E}}

\def\bbP{\mathbb{P}}

\def\bbR{\mathbb{R}}

\def\cD{\mathcal{D}}
\def\cE{\mathcal{E}}
\def\cF{\mathcal{F}}
\def\cG{\mathcal{G}}

\def\cK{\mathcal{K}}

\def\cR{\mathcal{R}}

\def\cU{\mathcal{U}}

\def\cX{\mathcal{X}}
\def\cY{\mathcal{Y}}

\def\Ex{{\bbE}}

\def\exp{{\sf exp}}

\def\ones{\mathbf{1}}

\usepackage[hmargin={1.0in, 1.0in}, vmargin={1.0in, 1.0in}]{geometry}
\usepackage[onehalfspacing]{setspace}
\usepackage{booktabs} % for professional tables
\usepackage{hyperref}
\usepackage{url}
\usepackage{subcaption}
\usepackage{microtype}
\usepackage{graphicx}
\usepackage{booktabs} % for professional tables
\usepackage{amsmath}
\usepackage{amssymb}
\usepackage{mathtools}
\usepackage{amsthm}
\usepackage{xcolor}
\usepackage{float}
\usepackage{algorithm}
\usepackage{algpseudocode}
\graphicspath{ {Figures/} }
\usepackage{bm}
\usepackage{natbib}

\theoremstyle{plain}
\newtheorem{theorem}{Theorem}[section]

\newtheorem{lemma}[theorem]{Lemma}
\newtheorem{procedure}[theorem]{Procedure}

\theoremstyle{definition}
\newtheorem{definition}[theorem]{Definition}

\theoremstyle{remark}

\title{Discrimination-Free Insurance Pricing with Privatized Sensitive Attributes}

\author{Tianhe Zhang \thanks{Correspondence to: Tianhe Zhang: tzhang349@wisc.edu} \ 
\thanks{Department of Risk and Insurance, University of Wisconsin-Madison}
\and
Suhan Liu 
\thanks{Department of Statistics and Operations Research, University of North Carolina-Chapel Hill}
\and
Peng Shi 
\footnotemark[2]
}

\begin{document}

\maketitle

% \vspace{-15pt}
\begin{abstract}
% \vspace{-5pt}
Fairness has become an important concern in insurance pricing as insurers increasingly rely on machine learning models to predict expected losses. At the same time, regulatory and privacy constraints often restrict insurers’ ability to access or use sensitive attributes such as gender or race. Recent actuarial research addresses fairness in this context through the concept of the discrimination-free premium, which removes both the direct and indirect effects of sensitive attributes while preserving actuarial consistency. However, implementing this approach typically requires access to the sensitive attributes themselves, which may not be available in practice.

This paper studies the estimation of discrimination-free insurance premiums when sensitive attributes are observed only in privatized or noise-perturbed form. We consider a multi-party data setting in which insurers observe non-sensitive attributes and outcomes, while a trusted third party holds privatized sensitive attributes generated through a privacy mechanism. Within this framework, we develop statistical methods for estimating discrimination-free premiums using only the privatized attributes. We study two settings of practical relevance: when the privacy mechanism is known and when its noise level is unknown. For both cases, we establish theoretical guarantees for the proposed estimators. Numerical experiments and empirical applications demonstrate that the proposed approach enables fair insurance pricing while respecting privacy and regulatory constraints.
\end{abstract}

{\it Keywords: Insurance pricing; Discrimination-free premium; Algorithmic fairness; Privacy-preserving learning; Noisy sensitive attributes}

\newpage
% \vspace{-10pt}

\section{Introduction}
% \vspace{-5pt}
\label{sec:1}

Fairness has become a growing concern in insurance pricing. Insurance premiums determine how risks and costs are allocated across individuals, placing fairness at the center of regulatory oversight and public trust in insurance markets. At the same time, insurance pricing increasingly relies on statistical and machine learning algorithms that analyze large volumes of data to predict expected losses \citep{blier2020machine}. While these methods can improve predictive accuracy, they may also produce outcomes that disadvantage individuals or groups defined by protected characteristics such as gender, race, or ethnicity. As algorithmic decision-making becomes more prevalent, ensuring fairness in insurance pricing has become a key challenge for both regulators and insurers \citep{charpentier2024insurance}.

This paper studies fair insurance pricing in settings where sensitive attributes are not directly observable. In practice, regulators impose restrictions on the use of sensitive attributes in insurance pricing. For example, Directive 2004/113/EC (“Gender Directive”) issued by the Council of the European Union prohibits insurers from using gender as a rating factor when pricing insurance products. More recently, the state of Colorado in the United States enacted Senate Bill (SB) 21-169, which prohibits unfair discrimination in insurance practices based on characteristics including race, national origin, religion, sex, sexual orientation, disability, and gender identity. Regulators have also raised concerns that complex predictive models may unintentionally encode discriminatory patterns through correlated variables or proxy relationships \citep{van2025ai}. As a result, insurers often cannot directly access or use protected attributes when developing pricing models. The objective of this paper is to develop statistical methods that enable fair insurance pricing under such constraints while avoiding both direct and indirect discrimination.

Fairness in insurance differs in important ways from fairness notions commonly studied in the machine learning literature. In the broader algorithmic fairness literature, several formal definitions have been proposed, including demographic parity, equalized odds, and predictive parity \citep{10.1145/2090236.2090255, 10.5555/3157382.3157469, Zafar2017FairnessConstraints, 10.5555/3294996.3295162}. These definitions focus on ensuring equitable outcomes across demographic groups in algorithmic decision-making systems. By contrast, insurance pricing traditionally follows the principle of actuarial fairness, under which premiums should reflect the expected cost associated with the insured risk \citep{Miller2009disparate, FreesHuang2023}. Accurate risk classification is therefore fundamental to maintaining financial sustainability and preventing unintended cross-subsidization across policyholders. Consequently, fairness considerations in insurance must be reconciled with the actuarial objective of accurately estimating expected losses.

Recent work in actuarial science has addresses this tension through the concept of a discrimination-free premium. \cite{Lindholm_Richman_Tsanakas_Wuthrich_2022_DFP} propose a causal framework for defining premiums that remove both the direct and indirect effects of sensitive attributes while preserving actuarial consistency. Under this framework, pricing rules are constructed so that sensitive attributes do not influence premiums at the individual level, either directly or through correlated proxy variables. In theory, this approach provides a principled way to incorporate fairness considerations into actuarial pricing models. In practice, however, implementing discrimination-free pricing is challenging under current regulatory and privacy constraints. Sensitive attributes are often unavailable during model development because access to such information is restricted or because only privatized or noise-perturbed versions of the attributes can be used. Yet implementing the discrimination-free premium typically requires knowledge of the sensitive attributes themselves.

To overcome this practical limitation, we develop a statistical framework for estimating discrimination-free premiums when sensitive attributes are observed only through a privacy mechanism that introduces noise. Our work connects the actuarial literature on fair pricing with statistical methods for learning under corrupted or privatized data \citep{Warner_1965, Maaten2013LearningWM, li2016learning}. To capture the institutional realities of modern insurance data ecosystems, we consider a multi-party learning framework in which insurers observe non-sensitive attributes and outcomes but do not observe the true sensitive attributes. Instead, privatized sensitive attributes are maintained by a trusted third party that facilitates model training. Such arrangements are increasingly common as insurers collaborate with external analytics vendors and data intermediaries to implement complex machine learning models while respecting privacy constraints. Within this framework, discrimination-free premiums can be estimated even when sensitive attributes are available only in noisy or privatized form. We examine two settings of practical interest. In the first, the trusted third party has full knowledge of the privacy mechanism generating the privatized attributes, including the noise rate. In the second, the privacy mechanism is known but the noise rate itself is unknown.

The proposed method offers several advantages. First, it enables the construction of discrimination-free pricing models without requiring insurers to access sensitive attributes directly, thereby strengthening privacy protection. Second, the framework is flexible with respect to how sensitive attributes are collected and privatized, relying only on the availability of their privatized versions. Third, the proposed estimation procedures are straightforward to implement and come with formal statistical guarantees. The contributions of this paper are threefold. First, we develop a statistical method for estimating discrimination-free premiums when sensitive attributes are observed only through a privacy mechanism. Second, we establish theoretical guarantees for the proposed estimators under both known and unknown noise rates. Third, we demonstrate the empirical performance of the proposed approach and analyze the impact of noise-rate misspecification on pricing accuracy.

The remainder of the paper is organized as follows. Section \ref{sec:2} reviews the related literature on fairness in machine learning and insurance pricing. Section \ref{sec:3} introduces the discrimination-free pricing rule and the trusted third-party framework underlying our theoretical development. Section \ref{sec:4} presents the proposed methodology for estimating discrimination-free premiums under both true and privatized sensitive attributes, together with statistical guarantees when the noise rate is known or unknown. Section \ref{sec:5} reports numerical experiments illustrating the main theoretical results. Section \ref{sec:6} applies the proposed method to real-world datasets in both regression and classification settings. Section \ref{sec:7} concludes. Additional technical details and empirical results are provided in the Appendix.

% \vspace{-10pt}
\section{Relation to Existing Literature}
% \vspace{-5pt}
\label{sec:2}

Our work lies at the intersection of three strands of research: algorithmic fairness in machine learning, fairness in insurance pricing, and statistical learning under privacy or noise constraints.

\subsection{Algorithmic Fairness in Machine Learning}

The machine learning literature has developed several formal frameworks for evaluating fairness in algorithmic decision-making. One prominent concept is individual fairness, which requires that similar individuals receive similar treatment \citep{10.1145/2090236.2090255}. Another widely studied class consists of group fairness criteria, such as demographic parity and equalized odds, which aim to equalize statistical outcomes across demographic groups \citep{10.5555/3157382.3157469, Zafar2017FairnessConstraints}. More recent work introduces causal notions of fairness, including counterfactual fairness, which evaluates whether a decision would change if an individual's sensitive attribute were different while holding other factors constant \citep{10.5555/3294996.3295162}.

These fairness definitions have motivated a wide range of algorithmic mitigation strategies, including pre-processing, in-processing, and post-processing approaches \citep{10.5555/3294996.3295155, JMLR:v21:19-966, 10.5555/3327144.3327203, pmlr-v80-agarwal18a, pmlr-v97-agarwal19d, pmlr-v65-woodworth17a}. However, their direct application to insurance pricing is complicated by the phenomenon of proxy discrimination. Machine learning models often identify predictive variables that strongly correlate with protected attributes, effectively reconstructing the forbidden attribute through neutral features. As a result, simply removing sensitive attributes from the model—often referred to as fairness through unawareness—may not eliminate discriminatory outcomes.

\subsection{Fairness in Insurance Pricing}

In insurance, the traditional benchmark for pricing is actuarial fairness, under which premiums equal the expected value of losses \citep{Miller2009disparate}. As insurers increasingly adopt high-dimensional machine learning models—such as gradient boosting and neural networks—to improve predictive accuracy \citep{blier2020machine, ShiZhangShi2024}, concerns about algorithmic discrimination have intensified.

Recent actuarial research distinguishes between direct discrimination, where sensitive attributes explicitly influence premiums, and indirect discrimination, where non-sensitive variables serve as proxies for protected characteristics \citep{FreesHuang2023, XinHuang2024anti, cote2025fair}. To reconcile predictive accuracy with fairness considerations, \citet{Lindholm_Richman_Tsanakas_Wuthrich_2022_DFP} introduce the concept of the discrimination-free premium, which removes both the direct and indirect causal effects of sensitive attributes while preserving actuarial consistency. Subsequent work has extended this framework in several directions. For example, \citet{AraizaIturriaHardyMarriott2024} employ path-specific counterfactual fairness to decompose the influence of sensitive attributes along different causal pathways, while \citet{LindholmRichmanTsanakasWuthrich2024} propose probabilistic measures to quantify unfair differentiation across risk groups. However, these approaches typically assume that the insurer has access to a dataset containing the sensitive attributes required to perform the fairness adjustments. In practice, this assumption is increasingly unrealistic due to regulatory restrictions and privacy constraints.

\subsection{Learning Under Privacy and Noise Constraints}

Our work also relates to the literature on statistical learning with corrupted or privatized features. Early foundations were established by the randomized response technique \citep{Warner_1965}, which introduced randomized reporting to protect respondent privacy. More recently, this idea has been formalized through local differential privacy, which studies statistical inference when sensitive attributes are intentionally perturbed to preserve privacy \citep{Duchi2013LDP, 10.5555/2946645.2946662}.

A growing body of research examines fairness under noisy or partially observed sensitive attributes \citep{10.1145/3442188.3445887, pmlr-v139-celis21a, Ghosh_2023}. These studies typically focus on classification problems and enforce fairness through optimization constraints such as equalized error rates across groups. In contrast, our work focuses on regression-based loss prediction and the actuarial objective of estimating expected losses. By integrating privacy-aware statistical learning with the discrimination-free premium framework, we provide a practical method for implementing fair insurance pricing when sensitive attributes are only available in privatized form.

\section{Pricing Framework and Problem Setup}
\label{sec:3}

This section introduces the pricing framework and data environment considered in the paper and formally states the statistical inference problem addressed in the remainder of the paper.

\subsection{Pricing Rules}

Let $\bbP_{X, D, Y}$ be an unknown distribution, where $n$ i.i.d. samples $(x_i,d_i,y_i) \sim \bbP_{X,D,Y}$, $i = 1, \ldots, n$ are drawn from. $X \in \cX$ denotes non-sensitive attributes, $D \in \cD$ denotes true sensitive attributes that we assume to be discrete (e.g., gender or ethnicity), and $Y \in \cY$ denotes the outcome of interest, which may be either discrete or continuous, such as claim count or loss severity.

Insurance pricing models aim to estimate the expected loss associated with each policyholder. In the fairness literature, three pricing rules are commonly discussed.
\begin{definition} {\bf(Best-estimated Price)} The best-estimated price for $Y$ given $(X,D)$ is defined as: 
\label{def:bep}
\[
\mu(X, D):= \mathbb{E}[Y | X,D].
\]
This price is actuarially optimal in the sense that it minimizes prediction error when both sensitive and non-sensitive attributes are available. However, it explicitly depends on the sensitive attribute $D$ and therefore leads to direct discrimination.
\end{definition} 
 
\begin{definition} {\bf (Unawareness Price)} The unawareness price for $Y$ given $X$ is defined as:
\label{def:up}
\[
\mu(X):= \mathbb{E}[Y | X].
\]
This pricing rule excludes sensitive attributes from the model. Nevertheless, indirect discrimination may still occur because the sensitive attribute may be statistically correlated with the observable features $X$. By the law of total expectation,
\[
\mu(X) = \int_{d} \mu(X,d)\, d \, \mathbb{P}(d | X).
\]
which shows that the prediction implicitly depends on the conditional distribution of the sensitive attribute.
\end{definition}

\begin{definition} {\bf (Discrimination-free Price)} To address both direct and indirect discrimination, \citet{Lindholm_Richman_Tsanakas_Wuthrich_2022_DFP} propose the discrimination-free price:
\label{def:dfp}
\[
h^*(X) := \int_{d} \mu(X,d) \, d \, \mathbb{P}^*(d),
\]
where $\mathbb{P}^*(d)$ is a reference distribution over the sensitive attribute $D$, typically taken as the marginal distribution $\mathbb{P}(D)$ or another socially justified prior. This definition ensures that sensitive attributes do not influence pricing at the individual level, either directly or indirectly. When the conditional independence condition $Y \perp D \mid X$ holds, the discrimination-free price coincides with the unawareness price.
\end{definition}

\subsection{Data Setting with Privatized Sensitive Attributes}
Estimating the discrimination-free price requires knowledge of $\mu(X,d)$, which depends on the sensitive attribute $D$. In practice, however, access to sensitive attributes is often restricted due to regulatory or privacy constraints. 

\begin{figure}[H]
    \centering
    \includegraphics[width = 1\linewidth]{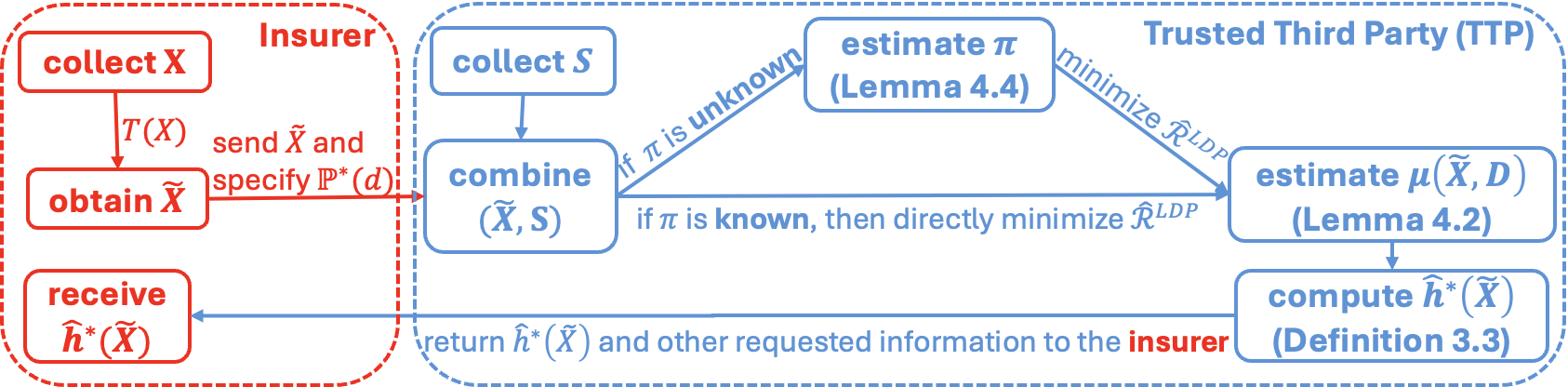}
    \caption{Insurer-TTP Interaction Diagram.}
    % \vspace{-12pt}
    \label{fig:insurer-TTP-general}
\end{figure}

Modern insurance data ecosystems frequently involve multiple parties that hold different components of the data. Insurers typically observe non-sensitive attributes and outcomes, while sensitive attributes may be collected by a separate entity such as a data intermediary or analytics vendor. In such environments, sensitive attributes may only be available in privatized or noise-perturbed form. To capture this environment, we consider a multi-party data setting involving a trusted third party (TTP). The insurer observes $(X,Y)$ but does not have direct access to the true sensitive attribute $D$. Instead, the TTP holds a privatized version of the sensitive attribute, denoted by $S$, which is generated through a privacy mechanism that introduces noise. The interaction between the insurer and the TTP is illustrated in Figure \ref{fig:insurer-TTP-general}. The insurer provides transformed non-sensitive attributes, denoted by $\tilde{X}$, together with the outcome variable to the TTP, which combines this information with the privatized sensitive attributes to train the pricing model.

% \vspace{-8pt}
% \vspace{-8pt}

Such multi-party arrangements arise naturally in practice. Regulatory constraints may limit insurers' ability to directly access sensitive attributes, while insurers increasingly rely on external vendors to implement complex machine learning models. In these environments, sensitive attributes may only be available through privacy-preserving mechanisms or through intermediaries responsible for managing sensitive data. Noise in sensitive attributes may arise from several sources. For example, privacy-preserving data collection mechanisms may deliberately perturb sensitive information to protect confidentiality. Measurement errors may also occur when sensitive attributes are self-reported or imputed from external data sources. In addition, sensitive data may be intentionally privatized when transmitted between organizations to comply with data protection regulations.

The framework described above also encompasses two common industry scenarios as special cases. In the first scenario, the insurer receives privatized sensitive attributes from a third party and trains the pricing algorithm internally. In the second scenario, the insurer collects both sensitive and non-sensitive data but delegates model training to an external vendor.

\subsection{Inference Problem}

Under the data setting described above, the true sensitive attribute $D$ is not observable to the insurer. Instead, only the privatized attribute $S$, generated through the privacy mechanism, is available. The central question studied in this paper is therefore: How can the discrimination-free premium $h^*(X)$ be estimated when the sensitive attribute $D$ is unobserved and only a privatized version $S$ is available? 

The discrimination-free price depends on two components: the conditional mean outcome $\mu(X, D)$ and a weighting distribution over sensitive attributes $\mathbb{P}^*(d)$. A natural and practical choice for $\mathbb{P}^*(d)$ is the empirical distribution of the sensitive attribute, although other choices may be used to satisfy desirable statistical properties. The primary challenge therefore lies in estimating $\mu(X, D)$ when the sensitive attribute is not directly observed. In the next section, we develop statistical methods for estimating the discrimination-free price under this setting and establish theoretical guarantees for the resulting estimators. Figure \ref{fig:insurer-TTP-general} not only depicts the information flow between the insurer and the trusted third party, but also organizes the main definitions and theorems that underpin our framework. It therefore serves as a roadmap connecting the major components of the TTP setting to the corresponding theoretical results and clarifies how these results build on one another.

\section{Discrimination-free Pricing for Insurance}
% \vspace{-5pt}
\label{sec:4}

% The ultimate goal of our framework is to compute the discrimination-free premium function $h^*(X)$, which depends on two key components: the conditional mean outcome $\mu(X, D)$ and a weighting distribution over sensitive attributes, $\mathbb{P}^*(d)$. A natural and practical choice for $\mathbb{P}^*(d)$ is its empirical distribution. More generally, it can be treated as a tuning parameter selected to satisfy desirable statistical properties, such as unbiasedness. Accordingly, the primary challenge lies in accurately estimating $\mu(X, D)$, particularly when the sensitive attributes are not fully observed. 

In this sections, we study the estimation discrimination-free premium under two settings: when the sensitive attributes are fully observed, and when only noised or privatized versions are available. For the latter, we further distinguish between cases where the noise mechanism is known and where it is unknown.

% \vspace{-5pt}
\subsection{Pricing under True Sensitive Attributes}
% \vspace{-5pt}
\label{sec:4.1}

We begin by considering the setting where the trusted third party (TTP) has access to the true sensitive attributes. The procedure involves two parties: the insurer and the TTP. In the first step, the insurer applies a transformation $T$ to the non-sensitive attributes $X$, producing transformed inputs $\tilde{X} := T(X)$. The insurer then transmits the dataset $\{\tilde{X}, Y\}$ to the TTP. In the second step, the TTP, having access to the corresponding sensitive attributes $D$, estimates the conditional outcome $\mu(\tilde{X}, D)$ and computes the discrimination-free premium $h^*(\tilde{X})$ according to Definition~\ref{def:dfp}. The quantities $\mu(\tilde{X}, D)$ and $h^*(\tilde{X})$ are then returned to the insurer.

Let $\mathcal{F}$ be a hypothesis class. To estimate $\mu(\tilde{X},D)$, let $f = (f_1, \ldots, f_{|\cD|}) \in \cF^{|\cD|}$, define $f_k \in \mathcal{F}$ s.t. $f_k : T(\mathcal{X}) \to \mathbb{R}_+, \forall k \in [|\cD|]$. The TTP then estimates $\mu(\tilde{X}, D)$ by minimizing the following expected risk:

% {\color{blue} need to add relation between $\mu(\tilde{X}, D)$ and $f_k(\tilde{X})$}

% {\color{purple} TZ: done}

%the TTP observes $\{\tilde{X}_i), Y_i, D_i\}_{i=1}^n$, but TTP does not know the transformation $T$. TTP's task consists of two components: 1) learn $\mu(T(X),D)$, 2) output the discrimination-free price $h^*(T(X))$ using $\mu(T(X),D)$ following Definition \ref{def:dfp}. 

\begin{equation}
\label{eq:TTPpopLoss}
    \cR(f) = \sum_{k=1}^{|\cD|} \left( \Ex_{Y, \tilde{X} | D = k} \Bigr[ L \bigr( f_k(\tilde{X}), Y \bigr) \Bigr] \cdot \bbP(D = k) \right),  
\end{equation}
for a generic loss function $L$. This formulation is flexible and agnostic to the choices of transformation $T$, hypothesis class $\mathcal{F}$, and loss function $L$. Using a pre-specified weight $\bbP^*(d)$, the TTP then computes the discrimination-free premium as:
\begin{equation}
\label{eq:TTPDFP}
    h^*(\tilde{X}) = \sum_{k=1}^{|\cD|} f_k(\tilde{X}) \cdot \bbP^*(D = k).
\end{equation}
This procedure is outlined in an algorithmic manner in Algorithm \ref{alg:D} (MPTP).

\begin{algorithm} 
\caption{Multi-party Training Process w.r.t. $D$ (MPTP)}
\label{alg:D}
\textbf{Insurer Input:} data $\{x_i, y_i\}_{i=1}^n$, and hypothesis class $\cU$ (if $T$ is obtained via supervised learning) \\
\textbf{Insurer Output:} $\{\tilde{x}_i\}_{i=1}^n$ \\
\textbf{TTP Input:} data $\{\tilde{x}_i, y_i, d_i\}$, hypothesis class: $\cF$, risk function $\cR(f)$ (Eq. (\ref{eq:TTPpopLoss}))
\begin{algorithmic}
\Repeat
\State train $f$ by minimizing Eq. (\ref{eq:TTPpopLoss}) 
\Until{convergence}
\State compute $\{h^*(\tilde{x}_i)\}_{i=1}^n$ by Eq. (\ref{eq:TTPDFP})
\State \Return{$f, \{h^*(\tilde{x}_i)\}_{i=1}^n$}
\end{algorithmic}
\textbf{TTP Output:} $f, \{h^*(\tilde{x}_i)\}_{i=1}^n$
\end{algorithm}

\textbf{Remark 1:} The framework centers on the use of group-specific score functions $f_1, \ldots, f_{|\mathcal{D}|}$, which offers two key advantages:  1) The framework naturally extends to the setting with privatized sensitive attributes. 2) The computation of $h^*(\tilde{X})$ does not require the disclosure of group membership $D$, enabling its implementation by either the TTP or the insurer.

% {\color{purple} TZ: This is true, since computing $h(\tilde{X})$ only requires knowing $\bbP^*(d)$ and $\mu(\tilde{X}, D)$, for each individual $\tilde{X}$, one compute $\mu(\tilde{X},D=k), \forall k \in [|\cD]$, then average over $\bbP^*(D)$. No group membership needed in this process.}

\textbf{Remark 2.} The framework involves an inherent trade-off between model transparency and model complexity. This trade-off is jointly controlled by the insurer—through the choice of transformation $T$—and the TTP—through the choice of hypothesis class $\mathcal{F}$. For example, if $T$ is set to the identity transformation and $\mathcal{F}$ is chosen as the class of linear models, the resulting pricing rule is maximally transparent, reducing to a (generalized) linear model in terms of the original covariates $X$.

\textbf{Remark 3:} The optimized $f_k$’s are independent of the specific form of $h^*$. More broadly, they can be directly applied to any downstream tasks that depends only on the $f_k$’s themselves and not on how they were optimized. This includes, but is not limited to, computing discrimination-free insurance premiums.

% {\color{purple} TZ: the discrimination-free insurance premium $h^*$ here is an example of an independent downstream task which only utilizes trained $f_k$'s. This remark says our algorithm is not just suitable for computing discrimination-free insurance premiums, but also suitable for any independent downstream tasks that only utilize $f_k$'s}

\bigskip 

\label{eg:1}
\textbf{Example:} Consider an insurer that models insurance claims using a generalized linear model (GLM) within the exponential dispersion family. Both the insurer and the TTP adopt the deviance loss for model training:
\begin{align*}
    L = -2\phi (\ell(\mu,\phi) - \ell_s).
\end{align*}
where $\ell(\mu, \phi)$ is the log-likelihood of the trained model and $\ell_s$ is the saturated log-likelihood. Suppose we have $n$ i.i.d. samples $\{x_i, y_i, d_i\}_{i=1}^n$ drawn from an unknown population $\bbP_{X,D,Y}$. The insurer observes only $\{x_i, y_i\}_{i=1}^n$, while the TTP observes the sensitive attributes $\{d_i\}_{i=1}^n$. The insurer constructs a transformation $\tilde{x}_i = T(x_i)$ using a feed-forward neural network.

Let $u \in \cU$ be a hypothesis from a hypothesis class $\cU$, where $u: \mathcal{X} \to \mathbb{R}_+$. Suppose the neural network has $m$ layers, with $q_j$ hidden units in the $j^{\text{th}}$ layer. For input features $x_i \in \mathbb{R}^{q_0}$, define $z^{(j)}: \mathbb{R}^{q_{j-1}} \to \mathbb{R}^{q_j}$ as the transformation at layer $j$, and denote the composition from layer 1 to $m$ as $z^{(m:1)} = z^{(m)} \circ \cdots \circ z^{(1)}$. The resulting transformation is:
\begin{align*}
    T(x_i) = z^{(m:1)}(x_i) = z^{(m)}(z^{(m-1)}( \cdots( z^{(1)}(x_i)))),
\end{align*}
which is learned by the insurer via minimizing the empirical risk:
\begin{equation}
\label{eq:step1empLoss}
    \hat{\cR}(u) = \sum_{i=1}^n L(u(x_i), y_i).
\end{equation}

Once the transformation is learned, the insurer transmits $\{\tilde{x}_i\}_{i=1}^n$ (an $n \times q_m$ matrix) along with $\{y_i\}_{i=1}^n$ to the TTP. Then, TTP estimates $\mu(\tilde{x}_i, D = k) = f_k(\tilde{x}_i)$ by minimizing the empirical risk:
\begin{equation}
\label{eq:TTPempLoss}
    \hat{\cR}(f_1, \ldots, f_{|\cD|}) = \sum_{i=1}^n \sum_{k=1}^{|\cD|} L(f_k(\tilde{x}_i), y_i) \cdot \ones\{D_i = k\},
\end{equation}
As an example, each $f_k \in \cF$ could be specified as a linear model such that:
\begin{align*}
f_k(\tilde{x}_i)) = \beta_0^k + (\tilde{x}_i)_1\beta_1^k + \cdots + (\tilde{x}_i)_{q_m}\beta_{q_m}^k.    
\end{align*}
% {\color{red} 1) in above, $u(X_i)$ is the estimate of $T(X_i)$; 2) notation $z(\tilde{X}_i)_1$ not consistent? }
% {\color{blue} TZ: 1) $u \in \cU$ is a neural network used to estimate $Y$, so it $u(X_i)$ is an estimate of $Y_i$. $T(X_i)$ is the transformed $X_i$ we get in the last layer of $u$. 2) Yes, $z(\tilde{X}_i)_j$ does not seem to make sense, it should just be $(\tilde{X}_i)_j, j = 1, \ldots,q_m$, since the above example is essentially fitting a linear model w.r.t. $\tilde{X} \in \bbR^{q_m}$.}

Finally, the TTP computes the empirical discrimination-free price using Definition~\ref{def:dfp}:
\begin{equation}
\label{eq:TTPempDFP}
    \hat{h}^*(\tilde{x}_i) = \sum_{k=1}^{|\cD|} \hat{f}_k(\tilde{x}_i) \cdot \hat{\bbP}(D = k),
\end{equation}
and returns the estimated conditional means and fair prices $\{ \hat{\mu}(\tilde{x}_i, D), \hat{h}^*(\tilde{x}_i) \}_{i=1}^n$ to the insurer.

\textbf{Remark.} In this example, the transformation $T$ is obtained by training a neural network (represented by some function $u \in \cU$). However, our framework does not impose any constraints on how $T$ is constructed. In principle, $\tilde{X}$ may consist of engineered features derived through supervised or unsupervised learning methods, or alternatively, privatized representations designed specifically to ensure secure data transmission.

% \vspace{-5pt}
\subsection{Pricing under Privatized Sensitive Attributes with Known Noise Rates}
% \vspace{-5pt}
\label{sec:4.2}

We now consider a setting where the true sensitive attributes are not directly observable by the TTP. Instead, the TTP only has access to their privatized versions. This situation arises frequently in practice due to the adoption of privacy-preserving mechanisms during data collection or transmission (see Section~\ref{sec:1} for a detailed discussion). The central inquiry revolves around how the TTP can compute a fair premium—i.e., minimize the risk in Eq.~(\ref{eq:TTPpopLoss})—without direct access to the true sensitive attributes \( D \).

To address this challenge, we employ the concept of local differential privacy (LDP) and the randomized response mechanism in our framework. Let \( S \) denote the privatized version of the sensitive attribute \( D \). An $\epsilon$-LDP mechanism \( Q \) is defined as:

\begin{definition}
\label{def:pm}
A randomized mechanism \( Q(s \mid d) \) satisfies \(\epsilon\)-local differential privacy (LDP) if:
\[
\max_{s \in \cD,\, d, d' \in \cD} \frac{Q(S = s | D = d)}{Q(S = s | D = d')} \le e^{\epsilon}.
\]
This condition guarantees that the output \(S\) does not reveal much about the true input \(D\), ensuring strong individual-level privacy. Under this constraint, a standard privatization rule is the randomized response mechanism, defined as \citep{Warner_1965,10.5555/2946645.2946662}:
\[
Q(s | d) =
\begin{cases}
\pi := \frac{e^{\epsilon}}{|\mathcal{D}| - 1 + e^{\epsilon}}, & \text{if } s = d, \\
\bar{\pi} := \frac{1}{|\mathcal{D}| - 1 + e^{\epsilon}}, & \text{if } s \ne d,
\end{cases}
\]
where \(|\cD|\) is the number of possible sensitive attribute values. The output \(s\) is sampled from \(Q(\cdot | d)\) independently of covariates \(X\) and outcome \(Y\).
\end{definition}

The primary advantage of employing LDP is that it ensures the data collector cannot reliably infer the true value of the sensitive attribute for any individual observation, regardless of the accuracy of the reported information \citep{10.5555/3524938.3525593}. Consequently, any model trained on such privatized data inherently preserves differential privacy with respect to the sensitive attributes.

The setup mirrors that of Section~\ref{sec:4.1}: the insurer observes \( \{x_i, y_i\}_{i=1}^n \), applies a transformation \( T \), and transmits \( \{\tilde{x}_i, y_i\}_{i=1}^n \) to the TTP. The TTP's goal is to estimate \( \mu(X, D) \) using the insurer-provided data in combination with the privatized sensitive attributes \( \{s_i\}_{i=1}^n \). To enable this, Lemma \ref{lemma:Risk-LDP} establishes a population equivalent risk under privacy mechanism $Q(s_i|d_i)$ and Theorem~\ref{theorem:GEB} provides corresponding statistical guarantees for the proposed estimation procedure.

\begin{lemma}
\label{lemma:Risk-LDP}

Given the privacy parameter $\epsilon$, minimizing the risk (Risk-LDP) defined by Eq. (\ref{eq:Risk-LDP}) under $\epsilon$-LDP w.r.t. privatized sensitive attributes $S$ is equivalent to minimizing Eq. (\ref{eq:TTPpopLoss}) w.r.t. true sensitive attributes $D$ at the population level:
\begin{equation}
\label{eq:Risk-LDP}
    \cR^{LDP}(f) = \sum_{k=1}^{|\cD|} \sum_{j=1}^{|\cD|} \left( \boldsymbol{\Pi}_{kj}^{-1} \Ex_{Y, \tilde{X} \mid S = j} \left[ L \bigr(f_k(\tilde{X}),Y \bigr) \right] \cdot \sum_{l = 1}^{|\cD|} \boldsymbol{T}_{kl}^{-1} \bbP(S = l) \right),
\end{equation}
where $\boldsymbol{\Pi}^{-1}$ and $\boldsymbol{T}^{-1}$ are $|\cD| \times |\cD|$ matrices with row-sum of one. Denote $\bbP(D = k)$ as $p_k$, then each entry of $\boldsymbol{\Pi}^{-1}$ and $\boldsymbol{T}^{-1}$ takes the following forms:
\[
\begin{cases}
\boldsymbol{\Pi}_{ii}^{-1} = \frac{\pi + |\cD| - 2}{|\cD|\pi - 1}\frac{\pi p_i + \sum_{d''\setminus i} \bar{\pi}p_{d''}}{p_i}, \text{for $i \in |\cD|$}\\
\boldsymbol{\Pi}_{ij}^{-1} = \frac{\pi - 1}{|\cD|\pi - 1}\frac{\bar{\pi} p_i + \sum_{d''\setminus i} \pi p_{d''}}{p_i}, \text{for $i,j \in |\cD|$ s.t.,$i \ne j$}
\end{cases},
\]
and 
\[
\begin{cases}
\boldsymbol{T}_{ii}^{-1} = \frac{\pi + |\cD| - 2}{|\cD|\pi - 1}, \text{for $i \in |\cD|$}\\
\boldsymbol{T}_{ij}^{-1} = \frac{\pi - 1}{|\cD|\pi - 1}, \text{for $i,j \in |\cD|$ s.t.,$i \ne j$}
\end{cases}.
\]
\end{lemma}
Empirically, for a given policyholder $i$, the TTP computes $\hat{h}^*(\tilde{x}_i)$ using learned $\{\hat{f}_k(\tilde{x}_i)\}_{k=1}^{|\cD|}$, and returns $\hat{h}^*(\tilde{x}_i)$ and $\hat{\mu}(\tilde{x}_i,D=k)$ for $k=1,\ldots,|\cD|$ to the insurer. We summarize this procedure in an algorithmic manner in Algorithm \ref{alg:S-NE} (MPTP-LDP).

\begin{algorithm}
\caption{Multi-party Training Process w.r.t. $S$ (MPTP-LDP)} 
\label{alg:S-NE}
\textbf{Insurer Input:} data $\{x_i, y_i\}_{i=1}^n$, hypothesis class $\cU$ (if $T$ is obtained via supervised learning), and hypothesis class $\cK$ (if $X^*$ is obtained via supervised learning)  \\
\textbf{Insurer Output:} $\{\tilde{x}_i\}_{i=1}^n, \{x_i^*\}_{i=1}^n$ \\
\textbf{TTP Input:} data: $\{\tilde{x}_i, y_i, s_i\}_{i=1}^n, \{x_i^*, s_i\}_{i=1}^n$, hypothesis class $\cG$, risk function $\forall k \in [n_1]$, $\cR(g_k) = \sum_{j=1}^{m} L(g_k(X_{k,j}^*), S_{k,j})$ (see Lemma \ref{lemma:NE} and Procedure \ref{procedure:C_1}), hypothesis class $\cF$, and risk function $\cR(f)$ (Eq. (\ref{eq:Risk-LDP})), 
\begin{algorithmic}
\If { ($\pi, \bar{\pi}$) are unknown}
    \State compute $\hat{\pi}_k, \hat{\bar{\pi}}_k$ for $k \in [n_1]$ using Lemma \ref{lemma:NE} 
    \State compute $\hat{C}_1$ using $\hat{\pi}_k, \hat{\bar{\pi}}_k$ for $k \in [n_1]$, as described in Procedure \ref{procedure:C_1}
    \State compute $\hat{\pi}, \hat{\bar{\pi}}$ using $\hat{C}_1$ defined in Theorem \ref{theorem:GEB}
    \State compute $\hat{\boldsymbol{\Pi}}^{-1}$ using $\hat{\pi}, \hat{\bar{\pi}}$ as defined in Lemma \ref{lemma:Risk-LDP}
\Else 
    \State compute $\boldsymbol{\Pi}^{-1}$ using $\pi, \bar{\pi}$ as defined in Lemma \ref{lemma:Risk-LDP}
\EndIf
\Repeat
\State train $f$ by minimizing Eq. (\ref{eq:Risk-LDP}) 
\Until{convergence}
\State compute $\{h^*(\tilde{x}_i)\}_{i=1}^n$ using Eq. (\ref{eq:TTPDFP})
\State \Return{$f, \{h^*(\tilde{x}_i)\}_{i=1}^n$} 
\end{algorithmic}
\textbf{TTP Output:} $f, \{h^*(\tilde{x}_i)\}_{i=1}^n$
\end{algorithm}

\textbf{Remark:} The use of group-specific score functions enables straightforward construction of an equivalent risk for Eq. (\ref{eq:TTPpopLoss}) using only $S_i$. It is crucial not to view it as a limitation of our approach. As discussed in Section \ref{sec:4.1}, obtaining a closed-form equivalence is generally not feasible with a conventional score function $f(\tilde{X},D)$. Existing methods tackling similar challenges often rely on surrogate risks or confine themselves to specific loss functions \citep{li2016learning}.

\begin{theorem}
\label{theorem:GEB}
Let $\boldsymbol{T}^{-1}$ be the matrix in Lemma \ref{lemma:Risk-LDP} with entries $\boldsymbol{T}_{kk}^{-1} = C_1:= \frac{\pi + |\cD| - 2}{|\cD|\pi} - 1$ and $\boldsymbol{T}_{kj}^{-1} = C_2 := \frac{\pi - 1}{|\cD|\pi - 1}, (k \ne j)$. Let $L: \cY \times \cY \to \bbR_+$ be a loss function bounded above by $M$. Let $\hat{f} \in \underset{f \in \cF^{|\cD|}}{\arg\min} \hat{\cR}^{LDP}(f)$ and $f^* \in \underset{f \in \cF^{|\cD|}}{\arg\min} \cR(f)$. Define the LDP-weighted loss class $\cF_{LDP}=\{(\tilde{x},y,s) \mapsto \sum_{k=1}^{|\cD|} \boldsymbol{T}_{ks}^{-1} L(f_k(\tilde{x}),y) \mid f_k \in \cF \}$ and let $\Re_n(\cF_{LDP})$ be the Rademacher complexity of $\cF_{LDP}$. Then, for any $\delta \in (0,1)$, w.p. at least $1 - \delta$:
\[
\cR(\hat{f}) \le \cR(f^*) + 4 \Re_n(\cF_{LDP}) + 2M(C_1 + (|\cD| - 1)|C_2|) \sqrt{\log(1/\delta)/2n}.
\]
\end{theorem}

\textbf{Remark.} The bound grows linearly with the number of sensitive groups \( |\mathcal{D}| \). However, in typical insurance applications, \( |\mathcal{D}| \) tends to be small—for example, when categorizing policyholders by binary gender, a few age brackets, or self-reported race/ethnicity categories. In cases where \( |\mathcal{D}| \) is large, categorical embedding techniques (see \cite{ShiShi2023risk}) can be used to reduce dimensionality, provided that it is permissible under applicable regulations.

% \vspace{-5pt}
\subsection{Pricing under Privatized Sensitive Attributes with Unknown Noise Rates}
% \vspace{-5pt}
\label{sec:4.3}

This section extends the framework developed in Section \ref{sec:4.2} to the more challenging setting where the noise rates of the privacy mechanism are unknown. It is essential to note that constructing a population-equivalent risk under the LDP requires knowledge of the conditional probabilities $\pi$ and $\bar{\pi}$. However, obtaining such information often proves challenging in practice, particularly when sensitive attributes are subject to measurement errors (refer to Section \ref{sec:1} for detailed discussions).

Within our multi-party framework, we consider a setup akin to that of Section \ref{sec:4.2}, with the key distinction being that the TTP lacks knowledge of the true parameters of the privacy mechanism \( Q(s_i | d_i) \). To tackle this, we propose a methodology wherein the TTP first infers the noise rates $\pi$ and $\bar{\pi}$ from the data, and then uses these estimates to construct the population-equivalent risk following the approach outlined in Section \ref{sec:4.2}. 

The following lemma outlines the core idea behind estimating the noise rate $\pi$ using the concept of an anchor point—a covariate value where the sensitive attribute is known with certainty.

\begin{lemma}
\label{lemma:NE}
Consider an $\epsilon$-LDP mechanism with parameters $\pi \in (\frac{1}{|\mathcal{D}|}, 1]$ and $\bar{\pi} \in [0, \frac{1}{|\mathcal{D}|})$. Suppose there exists an anchor point \( x^* \) in the dataset s.t. \( \mathbb{P}(D = j^* | x^*) = 1 \) for some \( j^* \in [|\mathcal{D}|] \). Then:
\[
\pi = \mathbb{P}(S = j^* | x^*).
\]
Empirically, for a dataset with \( n \) observations, let:
\[
\boldsymbol{\eta}_n(j^*) = \left( \hat{\mathbb{P}}(S = j^* | x_1), \ldots, \hat{\mathbb{P}}(S = j^* | x_n) \right),
\]
then an estimator for \( \pi \) is given by:
\[
\hat{\pi} = \left\| \boldsymbol{\eta}_n(j^*) \right\|_{\infty}.
\]
\end{lemma}

In addition to Lemma~\ref{lemma:NE}, we introduce a set of assumptions and an accompanying estimation procedure that allow us to establish statistical guarantees when the noise rates are unknown. We begin by presenting a procedure for estimating $C_1$, a key quantity in Theorem~\ref{theorem:GEB}, which in turn facilitates the recovery of the underlying noise parameters.

\begin{procedure}{\bf{($C_1$-estimation)}}
\label{procedure:C_1}

\textit{Step 1: Grouping.} 
We partition the dataset \(\{ x_i, s_i \}_{i=1}^n\) into \(n_1\) equally sized groups, each containing \(m = n / n_1\) samples.

\textit{Step 2: Within-group Estimation.}
For each group \(k \in [n_1]\), let \(\{ x_{k,j}^*, s_{k,j} \}_{j=1}^{m}\) denote its samples. Within group \(k\), we compute the vector:
\[
\boldsymbol{\eta}_m^k(j^*) = \left( \hat{\mathbb{P}}_k(S = j^* | x_{k,1}), \ldots, \hat{\mathbb{P}}_k(S = j^* | x_{k,m}) \right),
\]
where \(\hat{\mathbb{P}}_k(S = j^* | X)\) is the estimated conditional probability under group \(k\). We then estimate the privacy parameter \(\hat{\pi}_k\) by taking the infinity norm:
\[
\hat{\pi}_k = \left\| \boldsymbol{\eta}_m^k(j^*) \right\|_\infty = \max_{j \in [m]} \hat{\mathbb{P}}_k(S = j^* | x_{k,j}).
\]
Using this, we compute a group-specific estimate of the scaling factor:
\[
\hat{C}_{1,k} = \frac{\hat{\pi}_k + |\mathcal{D}| - 2}{|\mathcal{D}| \hat{\pi}_k - 1}.
\]

\textit{Step 3: Aggregation.} 
The final estimator \(\hat{C}_1\) is obtained by averaging across groups:
\[
\hat{C}_1 = \frac{1}{n_1} \sum_{k=1}^{n_1} \hat{C}_{1,k}.
\]
    
\end{procedure} 

Next, we formalize two technical assumptions that underpin the derivation of Theorem~\ref{theorem:GEB-NE}, which builds on the estimated noise rates.

\textbf{Assumption A (Sub-exponentiality).}  
\label{assumption:A}
For each group \(k \in [n_1]\), define \(\hat{g}_k(x) = \hat{\mathbb{P}}_k(S = j^* | x)\). Assume there exists a constant \(M_g > 0\) s.t. the sub-exponential norm of \(\hat{C}_{1,k}\) satisfies:
\[
\left\| \hat{C}_{1,k} \right\|_{\psi_1} = \left\| \min_{i \in [m]} \frac{\hat{g}_k(x_{k,i}) + |\mathcal{D}| - 2}{|\mathcal{D}| \hat{g}_k(x_{k,i}) - 1} \right\|_{\psi_1} \le M_g, \forall k \in [n_1],
\]
where \(\|\cdot\|_{\psi_1}\) denotes the sub-exponential norm defined by \(\|X\|_{\psi_1} = \inf\{t > 0 : \mathbb{E}[e^{|X|/t}] \le 2\}\).

\textbf{Assumption B (Nearly Unbiasedness).}  
\label{assumption:B}
For each group \(k \in [n_1]\), \(\hat{C}_{1,k}\) is a nearly unbiased estimator of $C_1$. That is, there exists \(\theta > 0\) such that:
\[
\left| \mathbb{E}[\hat{C}_{1,k}] - C_1 \right| < \theta \quad \text{for all } k \in [n_1].
\]

Combining the estimation procedure with these assumptions, we arrive at the following statistical guarantee:
\begin{theorem}
\label{theorem:GEB-NE}

Let $\hat{\boldsymbol{T}}^{-1}$ be the same matrix as in Theorem \ref{theorem:GEB} but whose entries are estimated via Procedure \ref{procedure:C_1}. Let $L: \cY \times \cY \to \bbR_+$ be a loss function bounded above by $M$. Define the empirical LDP risk using $\hat{\boldsymbol{T}}^{-1}$ by: $\hat{\cR}_{\hat{\boldsymbol{T}}^{-1}}^{LDP}(f):= \frac{1}{n} \sum_{i=1}^n \sum_{k=1}^{|\cD|} \hat{\boldsymbol{T}}_{kS_i}^{-1} L(f_k(\tilde{x}_i),y_i)$. Let $\hat{f} \in \underset{f \in \cF^{|\cD|}}{\arg\min} \hat{\cR}_{\hat{\boldsymbol{T}}^{-1}}^{LDP}(f)$ and $f^* \in \underset{f \in \cF^{|\cD|}}{\arg\min} \cR(f)$. For $\tilde{\epsilon} > \theta > 0$, define the enlarged LDP-weighted loss class $\cF_{\tilde{\epsilon}} = \{ (\tilde{x},y,s) \mapsto \sum_{k=1}^{|\cD|} \boldsymbol{T}_{kS}^{-1}(c) L(f_k(\tilde{x},y) \mid f_k \in \cF, c \in [1, C_1 + \tilde{\epsilon}] \}$, where $\boldsymbol{T}^{-1}(c)$ denotes the inverse randomized-response matrix whose diagonal entries are $c$ and off-diagonal entries are $\frac{1-c}{|\cD| - 1}$. Let $\Re_n(\cF_{\tilde{\epsilon}})$ be the Rademacher complexity of $\cF_{\tilde{\epsilon}}$ If $M_g + \frac{C_1 + \theta}{\ln 2} > \tilde{\epsilon} > \theta$ and $n_1 \ge \frac{1}{c(\tilde{\epsilon} - \theta)^2}(M_g + \frac{C_1 +\theta}{\ln 2})^2 \ln(2/\delta)$ for some absolute constant $c > 0$. Then, for any $\delta \in (0, \frac{1}{3})$, w.p. at least $1 - 3\delta$,
\[
\cR(\hat{f}) \le \cR(f^*) + 4 \Re_n(\cF_{\tilde{\epsilon}}) + 2M(C_1 + (|\cD|-1)|C_2|+2\tilde{\epsilon}) \sqrt{\frac{\log(1/\delta)}{2n}} + 4M\tilde{\epsilon}.
\]

\end{theorem}

\textbf{Remark 1.} The error bound accounts for the estimation uncertainty in \(\hat{C}_1\). Despite \(\hat{C}_1\) does not appear explicitly in the bound, its variability is incorporated into the error term \(\tilde{\varepsilon}\).

% {\color{red} I need some clarity on \(\tilde{\varepsilon}\), you call it a parameter, do we also estimate it?}
% {\color{blue} TZ: $\tilde{\epsilon}$ is a parameter introduced in the bound that w.p. $1-\delta$, the event $\{|\hat{C}_1 - C_1| \le \tilde{\epsilon}\}$ holds. This is equivalent to w.p. $1 - \delta$ the event $\{C_1 - \tilde{\epsilon} \le \hat{C_1} \le C_1 + \tilde{\epsilon}\}$ holds. Since we stated a condition on the range of $\tilde{\epsilon}$ in the theorem, so maybe using $C_1 + \tilde{\epsilon}$ makes more sense than $\hat{C}_1$.}

%Although \(\hat{C}_1\) does not appear explicitly in the bound, it plays a critical role in linking the estimation procedure to the theoretical guarantees. Its estimation variability is absorbed into the error parameter \(\tilde{\varepsilon}\) (see proof in Appendix \ref{proof:theoremGEB-NE}).

\textbf{Remark 2.} Increasing \(n_1\) generally improves the accuracy of \(\hat{C}_1\), as it averages over more independent estimates. However, if \(n_1\) is chosen too large, the resulting group size \(m = n / n_1\) may be too small for Assumption A to hold. In practice, careful tuning of \(n_1\) is advised to ensure both statistical stability and validity of assumptions.

% \textbf{Remark 2:} As $n_1$ increases, $\hat{C}_1$ is more accurate, as it is the average of $n_1$ independent variables, resulting in a tighter bound. However, blindly choosing a large $n_1$ is not recommended, since Assumption A will not hold if $m = \frac{n}{n_1}$ is too small. Some light tuning may help select $n_1$ in practice.

\textbf{Remark 3:} The bound is more adversely affected by underestimation of $\pi$. Note that the term playing a key role in the error bound is $\frac{1}{\pi - \frac{1}{|\cD|}}$. Hence, when $\pi$ is close to $\frac{1}{|\cD|}$, underestimating $\pi$ can be far more detrimental than overestimating it - a point we illustrate in Section \ref{sec:5}.

% {\color{red} Further, we provide an empirical study on the effect of estimation error on the model performance in Section \ref{sec:5}.} {\color{blue} need to think more on this? I thought it is one-sided??}

% {\color{blue} need to check the numbering???}

% \vspace{-5pt}
\section{Numerical Experiments}
% \vspace{-5pt}
\label{sec:5}

% In this section, we first create a carefully designed toy dataset to illustrate our main results (Theorem \ref{theorem:GEB}, Theorem \ref{theorem:GEB-NE}) and then evaluate the performance of our proposed method using two real-world datasets, demonstrating that the experiment results on both synthetic and real data are in support of our theories. Building on these findings, we further provide some practical guidelines for implementing our method under various conditions.

In this section, we present a series of carefully designed numerical experiments to illustrate and support our main theoretical results in Theorems~\ref{theorem:GEB} and~\ref{theorem:GEB-NE}, which address pricing under known and unknown noise rates in privatized sensitive attributes, respectively.

We simulate a synthetic dataset with $n=5,000$ observations. The response variable $Y$ denotes a continuous insurance claim cost. There are two non-sensitive attributes $X$, representing the age of the policyholder and the smoking status. The age variable $X_{\text{A}}$ takes integer values between $18$ and $80$, and the smoking status variable $X_{\text{S}}$ is binary, indicating smoker (S) or non-smoker (NS). The sensitive attribute $D$, representing gender, is also binary (male (M) or female (F)). The data-generating process for the claim cost is specified as:
\[
\begin{aligned}
    Y =& 100 + 4\times X_{\text{A}} + 100 \times \boldsymbol{1}\{X_{\text{S}} = \text{S}\} + 120 \times \boldsymbol{1}\{D = \text{F}\} \\ 
    &+ 200 \times \boldsymbol{1}\{(X_{\text{A}} \in [20, 40]) \cap (D = \text{F})\} + \epsilon,\ \ \ \text{$\epsilon \sim N(0, 40^2)$}.
\end{aligned}
\]
The covariates are generated under the following assumptions: 1) $X_{\text{A}} \perp D$ and $X_{\text{A}} \perp X_{\text{S}}$; 2) $\bbP(X_{\text{S}} = \text{S}) = 0.3$, $\bbP(D = \text{F}) = 0.45$; 3) $\bbP(D = \text{F} | X_{\text{S}} = \text{S}) = 0.8$.

In all experiments, we use the identity transformation $T(X) = X$ and set the target marginal distribution $\bbP^*(D) = \hat{\bbP}(D)$ when computing the discrimination-free premium $\hat{h}^*(X)$. All models are trained using a three-layer feed-forward neural network with five neurons per hidden layer, optimized with mean squared error (MSE) loss.

\subsection{Premiums with Discrimination}
% \subsection{Synthetic Data}
% \label{sec:5.0}

\begin{figure}[H]
    \centering
    \begin{subfigure}{\textwidth}
        \centering
        \includegraphics[width=\textwidth]{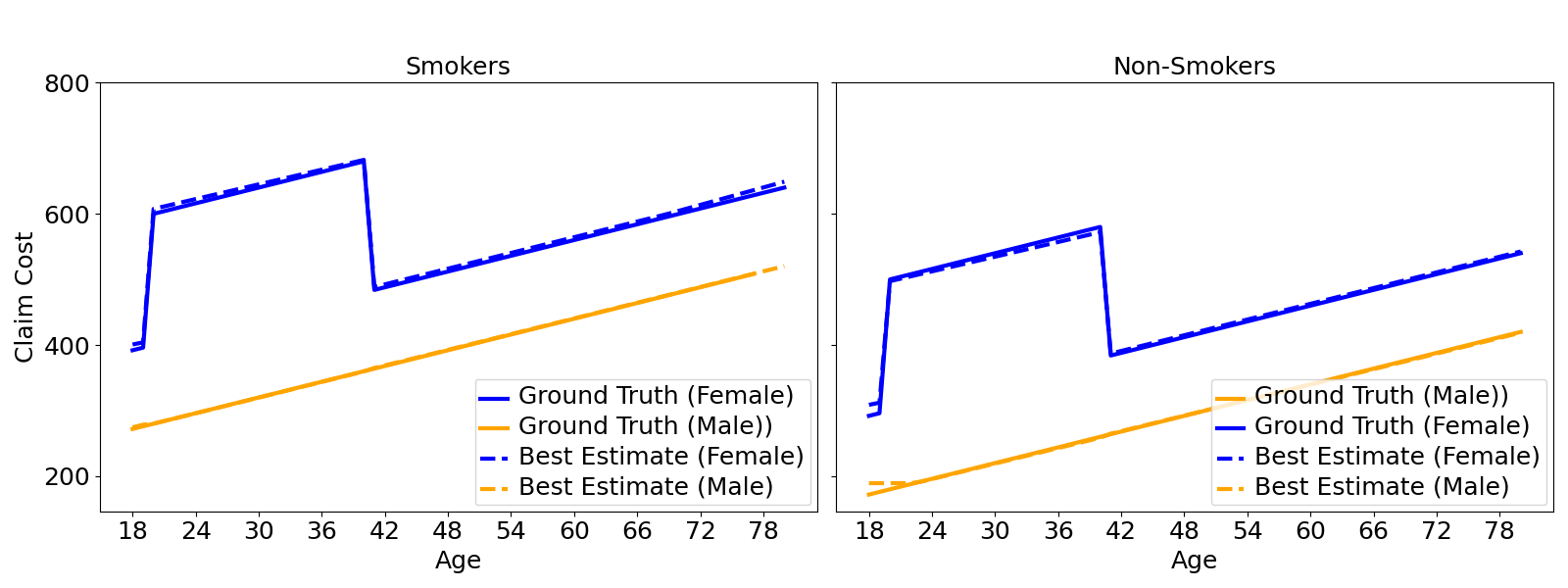}
        \caption{Premiums with direct discrimination}
        \label{fig:toy_gt_r1}
    \end{subfigure}
    \begin{subfigure}{\textwidth}
        \centering
        \includegraphics[width=\textwidth]{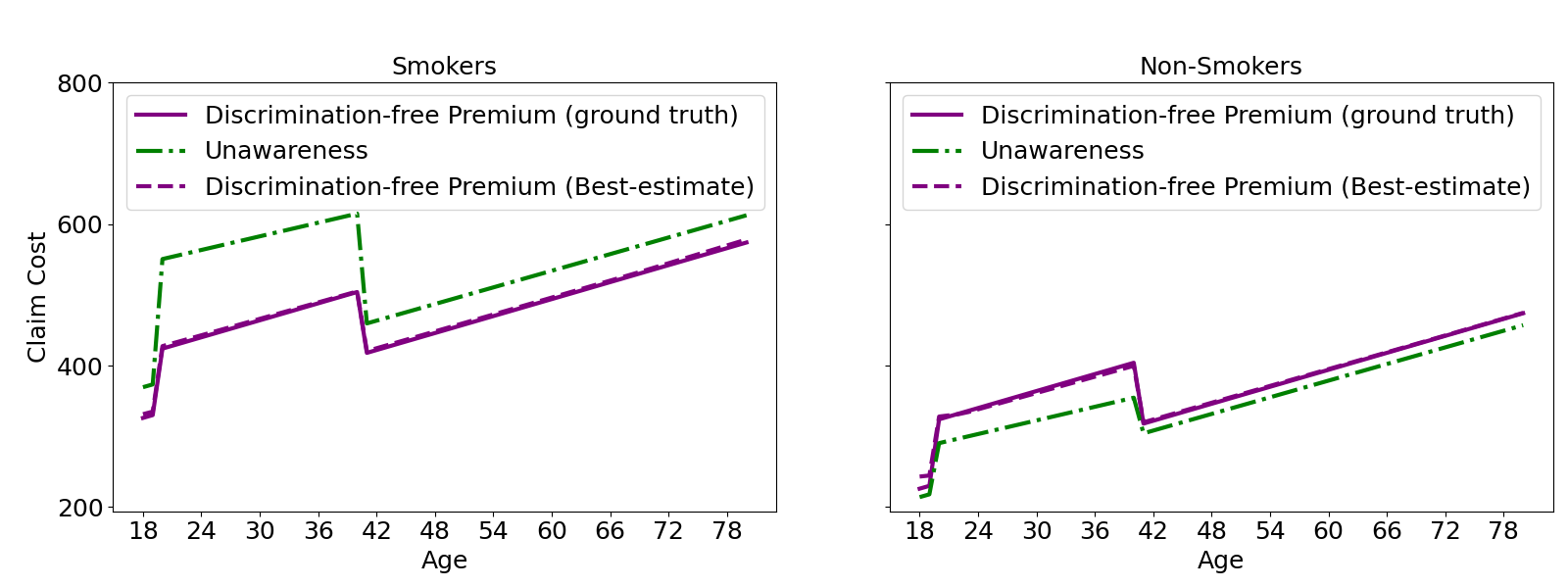} 
        \caption{Premiums with indirect discrimination}
        \label{fig:toy_gt_dfp_r1}
    \end{subfigure}
    \caption{Comparison of discriminatory premiums by risk class.}
    \label{fig:toy_gt_all_r1}
\end{figure}

In the first experiment, we compare several premium-setting strategies introduced in Section \ref{sec:3}. Assuming the sensitive attribute is accessible by the insurer, we compute discriminated premiums with direct and indirect discrimination, which are shown in Figure \ref{fig:toy_gt_all_r1}. 

Panel (a) displays the directly discriminatory premium, where gender is explicitly used to compute group-specific prices. This corresponds to the best-estimated price defined in Definition~\ref{def:bep}, obtained using Algorithm \ref{alg:D} (MTPT). As a benchmark, we overlay the true conditional expected costs. The best-estimate model accurately captures the ground truth for all risk classes, verifying the consistency of our model specification.

Panel (b) contrasts the unawareness premium (Definition \ref{def:up})—obtained by excluding gender from the pricing model—with the discrimination-free premium (Definition \ref{def:dfp}). Although gender is not explicitly used, the unawareness premium remains discriminatory due to indirect discrimination, driven by the statistical dependence between gender and smoking status in the data generating process. This illustrates that omitting sensitive attributes alone does not guarantee fairness in pricing, highlighting the importance of our fairness-aware methodology.

% {\color{red} TZ: I tried to use (MPTP Best-estimate), but that would make the legend for Fig (\ref{fig:toy_gt_dfp_r1}) too lengthy, so maybe just include a sentence in the text that the best-estimate is estimated using MPTP, as described in section \ref{sec:4.1}.}

\subsection{Privatized Sensitive Attributes with Known Noise Rates}

In our second experiment, we examine the setting where the sensitive attribute is privatized using a known noise rate. The privatized sensitive attribute $S$ is generated via the randomized response mechanism (Definition \ref{def:pm}) under three pre-specified noise levels corresponding to $\pi=0.9$, $0.8$, and $0.7$. This experiment illustrates how the performance of our method (Theorem~\ref{theorem:GEB}) is affected by the level of noise in the privatization process.

Recall that when sensitive attributes are privatized, the key challenge in computing the discrimination-free premium $h^*(X)$ lies in accurately estimating the group-conditional expectation $\mu(X,D)$. Therefore, rather than reporting final premiums, we focus on evaluating the estimation accuracy for $\mu(X,D)$, which more directly reflects the effectiveness of the method.

\begin{figure}[H]
    \centering
    \begin{subfigure}{\textwidth}
    \centering
        \includegraphics[width=0.95\textwidth]{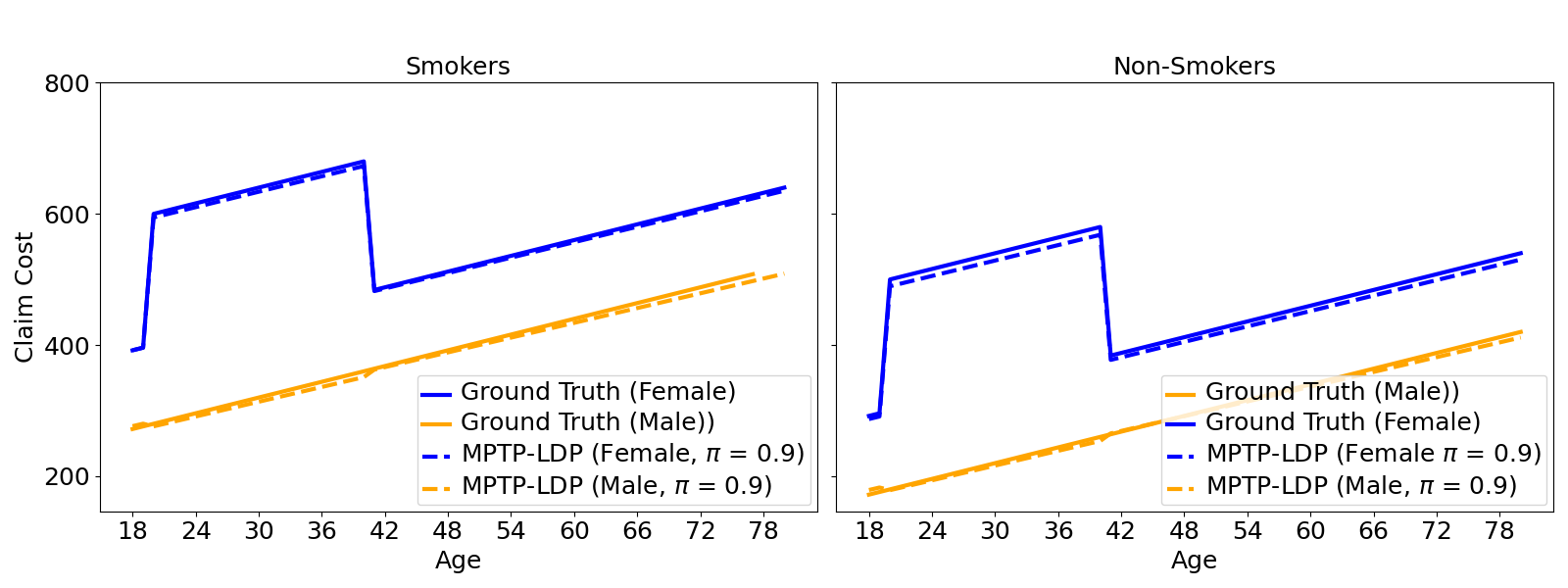}
        \caption{Low noise rate - $\pi = 0.9$}
        \label{fig:toy_0.9_r1}
    \end{subfigure}
    \begin{subfigure}{\textwidth}
    \centering
        \includegraphics[width=0.95\textwidth]{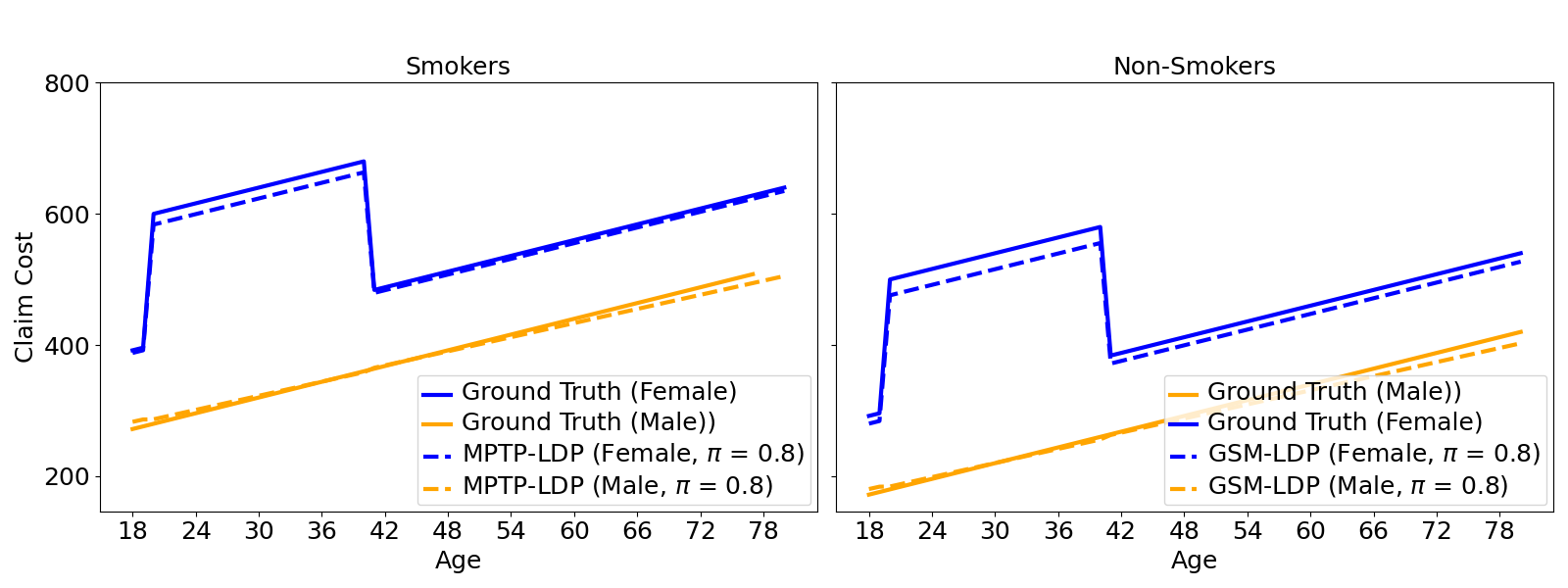} 
        \caption{Medium noise rate - $\pi = 0.8$}
        \label{fig:toy_0.8_r1}
    \end{subfigure}
    \begin{subfigure}{\textwidth}
    \centering
        \includegraphics[width=0.95\textwidth]{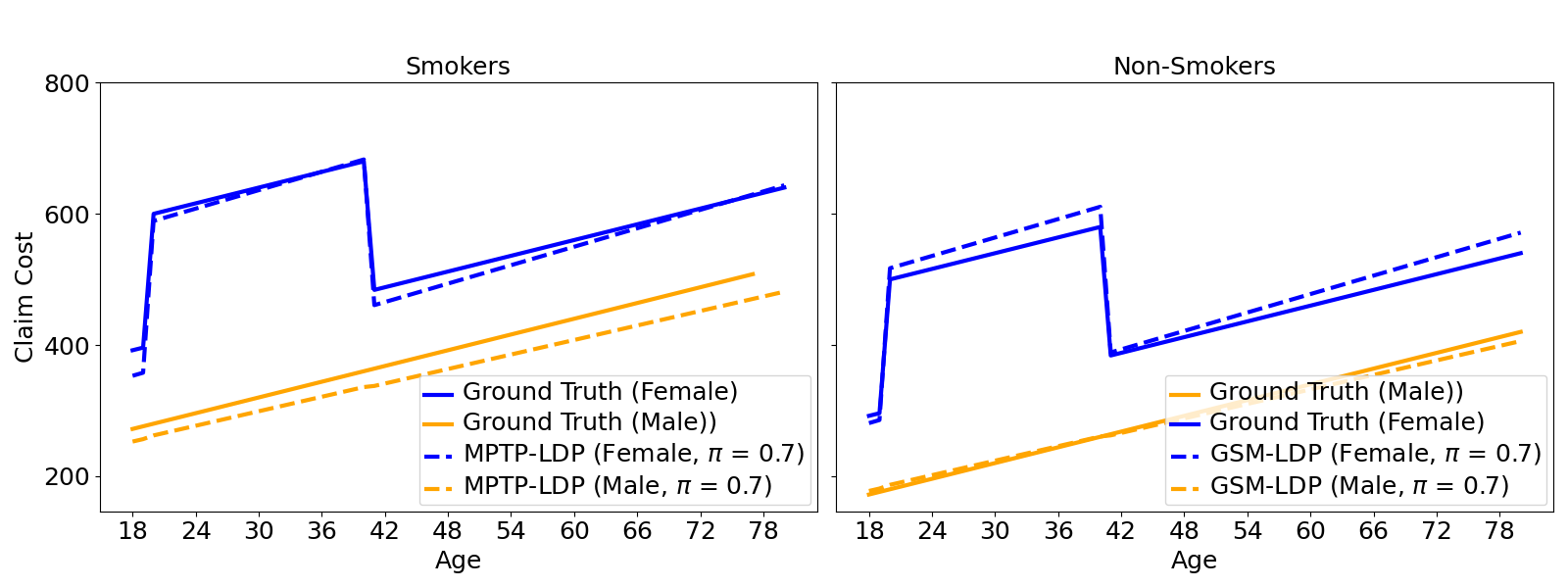} 
        \caption{High noise rate - $\pi = 0.7$}
        \label{fig:toy_0.7_r1}
    \end{subfigure}
        \caption{Group-specific expected claim costs under privatized sensitive attribute with known noise rates.}
    \label{fig:toy_exp2}
\end{figure}

Figure~\ref{fig:toy_exp2} displays the estimated group-specific prices across different noise levels. As expected, estimation accuracy deteriorates as $\pi$ decreases—that is, as more noise is introduced into $D$. This observation is consistent with the theoretical prediction in Theorem~\ref{theorem:GEB}. Notably, the degradation from  $\pi=0.9$ to $\pi=0.8$ is relatively mild, while the drop from $\pi=0.8$ to $\pi=0.7$ is more pronounced. This nonlinearity highlights the steep increase in generalization error as $\pi$ approaches the non-informative limit $\frac{1}{|\cD|}$. Thus, in practice, we recommend avoiding overly aggressive noise levels in the randomized response mechanism, as the resulting accuracy loss may outweigh the privacy gains.

\subsection{Privatized Sensitive Attributes with Unknown Noise Rates}

In the final experiment, we consider a more realistic scenario where the true noise rate $\pi$ is unknown and must be estimated from data (Lemma~\ref{lemma:NE}). Since this estimation is inherently error-prone, it is important to understand how sensitive our method is to the estimation error in $\hat{\pi}$.

\begin{figure}[H]
    \begin{subfigure}{\textwidth}
    \centering
        \includegraphics[width=0.95\textwidth]{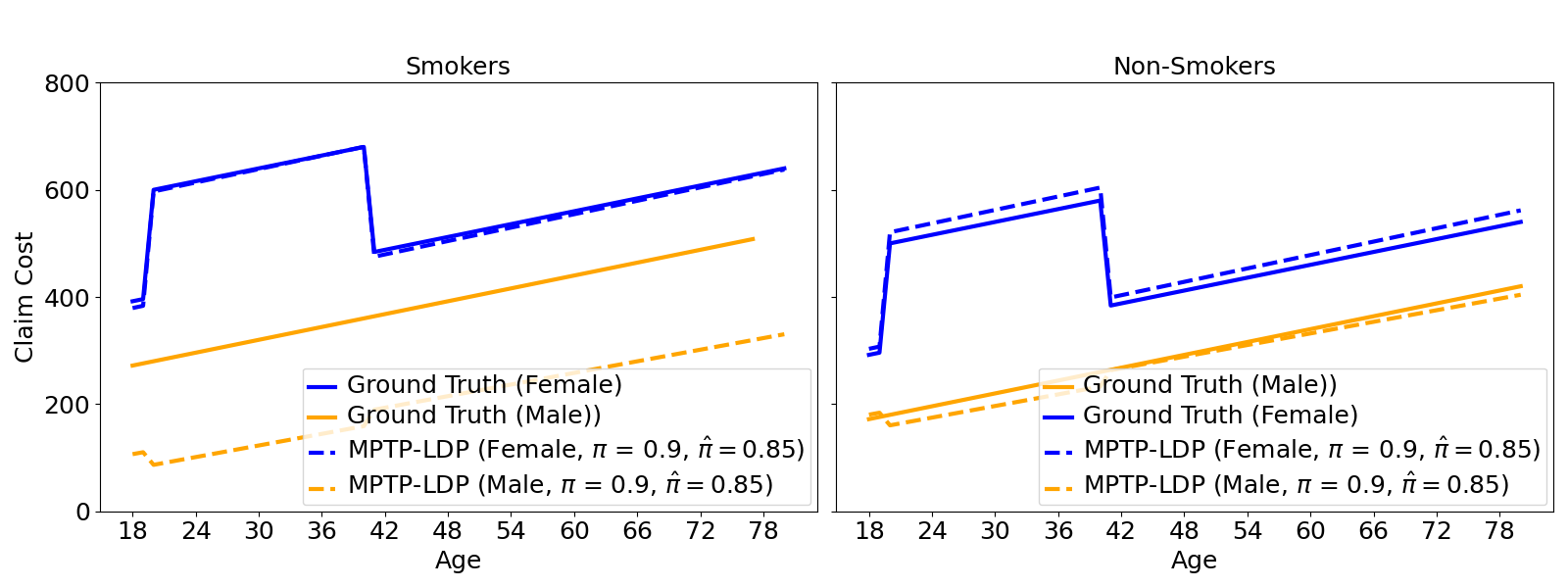}
        \caption{Underestimated noise rate with $\pi = 0.9$}
        \label{fig:toy_0.9_err_low_r1}
    \end{subfigure}
    \begin{subfigure}{\textwidth}
    \centering
        \includegraphics[width=0.95\textwidth]{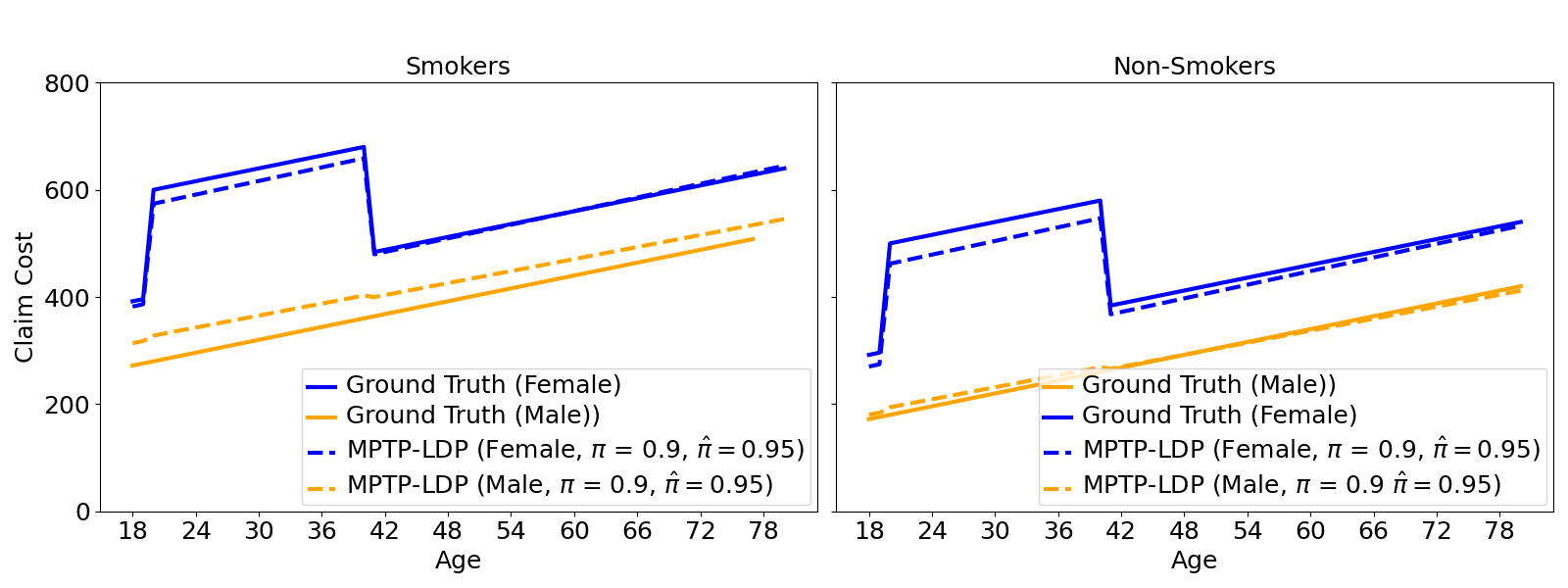} 
        \caption{Overestimated noise rate with $\pi = 0.9$}
        \label{fig:toy_0.9_err_high_r1}
    \end{subfigure}
        \caption{Group-specific expected claim costs under privatized sensitive attribute with unknown noise rates.}
    \label{fig:toy_exp3}
\end{figure}

To assess this sensitivity, we introduce controlled perturbations to the true value $\pi=0.9$, using $\hat{\pi} = \pi \pm 0.05$. These reflect both underestimation and overestimation cases. Figure~\ref{fig:toy_exp3} presents the resulting estimates of $\mu(X,D)$ under each case.

The results corroborate the theoretical insights from Theorem~\ref{theorem:GEB-NE} and its accompanying remarks. While both types of error degrade performance, underestimation of $\pi$ has a notably stronger impact. This asymmetry stems from the presence of the term $\frac{1}{\hat{\pi} - 1/|\cD|}$ in the generalization bound, which grows rapidly as $\hat{\pi}$ approaches $\frac{1}{|\cD|}$ from above. Thus, even a small underestimation in low-$\pi$ regimes can lead to substantial performance deterioration. These findings suggest a practical guideline: when the true noise rate is unknown, slightly conservative (i.e., overestimated) values of $\hat{\pi}$ may help mitigate risk and yield more stable performance.

% \begin{figure}[H]
%     \begin{subfigure}{\textwidth}
%     \centering
%         \includegraphics[width=0.95\textwidth]{Figures/toy_0.8_err_low_r1.png}
%         \caption{Underestimated noise rate with $\pi = 0.8$}
%         \label{fig:toy_0.8_err_low_r1}
%     \end{subfigure}
%     \begin{subfigure}{\textwidth}
%     \centering
%         \includegraphics[width=0.95\textwidth]{Figures/toy_0.8_err_high_r1.png} 
%         \caption{Overestimated noise rate with $\pi = 0.8$}
%         \label{fig:toy_0.8_err_high_r1}
%     \end{subfigure}
%         \caption{Group-specific expected claim costs under privatized sensitive attribute with unknown noise rates.}
%     \label{fig:toy_exp3*}
% \end{figure} 

% {\color{blue} TZ: Not really, but $\pi = 0.7$ looks very ugly, I attached $\pi = 0.8$ above, feel free to decide if we should go with $\pi = 0.8$}

% While a larger error in noise rate estimation inevitably introduces a larger bias in recovering the ground truth, we observe quite different behavior for underestimation and overestimation. As we already stated in the third remark following Theorem \ref{theorem:GEB-NE}, the generalization error bound is more adversely affected by the underestimation of $\pi$. This can be intuitively seen that when $\pi$ is close to $\frac{1}{|\cD|}$, if $\pi$ is underestimated, then the term $\frac{1}{\hat{\pi} - 1/|\cD|}$ can be arbitrarily large, making the generalization error bound loose.

\section{Real Data Analysis}
\label{sec:6}

We apply our proposed method to real-world datasets, evaluating its performance in both regression and classification tasks. Specifically, we use a health insurance dataset for a regression task (measured by mean squared error), and an automobile insurance dataset for a classification task (measured by cross-entropy loss). To conserve space, we present only the regression results here and defer the classification results to Appendix \ref{app:deferred_exp}.

The U.S. Health Insurance dataset contains 1,338 observations, 6 features, and 1 response variable (healthcare expenditure) \citep{10.5555/2588158}. We designate sex (with values ``Male'' and ``Female'') as the sensitive attribute $D$. The privatized version of this attribute is generated under varying privacy levels using a set of $\epsilon$ values in accordance with Definition \ref{def:pm}. 

We evaluate our method under two experimental settings: one where the noise rates are known, and another where the noise rates are unknown. In both cases, we adopt the following configuration:
\begin{itemize} \setlength\itemsep{-0.5em}
\item Hypothesis class $\cF$: linear models.
\item Representation $T(X) = \tilde{X}$, learned via a supervised feedforward neural network.
\item Three predefined noise levels, corresponding to $\pi = 0.9$, $0.8$, and $0.7$.
\end{itemize}

We split the data into training and test sets. For each setting, we estimate discrimination-free premium (Definition \ref{def:dfp}) using Algorithm \ref{alg:S-NE} (MPTP-LDP), where training is based on the privatized sensitive attribute $S$. As a benchmark, we compute the best-estimated premium (Definition \ref{def:bep}) using Algorithm \ref{alg:D} (MPTP) based on the true sensitive attribute $D$. We then compare the convergence behavior of the different pricing strategies by evaluating their test loss—the out-of-sample loss computed on the test set.

\subsection{Known Noise Rate}
\label{sec:6.1}

\begin{figure}[H]
    \centering
    \begin{subfigure}{0.49\textwidth}
        \centering
        \includegraphics[width=1\textwidth]{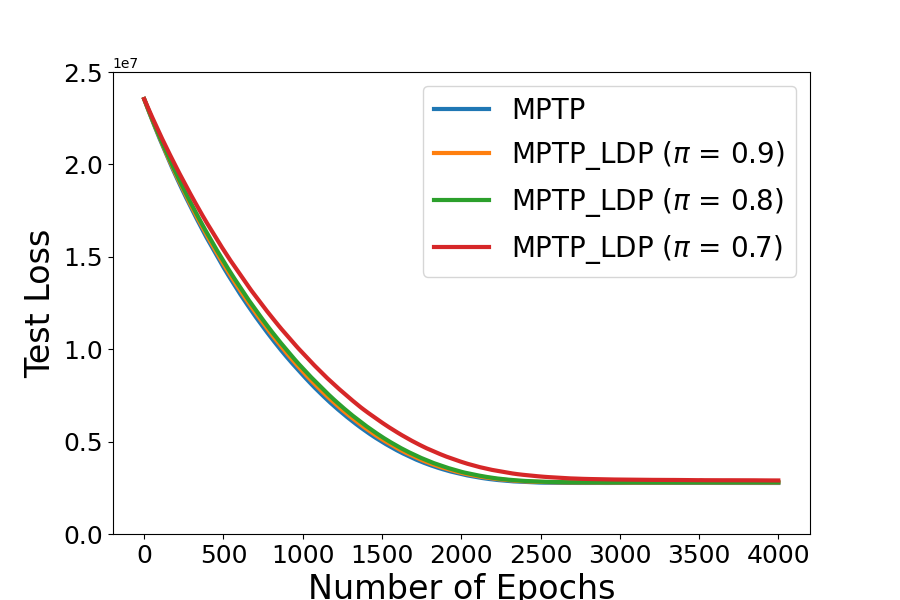}
        \caption{Original attributes $X$}
        \label{fig:health_1.0_mu_X_A_loss_r1}
    \end{subfigure}
    \begin{subfigure}{0.49\textwidth}
        \centering
        \includegraphics[width=1\textwidth]{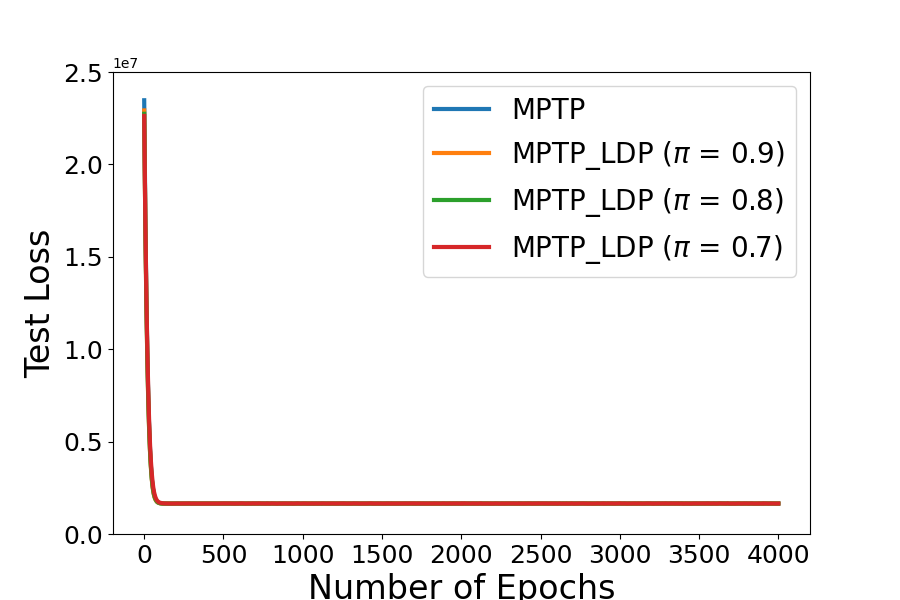} 
        \caption{Transformed attributes $\tilde{X}$}
        \label{fig:health_1.0_mu_TX_A_loss_r1}
    \end{subfigure}
    \caption{Comparison of test loss by noise rate for a given sample size, with known $\pi$.}
    \label{fig:health_1.0_loss_r1-LDP-a}
\end{figure}

Figure~\ref{fig:health_1.0_loss_r1-LDP-a} illustrates the impact of the noise rate on model performance under a fixed sample size, assuming the noise rate is known. The left panel presents results using the original attributes $X$, while the right panel uses the transformed attributes $\tilde{X}$ for premium estimation. As expected, higher noise rates (i.e., lower $\pi$) generally lead to increased error, consistent with our theoretical results in Theorem~\ref{theorem:GEB}.

Notably, the two panels exhibit a substantial difference in convergence behavior and robustness to noise. The model using $\tilde{X}$ demonstrates faster convergence and greater resilience to increasing noise levels. This improvement can be attributed to the fact that $\tilde{X}$ incorporates predictive information from the response $Y$ during training, effectively mitigating the impact of noise in the sensitive attribute.

%Figure \ref{fig:health_1.0_mu_X_A_loss_r1},\ref{fig:health_1.0_mu_TX_A_loss_r1} show how the noise rate affects loss approximation with a fixed sample size, where we observe a huge difference in terms of convergence rate and robustness against noise rate. This is expected since $\tilde{X}$ already incorporates some information from the response $Y$. Thus, the impact of noise perturbations is diminished, leading to increased robustness and faster convergence. Additionally, we note that a higher noise rate generally results in a larger error gap when the sample size remains fixed, which is consistent with our findings in Theorem \ref{theorem:GEB}. 

\begin{figure}[H]
    \begin{subfigure}{0.49\textwidth}
        \centering
        \includegraphics[width=\textwidth]{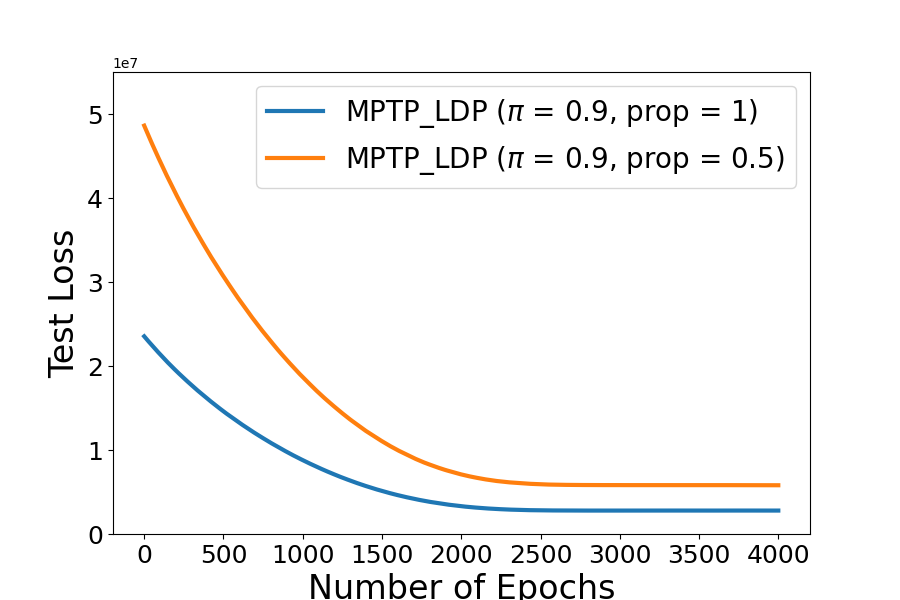}
         \caption{Original attributes $X$}
        \label{fig:health_p0.9_mu_X_A_loss_r1}
    \end{subfigure}
    \begin{subfigure}{0.49\textwidth}
        \centering
        \includegraphics[width=\textwidth]{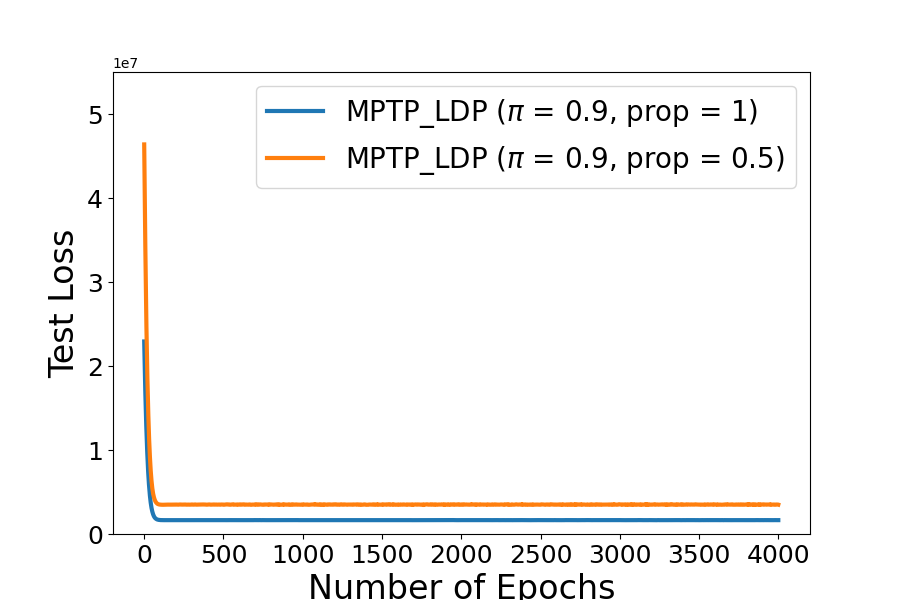} 
        \caption{Transformed attributes $\tilde{X}$}
        \label{fig:health_p0.9_mu_TX_A_loss_r1}
    \end{subfigure}
    \caption{Comparison of test loss by sample size for a given noise rate, with known $\pi$.}
    \label{fig:health_1.0_loss_r1-LDP-b}
\end{figure}

Figure \ref{fig:health_1.0_loss_r1-LDP-b} examines the effect of sample size under a fixed noise rate of $\pi = 0.9$. We compare performance using the full dataset versus a subsample containing half the observations. Once again, we observe a clear difference in convergence rates between models using $X$ and those using $\tilde{X}$. Moreover, for any fixed noise rate, increasing the sample size consistently reduces test loss, regardless of whether $X$ or $\tilde{X}$ is used—again aligning with Theorem \ref{theorem:GEB}.

From a practical perspective, when the sample size is sufficiently large, using a transformed representation can be especially advantageous in scenarios where: 1) the noise rate is high, 2) computing resources are limited, or 3) a tight generalization error is critical.

\subsection{Unknown Noise Rate}
\label{sec:6.2}

\begin{figure}[H]
    \centering
    \begin{subfigure}{1.2\textwidth}
        \includegraphics[width=0.3\textwidth]{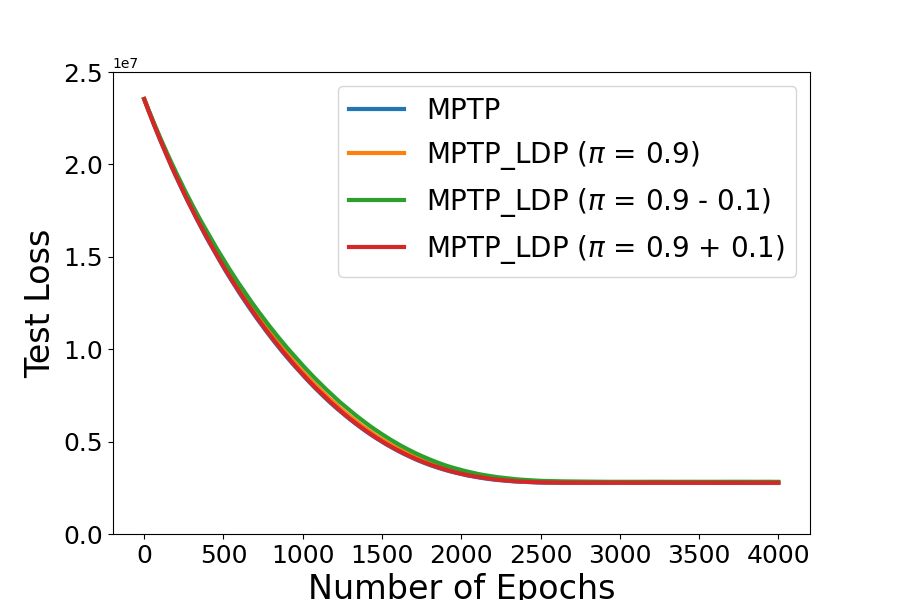}
        \includegraphics[width=0.3\textwidth]{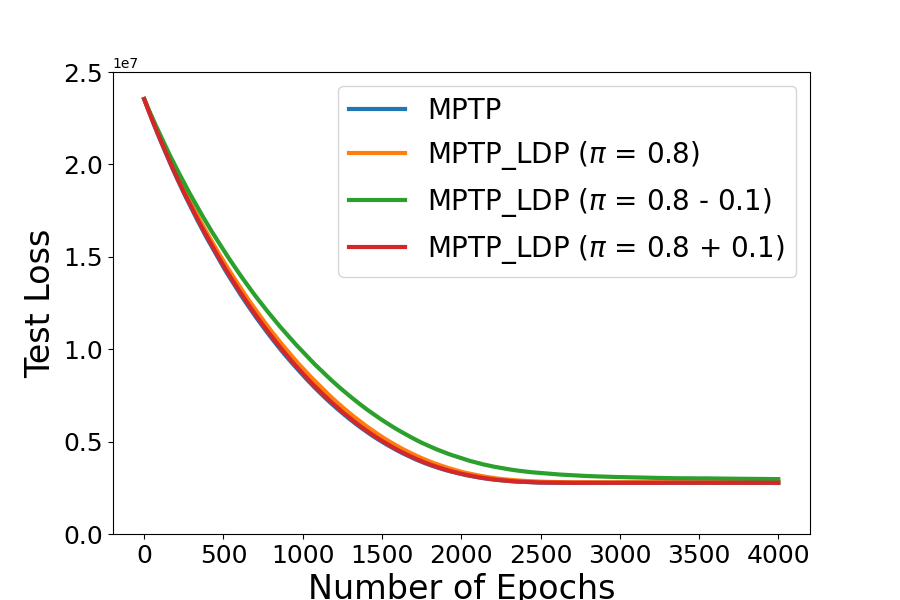} 
        \includegraphics[width=0.3\textwidth]
        {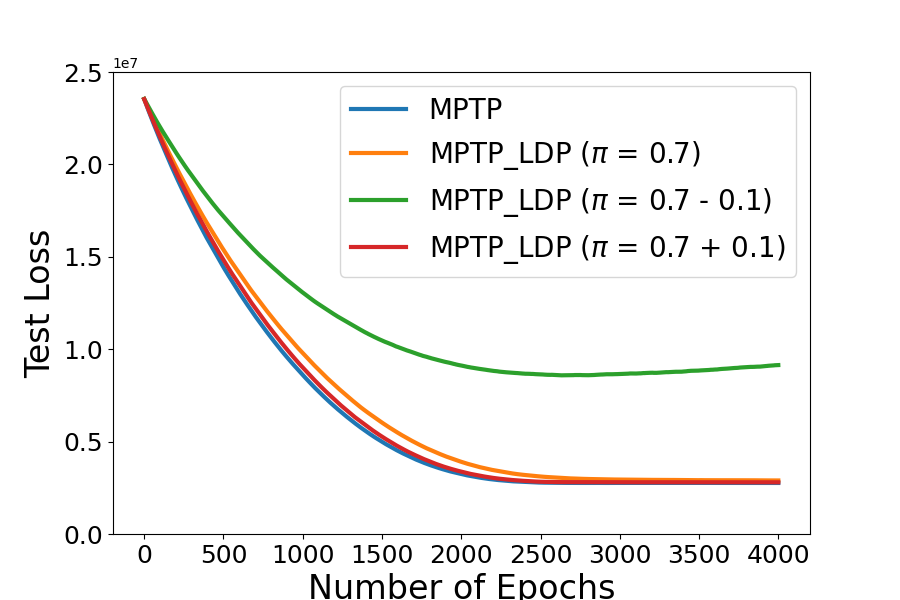} 
        \caption{Original attributes: $\pi=0.9, 0.8, 0.7$ from left to right}
    \end{subfigure}
    \begin{subfigure}{1.2\textwidth}
        \includegraphics[width=0.3\textwidth]{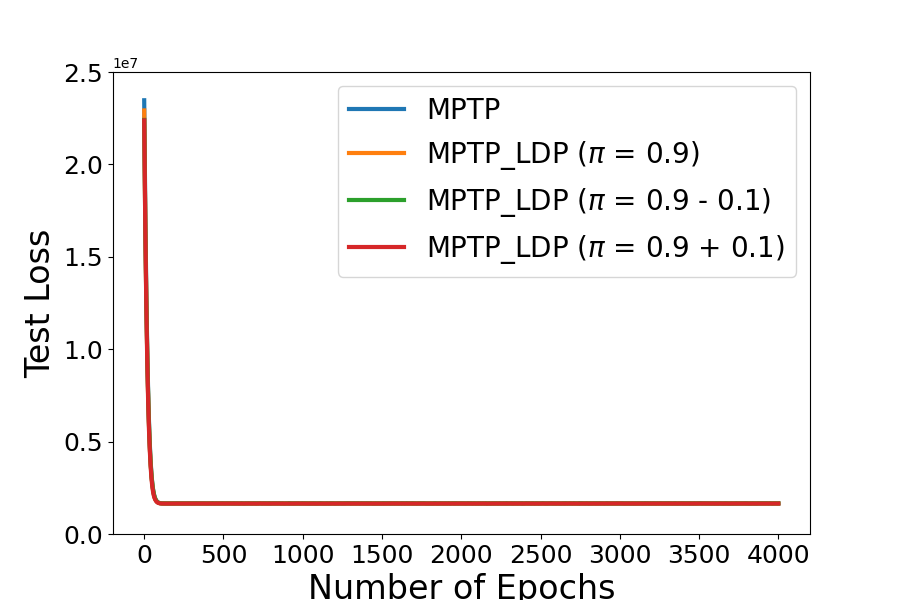}
        \includegraphics[width=0.3\textwidth]{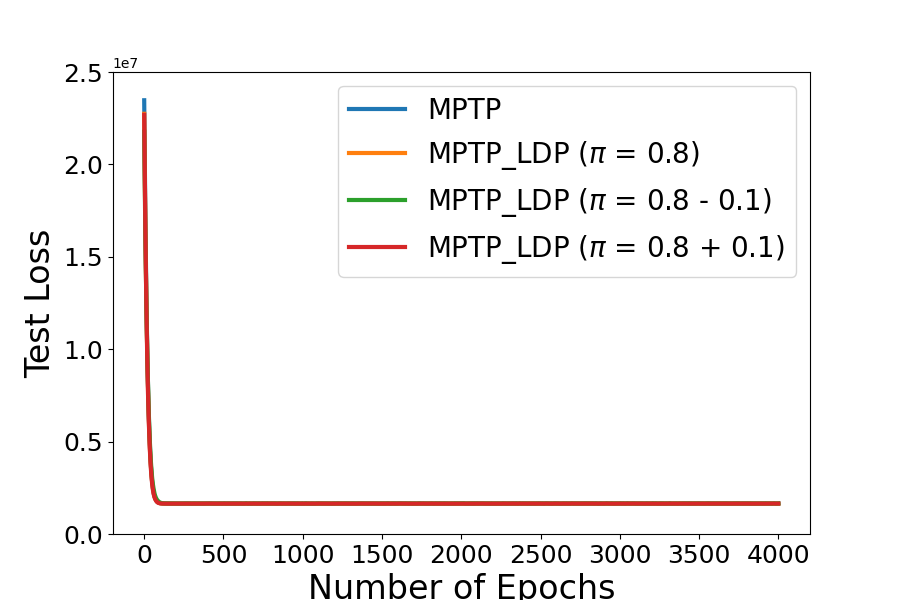} 
        \includegraphics[width=0.3\textwidth]
        {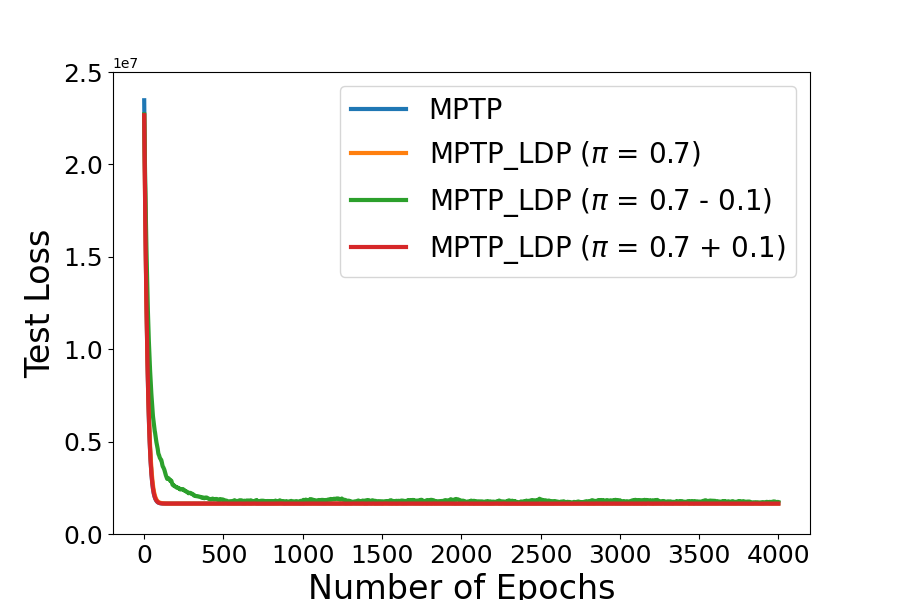} 
        \caption{Transformed attributes: $\pi=0.9, 0.8, 0.7$ from left to right}
    \end{subfigure}
    % \vspace{-8pt}
    \caption{Effect of estimation error in noise rate ($\pm 10\%$) on test loss.}
    % \vspace{-10pt}
    \label{fig:health_mu_TX_X_A_loss_ne_err_r1}
\end{figure}

We now consider the more practical setting where the noise rate used for privatization is unknown. To better understand the impact of estimation error on model performance, we begin with a controlled experiment in which the true noise rate $\pi$ is perturbed by a fixed margin. Specifically, we simulate underestimation and overestimation by adjusting $\pi$ by $\pm10\%$. We vary both the underlying value of $\pi$ and the input representation (original attributes $X$ versus transformed attributes $\tilde{X}$). The results are presented in Figure \ref{fig:health_mu_TX_X_A_loss_ne_err_r1}, where the pper panel corresponds to the orginal attributes and lower panel to the transformed attributes. 

The figure reveal several important insights. First, the effect of estimation error in $\pi$ is asymmetric: underestimating $\pi$ leads to worse performance than overestimating it. Second, this asymmetry becomes more pronounced as the true value of $\pi$ approaches $1/|D|$. For example, when $\pi=0.7$, the test loss under underestimation is substantially larger than under overestimation. This difference is much smaller when $\pi=0.9 ~\text{or}~ 0.8$. Third, using the transformed attributes $\tilde{X}$ generally improves convergence and reduces test error. This effect is especially evident when $\pi=0.7$, where the test loss with $\tilde{X}$ is noticeably closer to that observed at $\pi=0.9 ~\text{and}~ 0.8$, compared to the setting that uses the original attributes without transformation.

% We observe that models are less sensitive to estimation error when the true noise rate is low (i.e., when $\pi$ is large). ({\color{blue} why larger test loss}). For example, when $\pi = 0.8$, even a ±15\% estimation error does not significantly affect convergence. In contrast, estimation errors at lower values of $\pi$ can lead to larger test loss and slower convergence.

% A particularly notable finding is the asymmetric impact of under- and over-estimation. Underestimating $\pi$ (i.e., assuming the noise is lower than it actually is) results in significantly worse performance than overestimating it. This aligns with the theoretical insight from Theorem \ref{theorem:GEB-NE}, which suggests that introducing a slight upward bias to $\hat{\pi}$ may improve stability. ({\color{blue} why})

% We also find that although transformation via $\tilde{X}$ facilitates convergence in general, its benefit diminishes when the noise rate is very high. In such cases, the quality of the transformed representation itself may be compromised by the noise, reducing its advantage.

Next, we conduct a more realistic experiment where the noise rate $\pi$ is estimated directly from data using the method described in Lemma \ref{lemma:NE}. Recall that we randomly and evenly divide the full dataset into $n_1$ groups and compute an estimate $\hat{\pi}_k$ for each group $k = 1, \ldots, n_1$. We then aggregate these to obtain a final estimate $\hat{\pi} = \frac{1}{n_1} \sum_{k = 1}^{n_1} \hat{\pi}_k$.

\begin{figure}[H]
    \centering
    \begin{subfigure}{0.49\textwidth}
        \centering
        \includegraphics[width=\textwidth]{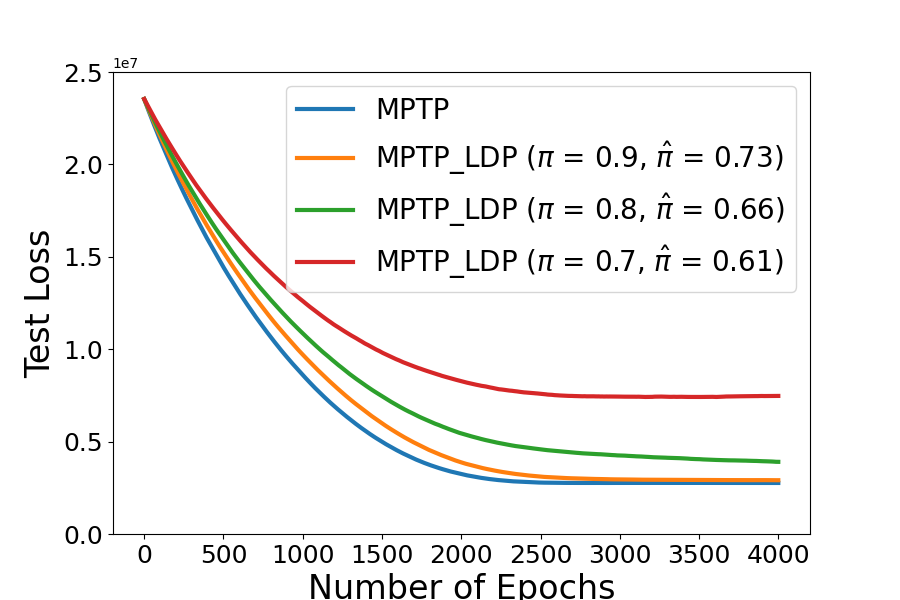} 
        \caption{Original attributes $X$}
        \label{fig:health_1.0_mu_X_A_loss_ne_n14_r1}
    \end{subfigure}
    \begin{subfigure}{0.49\textwidth}
        \centering
        \includegraphics[width=\textwidth]{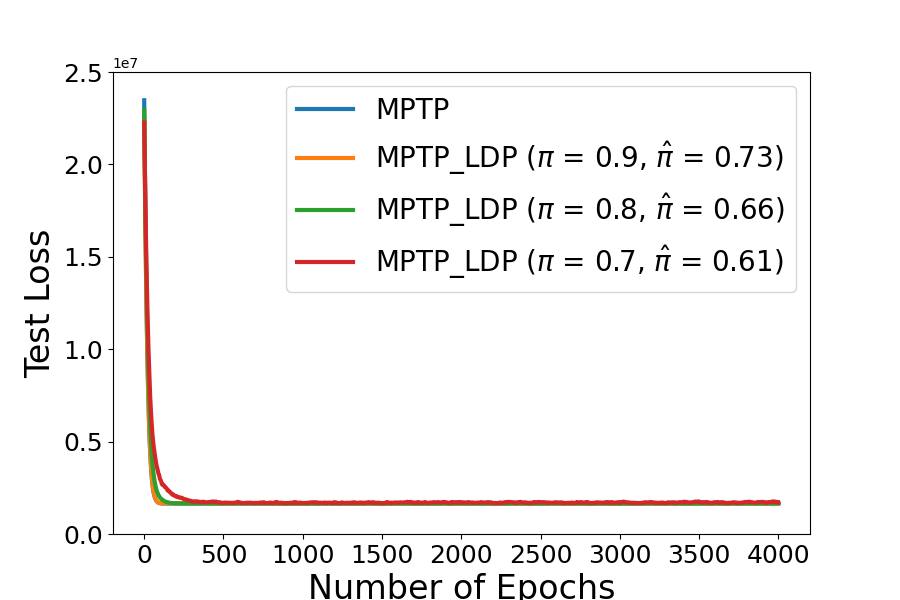} 
        \caption{Transformed attributes $\tilde{X}$}
        \label{fig:health_1.0_mu_TX_A_loss_ne_n14_r1}
    \end{subfigure}
    % \vspace{-8pt}
    \caption{Comparison of test loss by noise rate for a given sample size, with estimated $\pi$.}
    % \vspace{-10pt}
    \label{fig:health_1.0_mu_X_TX_A_loss_ne_r1}
\end{figure}

\begin{figure}[H]
    \centering
    \begin{subfigure}{0.49\textwidth}
        \centering
        \includegraphics[width=\textwidth]{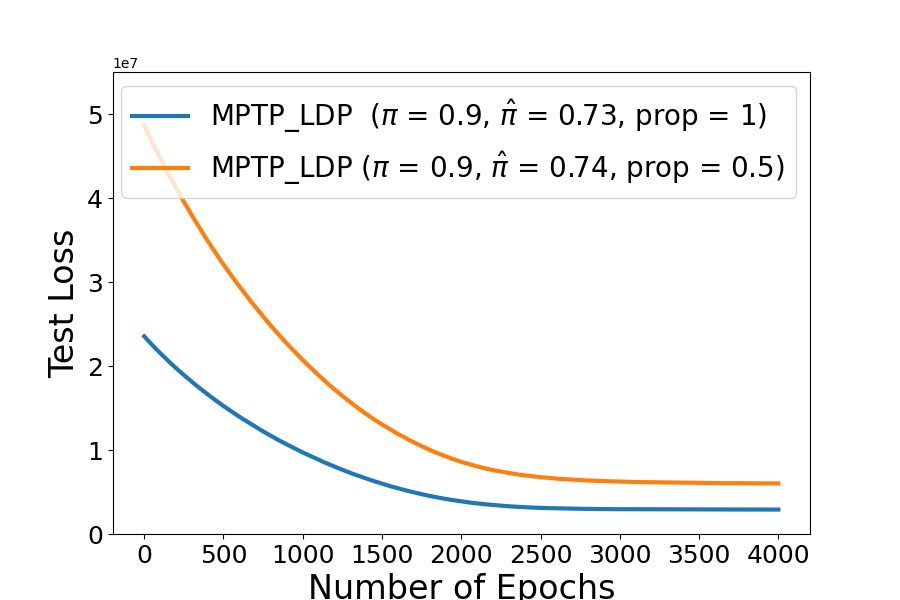} 
        \caption{Original attributes $X$}
        \label{fig:health_subset_p0.35_mu_X_A_loss_ne_n14_r1}
    \end{subfigure}
    \begin{subfigure}{0.49\textwidth}
        \centering
        \includegraphics[width=\textwidth]{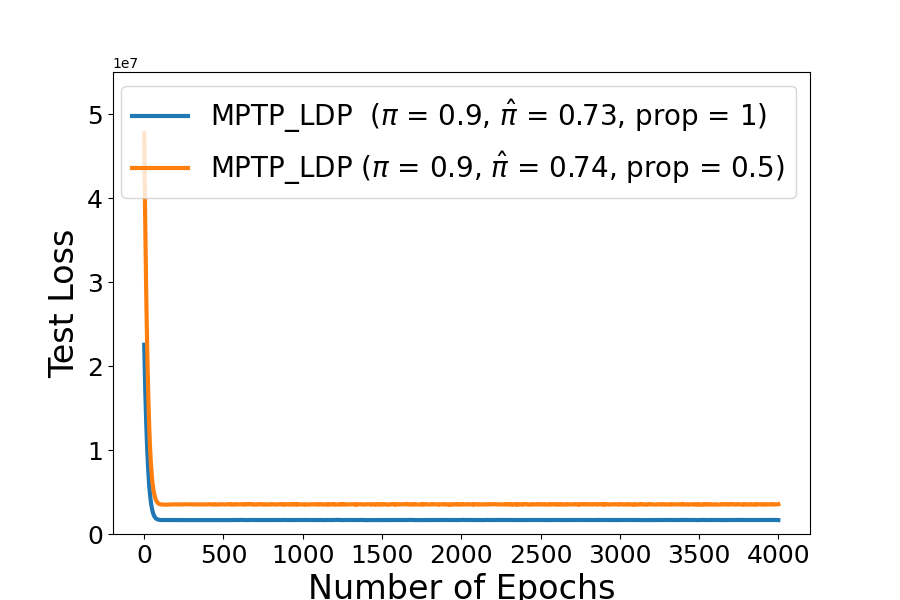} 
        \caption{Transformed attributes $\tilde{X}$}
        \label{fig:health_subset_p0.35_mu_TX_A_loss_ne_n14_r1}
    \end{subfigure}
    % \vspace{-8pt}
    \caption{Comparison of test loss by sample size for a given noise rate, with estimated $\pi$.}
    % \vspace{-10pt}
    \label{fig:health_subset_mu_X_TX_A_loss_ne_r1}
\end{figure}

As in Section~\ref{sec:6.1}, we examine how performance varies with both the true noise rate and the sample size. The corresponding results are presented in Figures~\ref{fig:health_1.0_mu_X_TX_A_loss_ne_r1} and~\ref{fig:health_subset_mu_X_TX_A_loss_ne_r1}, respectively. In both cases, we set $n_1=4$ for estimating $\pi$. and we compare performance using the original attributes $X$ and the transformed attributes $\tilde{X}$. The observed patterns are consistent with our theoretical predictions.

Figure~\ref{fig:health_1.0_mu_X_TX_A_loss_ne_r1} explores three values of $\pi$, and in all scenarios, the estimated $\pi$ is lower than the true value. The left panel clearly shows that the adverse effect of underestimation is more severe when $\pi$ is closer to $1/|D|$. The right panel reinforces the point that an informative transformation, such as $\tilde{X}$, can mitigate the impact of estimation error. Figure~\ref{fig:health_subset_mu_X_TX_A_loss_ne_r1} focuses on a single noise level corresponding to $\pi=0.9$, and compares two sample sizes: full and half. The results reflect a combined effect of sample size and estimation accuracy. In this case, the estimates of $\pi$ from both the full and half samples are similar, and both underestimate the true value. As a result, the effect of sample size appears to be the dominant factor. While the transformation $\tilde{X}$ does improve convergence, it does not fully compensate for the performance loss due to the reduced sample size in this particular setting.

%The results reflect a compound effect of sample size and estimation accuracy. In this case, the full sample leads to underestimation of $\pi$, while the half sample results in a slight overestimation or near-accurate estimate. As a result, the effect of sample size appears to dominate. We also observe that while the transformation $\tilde{X}$ improves convergence, it does not fully offset the dominant effect of reduced sample size in this particular setting.

% \vspace{-10pt}

\begin{figure}[H]
        \centering
    \begin{subfigure}{1.2\textwidth}
        \includegraphics[width=0.3\textwidth]{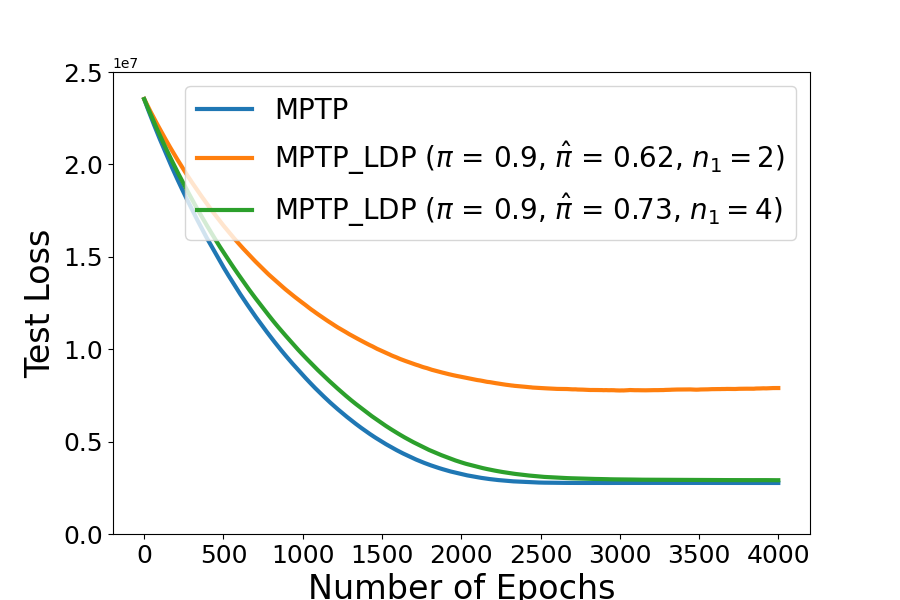}
        \includegraphics[width=0.3\textwidth]{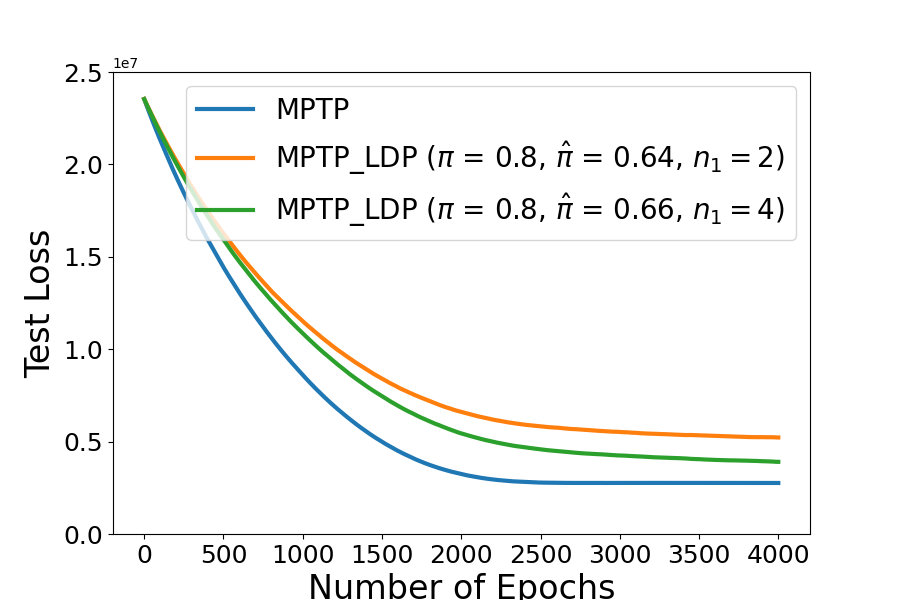} 
        \includegraphics[width=0.3\textwidth]{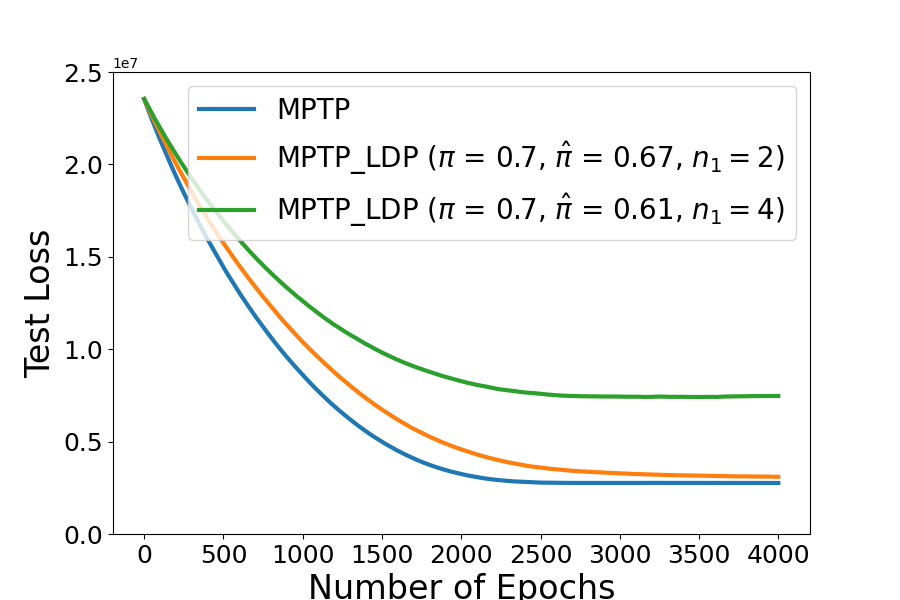} 
        \caption{Original attributes: $\pi=0.9, 0.8, 0.7$ from left to right}
    \end{subfigure}
    \begin{subfigure}{1.2\textwidth}
        \includegraphics[width=0.3\textwidth]{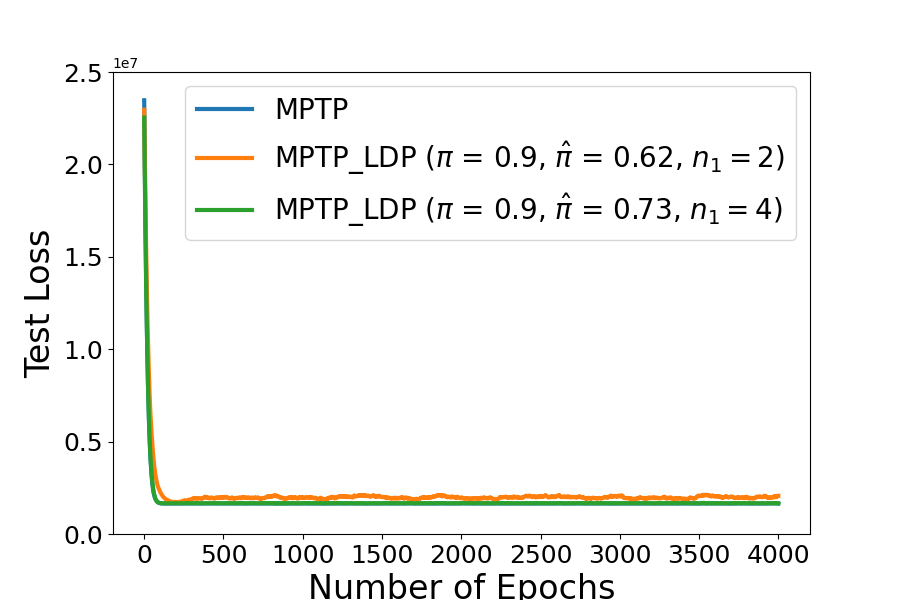}
        \includegraphics[width=0.3\textwidth]{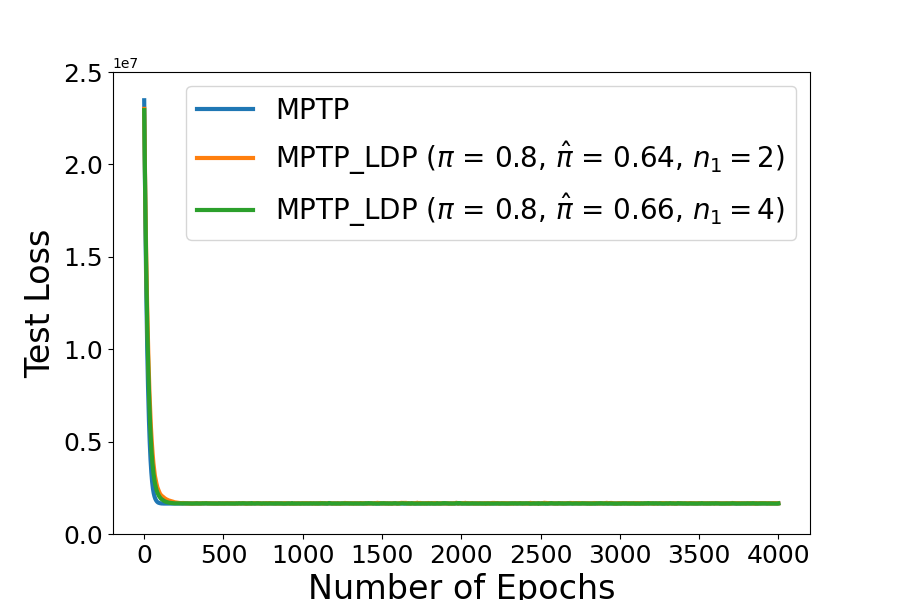} 
        \includegraphics[width=0.3\textwidth]{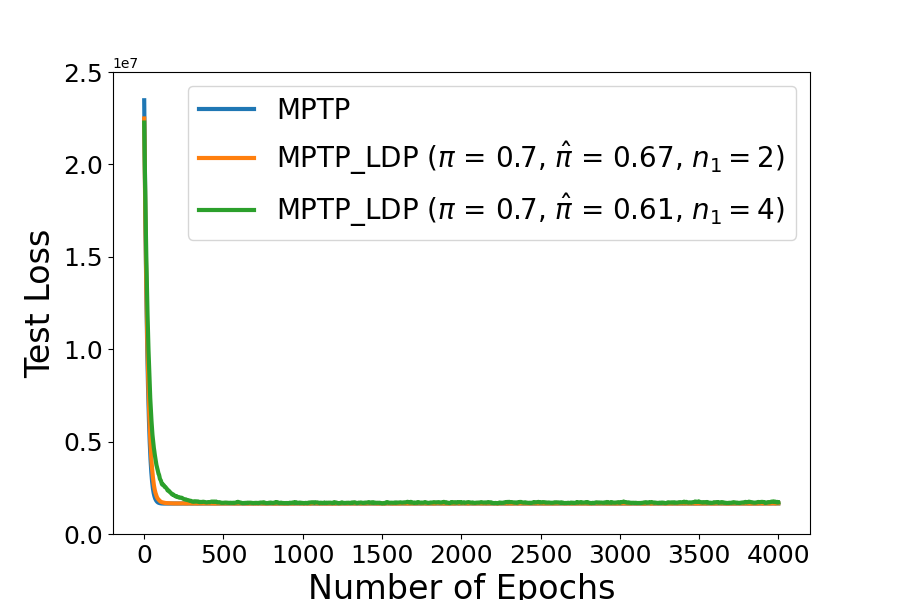} 
        \caption{Transformed attributes: $\pi=0.9, 0.8, 0.7$ from left to right}
    \end{subfigure}
    % \vspace{-8pt}
    \caption{Effect of group size ($n_1= 2 ~\text{or}~4$) on test loss.}
    \label{fig:health_1.0_mu_X_TX_A_loss_ne_n1_r1}
\end{figure}

Last, we examine the effect of the number of groups $n_1$ on model performance. Specifically, we consider three noise levels, $\pi=0.9, 0.8, 0.7$, and two types of input representations: original and transformed attributes. For each of the six combinations, we compare model performance under $n_1=2$ and $n_1=4$. The results are summarized in Figure~\ref{fig:health_1.0_mu_X_TX_A_loss_ne_n1_r1}, where the upper and lower panels correspond to the original and transformed attributes, respectively.

Several important takeaways emerge from this figure. In the upper panel, we observe that for lower noise levels ($\pi=0.9~\text{and}~0.8$), having more groups ($n_1=4$) improves model performance, as evidenced by reduced test loss. However, when $\pi=0.7$, this relationship reverses: a larger group size leads to worse performance. This pattern aligns with our theoretical insights. When the effective sample size is sufficiently large, a larger $n_1$ enables more accurate and robust estimation of $\pi$, which in turn improves predictive performance. However, when the effective sample size is small, as is the case with a high noise rate like $\pi=0.7$, a larger $n_1$ can reduce estimation stability and degrade model performance. In the lower panel, these patterns are less pronounced. This suggests that the informative transformation (i.e. using $\tilde{X}$) can partially offset the detrimental effects of reduced effective sample size due to high noise rates.

% \vspace{-10pt}

\section{Conclusion}
% \vspace{-5pt}
\label{sec:7}

This paper studies discrimination-free insurance pricing in settings where sensitive attributes are not directly observable. While recent actuarial research has proposed the discrimination-free premium as a principled way to eliminate both direct and indirect discrimination, implementing this concept in practice is challenging because it typically requires access to sensitive attributes. In many real-world environments, however, such attributes are unavailable due to regulatory restrictions or privacy constraints, or they are only available in privatized form.

To address this challenge, we develop a statistical framework for estimating discrimination-free premiums when sensitive attributes are observed only through a privacy mechanism. Our approach operates within a multi-party data setting in which insurers observe non-sensitive attributes and outcomes while a trusted third party holds privatized sensitive attributes. Within this framework, we propose estimation procedures that allow discrimination-free premiums to be constructed without requiring direct access to the true sensitive attributes. We analyze two practically relevant scenarios: when the privacy mechanism is known and when its noise level is unknown. For both cases, we establish theoretical guarantees for the proposed estimators and demonstrate their empirical performance through simulation studies and real data applications.

The increasing use of machine learning in insurance pricing raises important questions about fairness, transparency, and privacy. By integrating the discrimination-free premium framework with statistical methods for learning under noisy or privatized data, this paper provides a practical approach to fair insurance pricing under realistic data constraints. Future research may extend this framework to more complex data environments, such as settings with multiple sensitive attributes, richer privacy mechanisms, or dynamic pricing models.

\bibliographystyle{chicago}
\bibliography{iclr2025_conference}

\newpage
%\appendix
\section*{Appendix}
\renewcommand{\theequation}{A.\arabic{equation}}
\renewcommand{\thetable}{A.\arabic{table}}
\renewcommand{\thefigure}{A.\arabic{figure}}
\renewcommand{\thesection}{A\alph{section}}
\setcounter{figure}{0}
\setcounter{table}{0}
\setcounter{equation}{0}
\setcounter{section}{0}

\subsection{Proofs}
\label{app:deferred_proofs}

\subsubsection{Lemma \ref{lemma:Risk-LDP}}
\label{proof:lemmaRisk-LDP}

% \textbf{Lemma \ref{lemma:Risk-LDP}} Given the privacy parameter $\epsilon$, minimizing the following risk (Risk-LDP) Eq. (\ref{eq:Risk-LDP}) under $\epsilon$-LDP w.r.t. privatized sensitive attributes $S$ is equivalent of minimizing Eq. (\ref{eq:TTPpopLoss}) w.r.t. true sensitive attributes $D$ at the population level:
% \begin{equation}
%     \cR^{LDP}(f_1, \ldots, f_{|\cD|}) = \sum_{k=1}^{|\cD|} \sum_{j=1}^{|\cD|} \left( \boldsymbol{\Pi}_{kj}^{-1} \Ex_{Y, \tilde{X} \mid S = j} \left[ L \bigr(f_k(\tilde{X}),Y \bigr) \right] \cdot \sum_{l = 1}^{|\cD|} \boldsymbol{T}_{kl}^{-1} \bbP(S = l) \right),
% \end{equation}
% where $\boldsymbol{\Pi}^{-1}$ and $\boldsymbol{T}^{-1}$ are $|\cD| \times |\cD|$ row-stochastic matrics.

\begin{proof}

\textbf{Step 1:}

Since the $\epsilon$-LDP randomization mechanism operates independently of $X$ and $Y$, the distribution of the privatized output $S$ depends solely on the privacy parameter $\epsilon$ and the distribution of $D$. In other words, once $\epsilon$ and the distribution of $D$ are specified, the distribution of $S$ is fully determined.

\textbf{Step 2:} Recover distributions w.r.t. $D$. 

This step is inspired by Proposition 1 in \cite{10.5555/3524938.3525593}. Let $\cE_1$ and $\cE_2$ be two events defined over the joint space of $(\tilde{X}, Y, \hat{Y})$. Consider the following probability:
\[
\begin{aligned}
    & \bbP(\cE_1, \cE_2 \mid S=d) \\ 
    =& \sum_{d' \in D} \bbP(\cE_1, \cE_2 \mid S=d, D=d') \bbP(D=d' \mid S=d)\\
    =& \sum_{d' \in D} \bbP(\cE_1, \cE_2 \mid D=d') \bbP(D=d' \mid S=d)\\
    =& \sum_{d' \in D} \bbP(\cE_1, \cE_2 \mid D=d') \frac{ \bbP(S=d \mid D=d') \bbP(D=d')}{\sum_{d'' \in D} \bbP(S=d \mid D=d'') \bbP(D=d'')}\\
    =& P(\cE_1, \cE_2 \mid D=d)\frac{\pi \bbP(D = d)}{\pi \bbP(D = d) + \underset{d''\setminus d}{\sum}\bar{\pi}\bbP(D = d^{''})}+\sum_{d'\setminus d}P(\cE_1, \cE_2 \mid D=d')\frac{\bar{\pi}\bbP(D = d')}{\pi \bbP(D = d) + \underset{d''\setminus d}{\sum}\bar{\pi}\bbP(D = d^{''})}.
\end{aligned}
\]

Set $\cE_1 = Y, \cE_2 = \tilde{X}$. This yields: 
\[
\begin{aligned}
    & \bbP(Y, \tilde{X} \mid S=d) \\ 
    =& \sum_{d' \in D} \bbP(Y, \tilde{X} \mid S=d, D=d') \bbP(D=d' \mid S=d)\\
    =& \sum_{d' \in D} \bbP(Y, \tilde{X} \mid D=d') \bbP(D=d' \mid S=d)\\
    =& \sum_{d' \in D} \bbP(Y, \tilde{X} \mid D=d') \frac{ \bbP(S=d \mid D=d') \bbP(D=d')}{\sum_{d'' \in D} \bbP(S=d \mid D=d'') \bbP(D=d'')}\\
    =& P(Y, \tilde{X} \mid D=d)\frac{\pi \bbP(D = d)}{\pi \bbP(D = d) + \underset{d''\setminus d}{\sum}\bar{\pi}\bbP(D = d^{''})}+\sum_{d'\setminus d}P(Y, \tilde{X} \mid D=d')\frac{\bar{\pi}\bbP(D = d')}{\pi \bbP(D = d) + \underset{d''\setminus d}{\sum}\bar{\pi}\bbP(D = d^{''})}.
\end{aligned}
\]

Let $p_d = \bbP(D = d)$ denote the marginal distribution of $D$. Define $\boldsymbol{\Pi}$ as a $|\cD| \times |\cD|$ matrix whose entries are given by:
\[
\begin{cases}
\boldsymbol{\Pi}_{ii} = \frac{\pi p_i}{\pi p_i + \underset{d''\setminus i}{\sum}\bar{\pi}p_{d''}}, \text{for $i \in D$}\\
\boldsymbol{\Pi}_{ij} = \frac{\bar{\pi}p_j}{\pi p_i + \underset{d''\setminus i}{\sum}\bar{\pi}p_{d''}}, \text{for $i,j \in D$ s.t.,$i \ne j$}
\end{cases}.
\]
Using this, we obtain the following system of linear equations: 
\[
\begin{bmatrix}
\bbP(Y,\tilde{X} \mid S=1)\\
.\\
.\\
.\\
\bbP(Y,\tilde{X} \mid S=|\cD|)
\end{bmatrix}
= \boldsymbol{\Pi}
\begin{bmatrix}
\bbP(Y,\tilde{X} \mid D=1)\\
.\\
.\\
.\\
\bbP(Y,\tilde{X} \mid D=|\cD|)
\end{bmatrix}.
\]
Let us denote the linear relationship as $\boldsymbol{s}_1 = \boldsymbol{\Pi} \boldsymbol{d}_1$, wherewhere $\boldsymbol{s_1} = \bbP(Y,\tilde{X} \mid S)$, $\boldsymbol{d_1} = \bbP(Y,\tilde{X} \mid D)$.

Since $\boldsymbol{\Pi}$ is a row-stochastic and invertible matrix, we can compute $\boldsymbol{\Pi}^{-1}$ explicitly. The entries of $\boldsymbol{\Pi}^{-1}$ take the following form:
\[
\begin{cases}
\boldsymbol{\Pi}_{ii}^{-1} = \frac{\pi + |\cD| - 2}{|\cD|\pi - 1}\frac{\pi p_i + \sum_{d''\setminus i} \bar{\pi}p_{d''}}{p_i}, \text{for $i \in D$}\\
\boldsymbol{\Pi}_{ij}^{-1} = \frac{\pi - 1}{|\cD|\pi - 1}\frac{\bar{\pi} p_i + \sum_{d''\setminus i} \pi p_{d''}}{p_i}, \text{for $i,j \in D$ s.t.,$i \ne j$}
\end{cases},
\]
Multiplying both sides of the equation $\boldsymbol{s}_1 = \boldsymbol{\Pi} \boldsymbol{d}_1$ by $\boldsymbol{\Pi}^{-1}$ yields:
\[
\begin{aligned}
    \bbP(Y,\tilde{X} \mid D=k) &= \sum_{j=1}^{|\cD|}\boldsymbol{\Pi}_{kj}^{-1} \bbP(Y,\tilde{X} \mid S=j) \\
    &= \boldsymbol{\Pi}_{k\cdot}^{-1} \bbP(Y,\tilde{X} \mid S)
\end{aligned}
\]
where $\boldsymbol{\Pi}^{-1}_{k \cdot}$ denotes the $k^{\text{th}}$ row of $\boldsymbol{\Pi}^{-1}$.

However, to fully recover the population distribution $\bbP(Y, \tilde{X} \mid D)$, one additional component remains to be estimated: the marginal distribution $\bbP(D = d)$. To estimate $\bbP(D = d)$, we again exploit the structure of the randomization mechanism. Specifically, we express $\bbP(S = d)$ in terms of the conditional distribution of $S$ given $D$:
% However, there is still one component that we do need to estimate in order to recover the population distribution of $\bbP(Y, \tilde{X} \mid D)$. We need to further estimate $\bbP(D = d)$. Using the same technique, to estimate $\bbP(D=d)$, first write $P(S=d)$ in terms of the conditional probability of $S$ given $D$ as:
\[
\begin{aligned}
    \bbP(S=d) &= \sum_{d'\in D} \bbP(S=d \mid D=d') \bbP(D=d') \\
       &= \bbP(S=d \mid D=d) \bbP(D=d) + \sum_{d'\setminus d} \bbP(S=d \mid D=d') \bbP(D=d') \\
       &= \pi p_d + \sum_{d'\setminus d}\bar{\pi}p_{d'}.
\end{aligned}
\]
We now specify the structure of the matrix $\boldsymbol{T}$ used in the system of linear equations. Let $\boldsymbol{T}$ be a $|\cD| \times |\cD|$ matrix with entries defined as:
\[
\begin{cases}
\boldsymbol{T}_{ii}=\pi, \text{for $i \in D$}\\
\boldsymbol{T}_{ij}=\bar{\pi}, \text{for $i,j \in D$ s.t.,$i \ne j$}
\end{cases}.
\]
With this construction, the system of equations becomes:
\[
\begin{bmatrix}
\bbP(S=1)\\
.\\
.\\
.\\
\bbP(S=|\cD|)
\end{bmatrix}
= \boldsymbol{T}
\begin{bmatrix}
\bbP(D=1)\\
.\\
.\\
.\\
\bbP(D=|\cD|)
\end{bmatrix},
\]
which we denote compactly as $\boldsymbol{s}_2 = \boldsymbol{T} \boldsymbol{d}_2$, where $\boldsymbol{s}_2 = \bbP(S)$ and $\boldsymbol{d}_2 = \bbP(D)$.

By the same argument as before, the matrix $\boldsymbol{T}$ is invertible. It can be verified that the entries of $\boldsymbol{T}^{-1}$ take the following form:
\[
\begin{cases}
\boldsymbol{T}_{ii}^{-1} = \frac{\pi + |\cD| - 2}{|\cD|\pi - 1}, \text{for $i \in D$}\\
\boldsymbol{T}_{ij}^{-1} = \frac{\pi - 1}{|\cD|\pi - 1}, \text{for $i,j \in D$ s.t.,$i \ne j$}
\end{cases}.
\]
Multiplying both sides of the equation by $\boldsymbol{T}^{-1}$ yields the closed-form expression for recovering the marginal distribution of $D$:
\[
\begin{aligned}
    \bbP(D=k) &= \sum_{j=1}^{|\cD|} \boldsymbol{T}_{kj} \bbP(S = j) \\
    &= \boldsymbol{T}_{k\cdot}^{-1} \bbP(S).
\end{aligned}
\]
where $\boldsymbol{T}^{-1}_{k \cdot}$ denotes the $k^{\text{th}}$ row of $\boldsymbol{T}^{-1}$.

\textbf{Step 3:} Recover the loss w.r.t. D.

At the population level, we have shown that the conditional distribution of $(Y, \tilde{X})$ given $D = k$ can be recovered as:
\[
\bbP(Y, \tilde{X} \mid D = k) = \boldsymbol{\Pi}_{k\cdot}^{-1} \bbP(Y,\tilde{X} \mid S),
\]
where $\bbP(D = k) = \boldsymbol{T}_{k \cdot}^{-1} \bbP(S)$ is used in the calculation of $\boldsymbol{\Pi}_{k \cdot}^{-1}$.

Substituting these expressions into the population-level loss, we recover the population analogue of Eq.~\eqref{eq:TTPpopLoss}:
\[
\begin{aligned}
    &\sum_{k=1}^{|\cD|} \left( \Ex_{Y,\tilde{X} \mid D = k}\left[ L \bigr( Y, f_k(\tilde{X}) \bigr) \right] \cdot \bbP(D = k) \right) \\ 
    =& \sum_{k=1}^{|\cD|} \left( \int_Y\int_{\tilde{X}} \bbP(Y,\tilde{X} \mid D = k) L \left( Y, f_k(\tilde{X}) \right) d\tilde{X}dY  \cdot \bbP(D = k) \right) \\
    =& \sum_{k=1}^{|\cD|} \left( \left[ \int_Y\int_{\tilde{X}} \sum_{j=1}^{|\cD|}\boldsymbol{\Pi}_{kj}^{-1} \bbP(Y,\tilde{X} \mid S=j) L \bigr( Y, f_k(\tilde{X}) \bigr) d\tilde{X}dY \right] \cdot \sum_{l = 1}^{|\cD|} \boldsymbol{T}_{kl}^{-1} \bbP(S = l) \right) \\
    =& \sum_{k=1}^{|\cD|} \left( \left[ \sum_{j=1}^{|\cD|} \int_Y\int_{\tilde{X}} \boldsymbol{\Pi}_{kj}^{-1} \bbP(Y, \tilde{X} \mid S = j) L \bigr( Y, f_k(\tilde{X}) \bigr) d\tilde{X}dY \right] \cdot \sum_{l = 1}^{|\cD|} \boldsymbol{T}_{kl}^{-1} \bbP(S = l) \right)\\
    =& \sum_{k=1}^{|\cD|} \sum_{j=1}^{|\cD|} \left( \boldsymbol{\Pi}_{kj}^{-1} \Ex_{Y, \tilde{X} \mid S = j} \left[ L \bigr(f_k(\tilde{X}),Y \bigr) \right] \cdot \sum_{l = 1}^{|\cD|} \boldsymbol{T}_{kl}^{-1} \bbP(S = l) \right).
\end{aligned}
\]

Therefore, we conclude that minimizing the original loss with respect to the true (but unobserved) distribution over $D$ is equivalent to minimizing:
\[
(f_{1^*}, \ldots, f_{|\cD|^*}) \leftarrow \underset{f_1, \ldots, f_{|\cD|}}{\arg \min} \sum_{k=1}^{|\cD|} \sum_{j=1}^{|\cD|} \left( \boldsymbol{\Pi}_{kj}^{-1} \Ex_{Y, \tilde{X} \mid S = j} \left[ L \bigr(f_k(\tilde{X}),Y \bigr) \right] \cdot \sum_{l = 1}^{|\cD|} \boldsymbol{T}_{kl}^{-1} \bbP(S = l) \right)
\]

% But, it shall be noted that empirically, although the entries of $\hat{\bbP}(Y, X \mid D = k), k = 1, \ldots, |\cD|$ sum to $1$, some entries might be in fact negative. Thus, we need to project the derived empirical estimator onto a simplex of the corresponding dimension for it to be valid.

% We now define the orthogonal projection of $x$ onto a simplex as:
% \[
% \text{proj}_{\bm{\mathit{\Delta}}}(\bm{\mathrm{x}}):= \arg\min_{\bm{\mathrm{y}}}\frac{1}{2}\|\mathrm{y}-\mathrm{x}\|_2^2, \  \text{s.t.}\medspace \bm{\mathrm{y}}^T\bm{1} = 1, \bm{\mathrm{y}}>0,
% \]
% which can be solved optimally in a non-iterative manner in time $\cO(|\cD|log|\cD|)$.

% Therefore, a well-defined empirical estimator $\hat{\bbP}(Y, X \mid D = k)$ is obtained through its orthogonal projection to the $|\cD| - 1$ simplex:
% \[
% \begin{aligned}
% \hat{\bbP}(Y, X \mid D = k) &= \text{proj}_{\bm{\mathit{\Delta}}} \Bigr( \sum_{j=1}^{|\cD|} \boldsymbol{\Pi}_{kj}^{-1} \hat{\bbP}(Y, X | S = j) \Bigr) \\
% &= \text{proj}_{\bm{\mathit{\Delta}}} \Bigr( \boldsymbol{\Pi}_{k \cdot}^{-1} \hat{\bbP}(Y, X \mid S) \Bigr)
% \end{aligned}
% \]
% where $\hat{\bbP}(Y, X \mid S)$ is the empirical distribution from data.

This completes the proof.
\end{proof}

%####################################################################################################
%####################################################################################################
\newpage
\subsubsection{\textbf{Theorem} \ref{theorem:GEB} }
\label{proof:theoremGEB}

% \textbf{Theorem \ref{theorem:GEB}}  For any $\delta \in (0,\frac{1}{2})$, $C_1 = \frac{\pi + |\cD| - 2}{|\cD|\pi - 1}$, denote $VC(\cF)$ as the VC-dimension of the hypothesis class $\cF$, and $K$ be some constant that depends on $VC(\cF)$. Let $f = \{f_k\}_{k=1}^{|\cD|}$ where $f_k \in \cF$ and let $L: Y \times Y \to \bbR_+$ be a loss function bounded by some constant $M$. Denote $k^* \leftarrow \underset{k}{\arg\max} |\hat{\cR}^{LDP}(f_k)\hat{\bbP}(D = k) - \cR^{LDP}(f_k)\bbP(D = k)|$, if $n \ge \frac{8 \ln{(\frac{|\cD|}{\delta})}}{\min_k \bbP(S = k)}$ \, then with probability $1 - 2\delta$:
% \[
% \hat{\cR}^{LDP}(f) \le \cR(f^*) + K\sqrt{\frac{VC(\cF) + \ln{( \frac{\delta}{2})}}{2n}} \frac{2C_1 M|\cD|}{\bbP(S = k^*)}.
% \]

\begin{proof}

From Lemma \ref{lemma:Risk-LDP}, we have:
\[
\cR(f) = \sum_{k=1}^{|\cD|} \sum_{j=1}^{|\cD|} \left( \boldsymbol{\Pi}_{kj}^{-1} \Ex_{Y,\tilde{X} \mid S = j} \left[ L(f_k(\tilde{X},Y)) \right] \cdot \sum_{l=1}^{|\cD|} \boldsymbol{T}_{kl}^{-1} \bbP(S = l) \right).
\]
Since 
\[
\sum_{l=1}^{|\cD|} \boldsymbol{T}_{kl}^{-1} \bbP(S = l) = \bbP(D = k),
\]
we can then equivalently write:
\[
\boldsymbol{\Pi}_{kj}^{-1} \bbP(D = k) \Ex_{Y, \tilde{X} \mid S = j} \left[ L(f_k(\tilde{X}), Y) \right].
\]
But, note that, for any $j,k$, we have:
\begin{align*}
    \boldsymbol{\Pi}_{kj} &= \bbP(D = k \mid S = j) \\
    &=\frac{\bbP(D = k, S = j)}{\bbP(S = j)} \\
    &= \frac{\boldsymbol{T}_{jk} \bbP(D = k)}{\bbP(S = j)}.
\end{align*}
Therefore, we get
\[
\boldsymbol{\Pi}_{kj}^{-1} = \frac{\boldsymbol{T}_{kj}^{-1}\bbP(S = j)}{\bbP(D = k)}.
\]
Then, we can write:
\[
\sum_{j=1}^{|\cD|} \frac{\boldsymbol{T}_{kl}^{-1} \bbP(S = j)}{\bbP(D = k)} \cdot \frac{\boldsymbol{T}_{jl}\bbP(D = l)}{\bbP(S = j)} = \frac{\bbP(D = l)}{\bbP(D = k)} \sum_{j=1}^{|\cD|} \boldsymbol{T}_{kj}^{-1} \boldsymbol{T}_{jl} = \ones \{k = l\}.
\]
This gives us:
\begin{align*}
    \cR(f) &= \sum_{k=1}^{|\cD|} \sum_{j=1}^{|\cD|} \boldsymbol{T}_{kj}^{-1} \bbP(S = j) \Ex_{Y,\tilde{X} \mid S = j} \left[ L(f_k(\tilde{X}), Y) \right] \\
    &= \sum_{k=1}^{|\cD|} \Ex \left[ \boldsymbol{T}_{kS}^{-1} L(f_k(\tilde{X},Y)) \right] \\
    &= \Ex \left[ \sum_{k=1}^{|\cD|} \boldsymbol{T}_{kS}^{-1} L(f_k(\tilde{X},Y)) \right].
\end{align*}
Then, we can write the empirical counterpart as:
\[
\hat{\cR}^{LDP}(f) = \frac{1}{n} \sum_{i=1}^n \sum_{k=1}^{|\cD|} \boldsymbol{T}_{kS_i}^{-1} L(f_k(\tilde{x}_i),y_i).
\]
Now, define, for each $f = (f_1 \ldots, f_{|\cD|})$
\[
h_f(\tilde{x},y,s) = \sum_{k=1}^{|\cD|} \boldsymbol{T}_{ks}^{-1} L(f_k(\tilde{x}),y).
\]
Then,
\[
\cR(f) = \Ex \left[ h_f(\tilde{x},y,s) \right],
\]
and its empirical counterpart is:
\[
\hat{\cR}^{LDP}(f) = \frac{1}{n}\sum_{i=1}^n h_f(\tilde{x}_i, y_i, s_i).
\]
Next, we bound the range of $h_f$. For fixed $s$,
\[
h_f(\tilde{x},y,s) = C_1 L(f_s(\tilde{x}),y) + C_2 \sum_{k \ne s} L(f_k(\tilde{x}),y).
\]
Notice that since $L$ is bounded above by $M$, $C_1 > 0$, and $C_2 \le 0$, we get
\[
M(|\cD|-1)C_2 \le h_f(\tilde{x},y,s) \le MC_1.
\]
Therefore, we know that
\[
|h_f(\tilde{x},y,s)| \le M(C_1 + (|\cD| - 1)|C_2|).
\]
Now, define the uniform deviation
\[
\Delta = \sup_{f \in \cF^{|\cD|}} \left| \cR(f) - \hat{\cR}^{LDP}(f) \right|.
\]
By the standard symmetrization inequality:
\[
\Ex[\Delta] \le 2 \Re_n(\cF_{LDP}).
\]
Changing one observation changes $\Delta$ by at most $\frac{M(C_1 + (|\cD| - 1))|C_2|}{n}$.

Hence, using McDiarmid's inequality, we have w.p. at least $1 - \delta$:
\[
\Delta \le 2 \Re_n(\cF_{LDP}) + M(C_1 + (|\cD| - 1) |C_2|) \sqrt{\frac{\log(1/\delta)}{2n}}.
\]
Let
\[
\alpha_n(\delta) = 2 \Re_n(\cF_{LDP}) + M(C_1 + (|\cD| - 1) |C_2|) \sqrt{\frac{\log(1/\delta)}{2n}},
\]
then on the event $\Delta \in \alpha_n(\delta)$, the empirical-risk bound follows the ERM property:
\[
\hat{\cR}^{LDP}(\hat{f}) \le \hat{\cR}^{LDP}(f^*) \le \cR(f^*) + \alpha_n(\delta).
\]
Therefore, we have:
\[
\hat{\cR}^{LDP}(\hat{f}) \le \hat{\cR}^{LDP}(\hat{f}) + \alpha_n(\delta) \le \hat{\cR}^{LDP}(f^*) + \alpha_n(\delta) \le \cR(f^*) + 2 \alpha_n(\delta).
\]
Therefore, we conclude that:
\[
\cR(\hat{f}) \le \cR(f^*) + 4 \Re_n(\cF_{LDP}) + 2M(C_1 + (|\cD| - 1)|C_2|) \frac{\log(1/\delta)}{2n}.
\]
This completes the proof.

\end{proof}

%####################################################################################################
%####################################################################################################

\newpage
\subsubsection{\textbf{Lemma} \ref{lemma:NE}}
\label{proof:lemmaNE}

% \textbf{Lemma \ref{lemma:NE}} Consider $\epsilon$-LDP setting with $\pi \in (\frac{1}{|\cD|}, 1]$ and $\bar{\pi} \in [0, \frac{1}{|\cD|})$. For some transformation of $X$, denoted by $X^* = \tilde{T}(X)$, assume there exists at least one anchor point $X_{\text{anchor}}^*$ in the dataset s.t. $\bbP(D = j^*|X_{\text{anchor}}^*) = 1$ for some $j^* \in [|\cD|]$. Then $\pi = \bbP(S = j^* |X_{\text{anchor}}^*)$. Empirically, for a dataset with $n$ observation, let $\boldsymbol{\eta}_{j^*}^n(X^*) = \left( \hat{\bbP}(S = j^*|X_1^*), \ldots, \hat{\bbP}(S = j^*|X_n^*)\right)$, then $\hat{\pi} = \left\| \boldsymbol{\eta}_{j^*}^n(X^*) \right\|_{\infty}$.

\begin{proof}

Notice that $\pi \in \left(\frac{1}{|\cD|}, 1\right]$, $\bar{\pi} \in \left[0, \frac{1}{|\cD|}\right)$, and consequently, we have $\pi > \bar{\pi}$. Hence, by Theorem 5 of \cite{pmlr-v139-zhang21k}, we are well-positioned to apply the noise rate estimation method (Theorem 3) from \cite{patrini2017making} to estimate $\pi$ and $\bar{\pi}$. Our $\epsilon$-LDP setting can be viewed as a special case of class-conditional noise (CCN), where the flip probability is the same across all groups in $\cD$.

Consider
\[
\begin{aligned}
\bbP(S = j^* | x^*) &= \sum_{k=1}^{|\cD|} \bbP(S = j^*|D = k) \cdot \bbP(D = k | x^*) \\
&\overset{(a)}{=} \sum_{k=1}^{|\cD|} \bbP(S = j^* | D = k) \cdot \ones\{j^* = k\} \\ 
&= \pi,
\end{aligned}
\]
where (a) follows from the definition of the anchor point:
\[
\bbP(D = j^* | x^*) = 1 \implies \bbP(D = k | x^*) = 0, \forall k \ne j^*, k, j^* \in [|\cD|].
\]

It is clear that $\bbP(S = j^* \mid x_i)$ attains its maximum when $\bbP(D = j^* \mid x_i) = 1$]. Since we know
\[
\begin{cases}
    \bbP(S = j^* | D = k) = \pi, \text{\ if $j^* = k$} \\
    \bbP(S = j^* | D = k) = \bar{\pi}, \text{\ if $j^* \ne k$}, \\
\end{cases}
\]
so $\bbP(S = j^* \mid x_i)$ is a weighted sum of $\pi$ and $\bar{\pi}$, with weights given by $\{\bbP(D = k \mid x_i)\}_{k=1}^{|\cD|}$. Since $\pi > \bar{\pi}$, the maximum is achieved at the anchor point. Hence, for empirical estimation, $\boldsymbol{\eta}_n(j^*) = \left( \hat{\bbP}(S = j^*|x_1), \ldots, \hat{\bbP}(S = j^*|x_n)) \right)$, then $\hat{\pi} = \| \boldsymbol{\eta}_n(j^*) \|_{\infty}$.

This completes the proof.
\end{proof}

%############################################################################################################################
%############################################################################################################################

\newpage
\subsubsection{\textbf{Theorem} \ref{theorem:GEB-NE}}
\label{proof:theoremGEB-NE}

% \textbf{Theorem \ref{theorem:GEB-NE}}
% For any $\delta \in (0,\frac{1}{3})$, $C_1 = \frac{\pi + |\cD| - 2}{|\cD|\pi - 1}$, denote $VC(\cF)$ as the VC-dimension of the hypothesis class $\cF$, and $K$ be some constant that depends on $VC(\cF)$. If Assumption A, B, and Lemma \ref{lemma:NE} hold, let $f = \{f_k\}_{k=1}^{|\cD|}$ where $f_k \in \cF$ and let $L: Y \times Y \to \bbR_+$ be a loss function bounded by some constant $M$. Denote $k^* \leftarrow \underset{k}{\arg\max} |\hat{\cR}^{LDP}(f_k) \hat{\bbP}(D = k) - \cR^{LDP}(f_k) \bbP(D = k)|$, if $n \ge \frac{8 \ln{(\frac{|\cD|}{\delta})}}{\min_k \bbP(S = k)}$, $n_1\ge \frac{1}{c(\Tilde{\epsilon}-\theta)^2}(M_g+\frac{C_1+\theta}{\ln 2})^2 \ln(\frac{2}{\delta})$, and $M_g+\frac{C_1+\theta}{\ln 2}>\Tilde{\epsilon}>\theta$,  where $c$ is an absolute constant, then with probability $1 - 3\delta$:
% \[
% \hat{\cR}^{LDP}(f) \le \cR(f^*) + K\sqrt{\frac{VC(\cF) + \ln{( \frac{\delta}{2})}}{2n}} \frac{2(C_1+\Tilde{\epsilon}) M|\cD|}{\bbP(S = k^*)}.  
% \]

\begin{proof}

First, we establish a concentration inequality with regard to $\hat{C}_1$ and $C_1$.

Since for any constant $L$, we have
\[
\begin{aligned}
    \|L\|_{\psi_1} &=\inf \{t>0 | \Ex [e^{|L|/t}] \le 2\} \\
    &= \inf \{t>0 | e^{|L|/t} \le 2\} \\
    &= \frac{|L|}{\ln 2},
\end{aligned}
\]
where $\|\cdot \|_{\psi_1}$ is a norm. It follows that the standardized statistic $\Tilde{C}_{1,k} = \hat{C}_{1,k} - \Ex [\hat{C}_{1,k}]$ is also sub-exponential:
\[
\begin{aligned}
    \| \Tilde{C}_{1,k} \|_{\psi_1}&\leq  \| \hat{C}_{1,k} \|_{\psi_1} + \| \Ex [\hat{C}_{1,k}] \|_{\psi_1} \\
    &\leq M_g + \frac{|\Ex [\hat{C}_{1,k}|]}{\ln 2}\\
    &\overset{(a)}{=} M_g+ \frac{C_1+\theta}{\ln 2},
\end{aligned}
\]
where (a) follows from Assumption B (Nearly Unbiasedness).

Define 
\[
K_C:= M_g + \frac{C_1 + \theta}{\ln 2}.
\]
Since the $n_1$ groups in Procedure \ref{procedure:C_1} are independent. Thus, the centered variables $\tilde{C}_{1,1}, \ldots, \tilde{C}_{1,n_1}$ are independent zeor-mean sub-exponential random variables. By Bernstein's inequality, we get:
\[
\bbP \left( \left| \frac{1}{n_1} \sum_{k=1}^{n_1} \tilde{C}_{1,k} \right| > t \right) \le 2 \exp \left\{ -cn_1 \min \left( \frac{t^2}{K_C^2}, \frac{t}{K_C} \right) \right\}.
\]
Now, write
\[
\hat{C}_1 - C_1 = \frac{1}{n_1} \sum_{k=1}^{n_1} \tilde{C}_{1,k} + \left( \frac{1}{n_1} \sum_{k=1}^{n_1} \Ex \left[ \hat{C}_{1,k} \right] - C_1 \right).
\]
Then, by Assumption B, we have:
\[
\left| \frac{1}{n_1} \sum_{k=1}^{n_1} \Ex \left[ \hat{C}_{1,k} \right] - C_1 \right| \le \theta.
\]
Then, if we have
\[
\left| \hat{C}_1 - C_1 \right| > \tilde{\epsilon}.
\]
Then, we get:
\[
\left| \frac{1}{n_1} \sum_{k=1}^{n_1} \hat{C}_{1,k} \right| > \tilde{\epsilon} - \theta.
\]
Note that $\tilde{\epsilon} - \theta > 0$, since $\tilde{\epsilon} > \theta$ and also $K_C > \tilde{\epsilon} > \theta$ implies $\tilde{\epsilon} - \theta < K_C$. Therefore, we get:
\[
\bbP \left( \left| \hat{C}_1 - C_1 \right| > \tilde{\epsilon} \right) \le 2 \exp \left\{ -cn_1\frac{(\tilde{\epsilon} - \theta)^2}{K_C^2} \right\}.
\]
With the condition
\[
n_1 \ge \frac{1}{c(\tilde{\epsilon} - \theta)^2}\left( M_g + \frac{C_1 + \theta}{\ln 2} \right)^2 \ln (\frac{2}{\delta}),
\]
we get:
\[
\bbP \left( \left| \hat{C}_1 - C_1 \right| > \tilde{\epsilon} \right) \le \delta.
\]
Note that clipping is non-expansive around $C_1 \ge 1$. Therefore, the same bound is also true for the clipped estimator. Hence, w.p. at least $1-\delta$,
\[
\left| \hat{C}_1 - C_1 \right| \le \tilde{\epsilon},
\]
and denote this event as $\cE_C:= \left\{ \left| \hat{C}_1 - C_1 \right| \le \tilde{\epsilon} \right\}$.

Now, note that for any value $c \ge 1$, define $\boldsymbol{T}^{-1}(c)$ by 
\[
\boldsymbol{T}_{ks}^{-1}(c) = 
\begin{cases}
    c \quad \quad \quad, k = s, \\
    \frac{1-c}{|\cD| - 1} \ \ , k \ne s,
\end{cases}
\]
for each $f = (f_1, \ldots, f_{|\cD|})$, define
\[
h_{f,c}(\tilde{x},y,s):= \sum_{k=1}^{|\cD|} \boldsymbol{T}_{ks}^{-1}(c) L(f_k(\tilde{x}),y).
\]
Notice that when $c = C_1$, this is precisely the true inverse-weighted loss:
\[
h_{f,C_1}(\tilde{x},y,s) = \sum_{k=1}^{|\cD|} \boldsymbol{T}_{ks}^{-1} L(f_k(\tilde{x}),y).
\]
Recall that in the proof of Theorem \ref{theorem:GEB}, we have:
\begin{align*}
    \cR(f) &= \Ex \left[ h_{f,C_1}(\tilde{x},y,s) \right] \\
    &= \Ex \left[ \sum_{k=1}^{|\cD|} \boldsymbol{T}_{kS}^{-1} L(f_k(\tilde{x}),Y) \right].
\end{align*}
Thus, we can write:
\[
\hat{\cR}_{\hat{\boldsymbol{T}}^{-1}}^{LDP} = \frac{1}{n} \sum_{i=1}^n h_{f,\hat{C}_1}(\tilde{x}_i,y_i,s_i).
\]
On the event $\cE_C$, we have:
\[
\hat{C}_1 \in [1, C_1 + \tilde{\epsilon}].
\]
Hence, for every $f \in \cF^{|\cD|}$, we have:
\[
h_{f,\hat{C}_1} \in \cF_{\tilde{\epsilon}}.
\]

Next, we bound the range of $\cF_{\tilde{\epsilon}}$. Fix $c \in [1, C_1 + \tilde{\epsilon}]$. For fixed $s$, we have:
\[
h_{f,c}(\tilde{x},y,s) = c L(f_s(\tilde{x}),y) + \frac{1-c}{|\cD|-1} \sum_{k \ne s} L(f_k(\tilde{x}),y),
\]
since we know $L$ is upper-bounded by $M$ and $c \ge 1$.

Notice that $\frac{1-c}{|\cD|-1} \le 0$, therefore:
\[
(1-c)M \le h_{f,c}(\tilde{x},y,s) \le cM.
\]
Therefore,
\[
\left| h_{f,c}(\tilde{x},y,s) \right| \le M(2(C_1 + \tilde{\epsilon})-1).
\]
Now, define
\[
\Delta_{\tilde{\epsilon}}:= \sup_{h \in \cF_{\tilde{\epsilon}}} \left| \Ex \left[ h(\tilde{x},y,s) \right] - \frac{1}{n}\sum_{i=1}^n h(\tilde{x}_i,y_i,s_i) \right|,
\]
by symmetrization, we have:
\[
\Ex \left[\Delta_{\tilde{\epsilon}} \right] \le 2 \Re_n(\cF_{\tilde{\epsilon}}).
\]
Note that changing one observation $(\tilde{x}_x,y_i,s_i)$ changes $\Delta_{\tilde{\epsilon}}$ by at most $\frac{M(2(C_1 + \tilde{\epsilon}) - 1)}{n}$,
by McDiarmid's inequality, w.p. at least $1 - \delta$, 
\[
\Delta_{\tilde{\epsilon}} \le 2 \Re_n(\cF_{\tilde{\epsilon}}) + M(2(C_1 + \tilde{\epsilon})-1) \sqrt{\frac{\ln(1/\delta)}{2n}}.
\]
Define
\[
\alpha_n(\delta):= 2 \Re_n(\cF_{\tilde{\epsilon}}) + M(2(C_1 + \tilde{\epsilon})-1) \sqrt{\frac{\ln(1/\delta)}{2n}}.
\]
Let
\[
\cE_R:= \left\{ \Delta_{\tilde{\epsilon}} \le \alpha_n(\delta) \right\},
\]
then, we get
\[
\bbP(\cE_R) \ge 1 - \delta.
\]
Recall that on $\cE_C$, we have:
\[
\left| \hat{C}_1 - C_1 \right| \le \tilde{\epsilon}.
\]
Since 
\[
C_2 = \frac{1 - C_1}{|\cD| - 1},
\]
then, we get
\[
\left| \hat{C}_2 - C_2 \right| = \frac{\left| \hat{C}_1 - C_1 \right|}{|\cD| - 1}.
\]
Then, for any fixed $s$, we have:
\begin{align*}
    \sum_{k=1}^{|\cD|} \left| \hat{\boldsymbol{T}}_{ks}^{-1} - \boldsymbol{T}_{ks}^{-1} \right| &= \left| \hat{C}_1 - C_1 \right| + (|\cD| - 1) \frac{\left| \hat{C}_1 - C_1 \right|}{|\cD| - 1} \\
    &= 2 \left| \hat{C}_1 - C_1 \right|.
\end{align*}
Thus, on $\cE_C$, we have:
\[
\max_s \sum_{k=1}^{|\cD|} \left| \hat{\boldsymbol{T}}_{ks}^{-1} - \boldsymbol{T}_{ks}^{-1} \right| \le 2 \tilde{\epsilon}.
\]
Therefore, for every $f$, we have:
\[
\left| h_{f, \hat{C}_1}(\tilde{x},y,s) - h_{f,C_1}(\tilde{x},y,s) \right| \le 2M\tilde{\epsilon}.
\]
Hence, uniformly over $f$, we get:
\[
\left| \Ex \left[ h_{f,\hat{C}_1}(\tilde{X},Y,S) \right] - \cR(f) \right| \le 2 M \tilde{\epsilon},
\]
and
\[
\left| \hat{\cR}_{\hat{\boldsymbol{T}}^{-1}}^{LDP}(f) - \hat{\cR}_{\boldsymbol{T}^{-1}}^{LDP}(f) \right| \le 2 M \tilde{\epsilon}.
\]
Now, on $\cE \cap \cE_R$, we know this event is w.p. at least $1 - 2\delta$. Since by definition, $\hat{f}$ minimizes the empirical risk with $\hat{\boldsymbol{T}}^{-1}$ involved, we have:
\[
\hat{\cR}_{\hat{\boldsymbol{T}}^{-1}}^{LDP}(\hat{f}) \le \hat{\cR}_{\hat{\boldsymbol{T}}^{-1}}^{LDP}(f^*).
\]
Since $h_{f^*,\hat{C}_1} \in \cF_{\tilde{\epsilon}}$ on $\cE_C$, then $\cE_R$ gives us:
\[
\hat{\cR}_{\hat{\boldsymbol{T}}^{-1}}^{LDP}(f^*) \le \Ex \left[ h_{f^*, \hat{C}_1}(\tilde{X},Y,S) \right] + \alpha_n(\delta).
\]
Thus, we have:
\[
\Ex \left[ h_{f^*, \hat{C}_1} (\tilde{X}, Y, S) \right] \le \cR(f^*) + 2 M \tilde{\epsilon}.
\]
Combining the bounds, we get:
\begin{align*}
    \hat{\cR}_{\hat{\boldsymbol{T}}^{-1}}^{LDP}(\hat{f}) & \le \cR^*(f) + \alpha_n(\delta) + 2 M \tilde{\epsilon} \\
    &= \cR(f^*) + 2 \Re_n(\cF_{\tilde{\epsilon}}) + M(C_1 + (|\cD|-1)|C_2|+2\tilde{\epsilon}) \sqrt{\frac{\log(1/\delta)}{2n}} + 2M \tilde{\epsilon}.
\end{align*}
Then, notice that on $\cE_C \cap \cE_R$,
\[
\cR(\hat{f}) \le \Ex \left[ h_{\hat{f},\hat{C}_1} (\tilde{X}, Y, S) \right] + 2 M \tilde{\epsilon}.
\]
By the uniform convergence on $\cE_R$,
\[
\Ex \left[ h_{\hat{f},\hat{C}_1} (\tilde{X}, Y, S) \right] \le \hat{\cR}_{\hat{\boldsymbol{T}}^{-1}}^{LDP} (\hat{f}) + \alpha_n(\delta).
\]
This gives:
\[
\cR(\hat{f}) \le \hat{\cR}_{\hat{\boldsymbol{T}}^{-1}}^{LDP} (\hat{f}) + \alpha_n(\delta) + 2M \tilde{\epsilon},
\]
by ERM, we know,
\[
\hat{\cR}_{\hat{\boldsymbol{T}}^{-1}}^{LDP} (\hat{f}) \le \hat{\cR}_{\hat{\boldsymbol{T}}^{-1}}^{LDP} (f^*).
\]
Thus, we have 
\[
\cR(\hat{f}) \le \hat{\cR}_{\hat{\boldsymbol{T}}^{-1}}^{LDP} (f^*) + \alpha_n(\delta) + 2 M \tilde{\epsilon},
\]
by uniform convergence, we get:
\[
\hat{\cR}_{\hat{\boldsymbol{T}}^{-1}}^{LDP} (f^*) \le \Ex \left[ h_{f^*,\hat{C}_1} (\tilde{X}, Y, S) \right] + \alpha_n(\delta),
\]
and also
\[
\Ex \left[ h_{f^*,\hat{C}_1} (\tilde{X}, Y, S) \right] \le \cR(f^*) + 2 M \tilde{\epsilon}.
\]
Combining results, we get:
\begin{align*}
    \cR(\hat{f}) &\le \cR(f^*) + 2 \alpha_n(\delta) + 4 M \tilde{\epsilon} \\
    &= \cR(f^*) + 4 \Re_n(\cF_{\tilde{\epsilon}}) + 2M(C_1 + (|\cD|-1)|C_2|+2\tilde{\epsilon}) \sqrt{\frac{\log(1/\delta)}{2n}} + 4M\tilde{\epsilon}.
\end{align*}
This completes the proof.

\end{proof}

\newpage
\subsection{Data Analysis - Auto Insurance}
\label{app:deferred_exp}

\subsubsection{Data Description}

The Auto Insurance dataset contains 8,150 observations, 17 features, and a single response variable \citep{Roy2021}. The response represents risk classification (high risk, low risk), which defines the task as a classification problem. In our experiment, we choose $D = \text{sex}$ to be the sensitive attribute, taking values``Male'' and ``Female''.The privatized sensitive attribute $S$ is generated under varying privacy levels, specified by different values of $\epsilon$'s, following Definition \ref{def:pm}. %$D$ was used to set the performance benchmark and is masked under any other settings.

\subsubsection{Experiments Setup}

We adopt the same experimental design as in the regression task for the Health Insurance dataset (Section~\ref{sec:6}), considering two scenarios: one where the noise rates are known and another where they are unknown. Across both scenarios, we use the following configuration: 1) Hypothesis class $\cF$: linear models; 2) Representation $T(X) = \tilde{X}$, learned via a supervised feedforward neural network; 3) Three predefined noise levels, corresponding to $\pi = 0.9$, $0.8$, and $0.7$.

The data is split into training and test sets. For each setting, we estimate the discrimination-free premium (Definition~\ref{def:dfp}) using Algorithm~\ref{alg:S-NE} (MPTP-LDP), and benchmark it against the best-estimated premium (Definition~\ref{def:bep}) computed using Algorithm~\ref{alg:D} (MPTP). The only difference from the regression analysis in Section~\ref{sec:6} is the use of cross-entropy loss, appropriate for the classification task.

\subsubsection{Empirical Results}

We compare the convergence behavior of the pricing strategies based on their test loss, with results summarized in this section. As noted above, the test loss here refers to cross-entropy computed on the test data, in contrast to the mean squared error used in the regression task.

In general, we observe similar patterns to those in the regression analysis presented in Section~\ref{sec:6}, further supporting our theoretical predictions. These results provide additional evidence for the robustness of our approach across both regression and classification tasks. Since no new insights emerge from these experiments, we do not describe the figures in detail. Instead, we provide a mapping below to help readers draw direct analogies to the figures and findings reported in Section~\ref{sec:6}.

\begin{itemize}
    \item Figure \ref{fig:auto_1.0_loss_r1-LDP-a}: analogous to Figure \ref{fig:auto_1.0_loss_r1-LDP-a}, comparison of test loss by noise rate for a given sample size, with known $\pi$.
    \begin{figure}[H]
    \centering
    \begin{subfigure}{0.49\textwidth}
        \centering
        \includegraphics[width=1\textwidth]{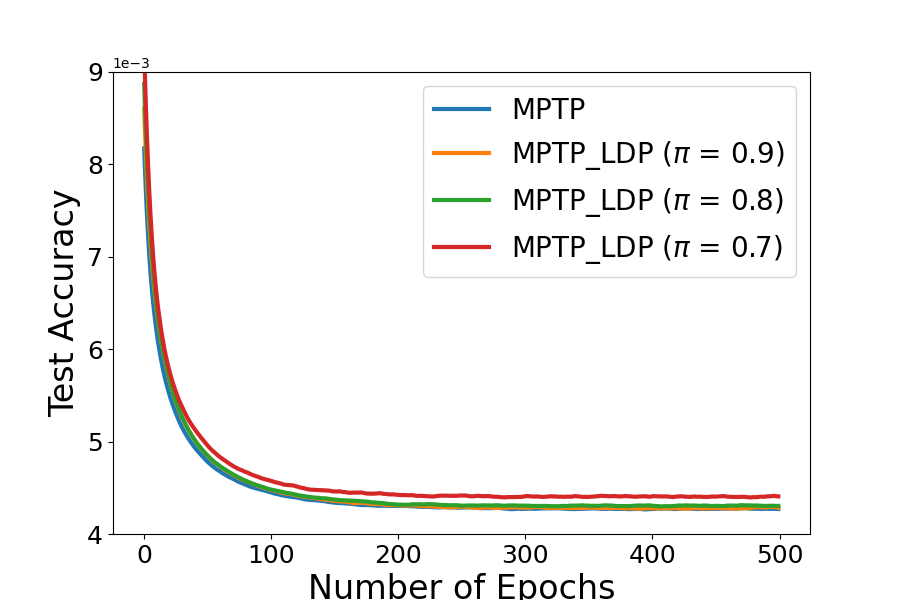}
        \caption{Original attributes $X$}
        \label{fig:auto_1.0_mu_X_A_loss_r1}
    \end{subfigure}
    \begin{subfigure}{0.49\textwidth}
        \centering
        \includegraphics[width=1\textwidth]{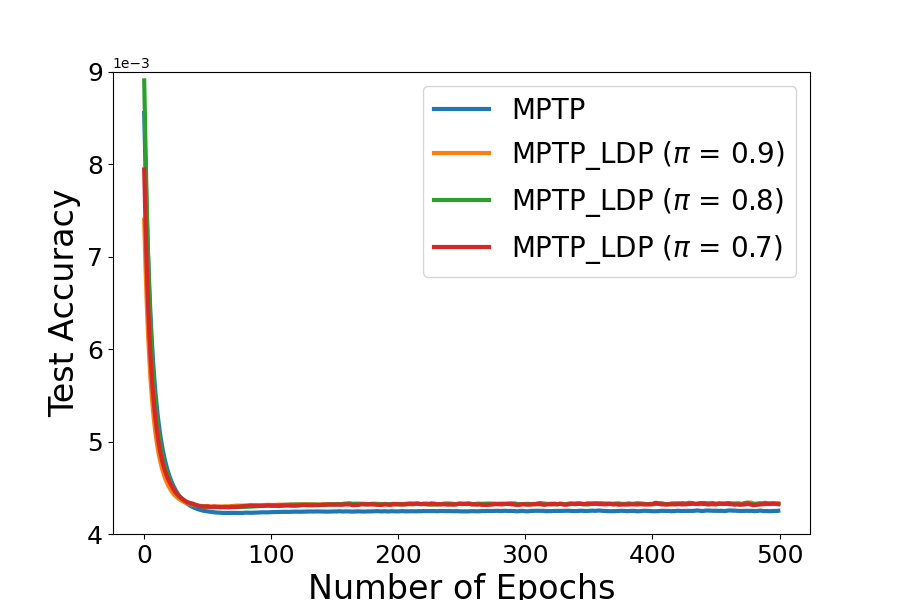} 
        \caption{Transformed attributes $\tilde{X}$}
        \label{fig:auto_1.0_mu_TX_A_loss_r1}
    \end{subfigure}
    \caption{Comparison of test loss by noise rate for a given sample size, with known $\pi$.}
    \label{fig:auto_1.0_loss_r1-LDP-a}
    \end{figure} 
    
    \item Figure \ref{fig:auto_1.0_loss_r1-LDP-b}: analogous to Figure \ref{fig:health_1.0_loss_r1-LDP-b}, comparison of test loss by sample size for a given noise rate, with known $\pi$.
    \begin{figure}[H]
    \begin{subfigure}{0.49\textwidth}
        \centering
        \includegraphics[width=\textwidth]{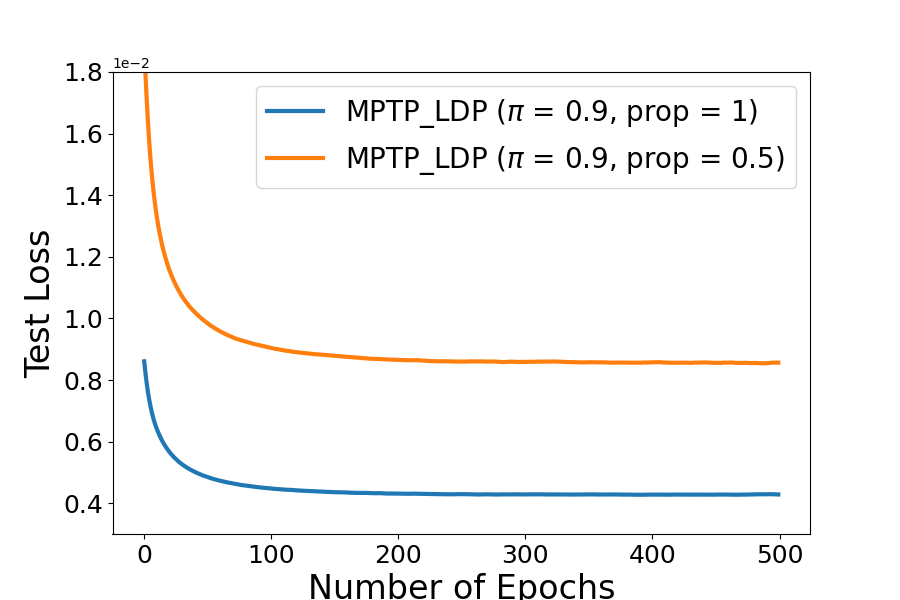}
         \caption{Original attributes $X$}
        \label{fig:auto_p0.9_mu_X_A_loss_r1}
    \end{subfigure}
    \begin{subfigure}{0.49\textwidth}
        \centering
        \includegraphics[width=\textwidth]{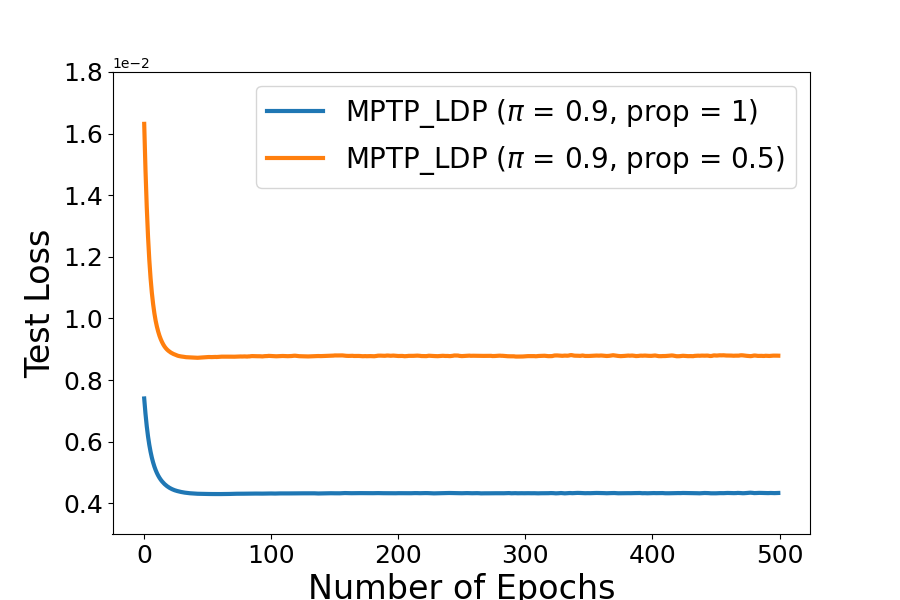} 
        \caption{Transformed attributes $\tilde{X}$}
        \label{fig:auto_p0.9_mu_TX_A_loss_r1}
    \end{subfigure}
    \caption{Comparison of test loss by sample size for a given noise rate, with known $\pi$.}
    \label{fig:auto_1.0_loss_r1-LDP-b}
    \end{figure} 
    
    \item Figure \ref{fig:auto_mu_TX_X_A_loss_ne_err_r1}: analogous to Figure \ref{fig:health_mu_TX_X_A_loss_ne_err_r1}, effect of estimation error in noise rate ($\pm 10\%$) on test loss.
    \begin{figure}[H]
    \centering
    \begin{subfigure}{1.2\textwidth}
        \includegraphics[width=0.3\textwidth]{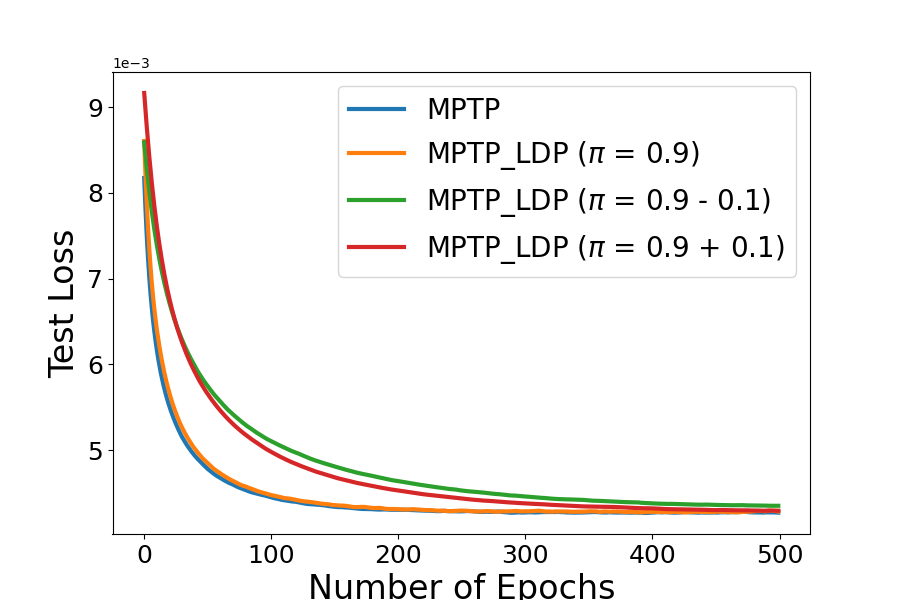}
        \includegraphics[width=0.3\textwidth]{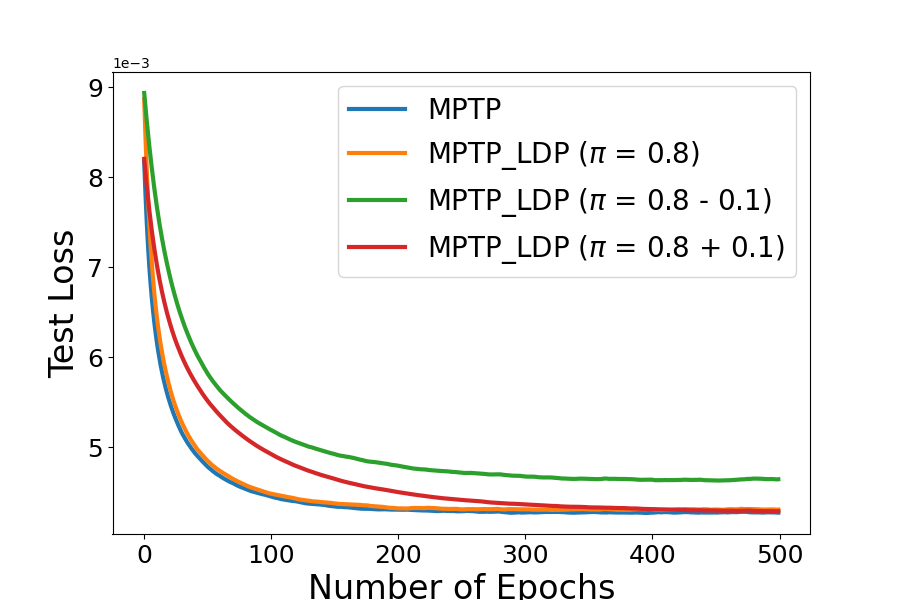} 
        \includegraphics[width=0.3\textwidth]
        {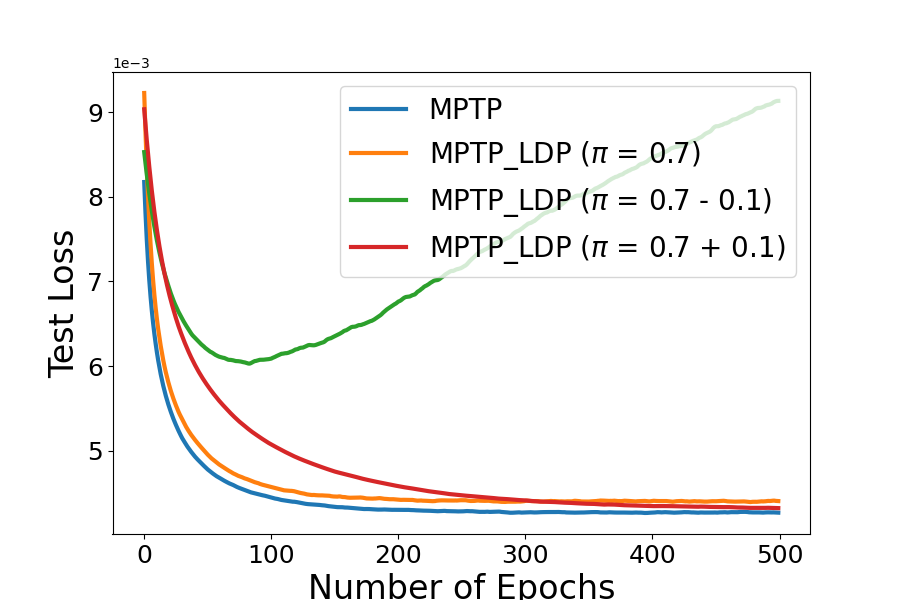} 
        \caption{Original attributes: $\pi=0.9, 0.8, 0.7$ from left to right}
    \end{subfigure}
    \begin{subfigure}{1.2\textwidth}
        \includegraphics[width=0.3\textwidth]{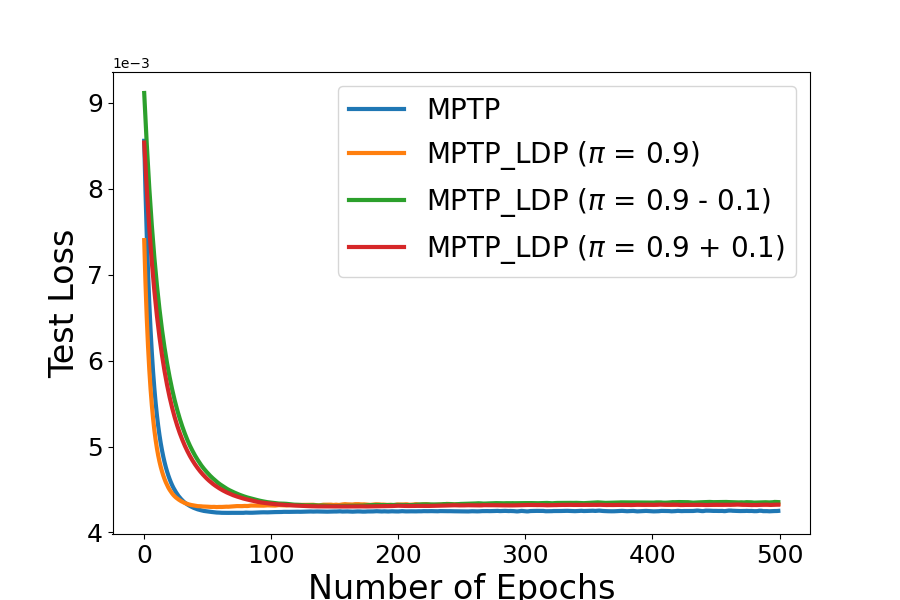}
        \includegraphics[width=0.3\textwidth]{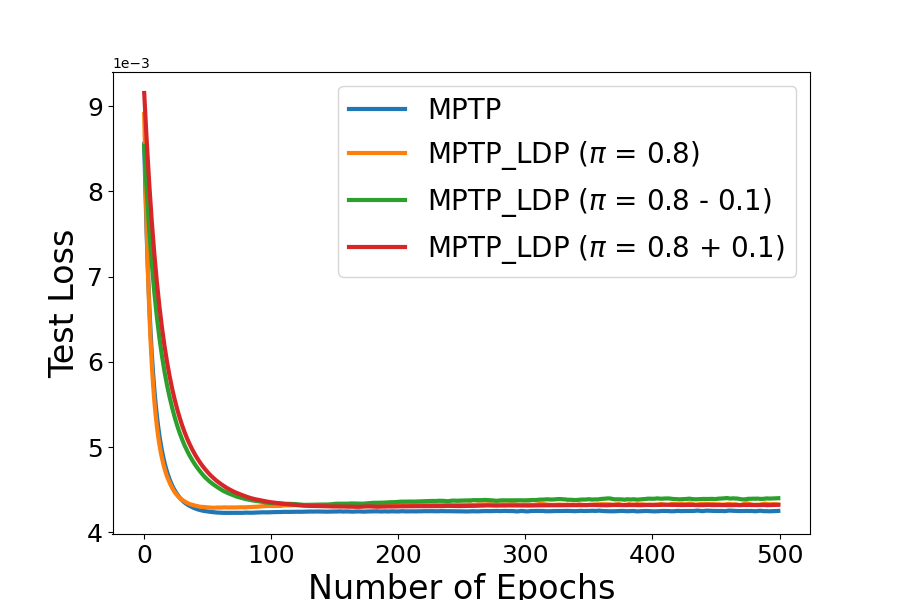} 
        \includegraphics[width=0.3\textwidth]
        {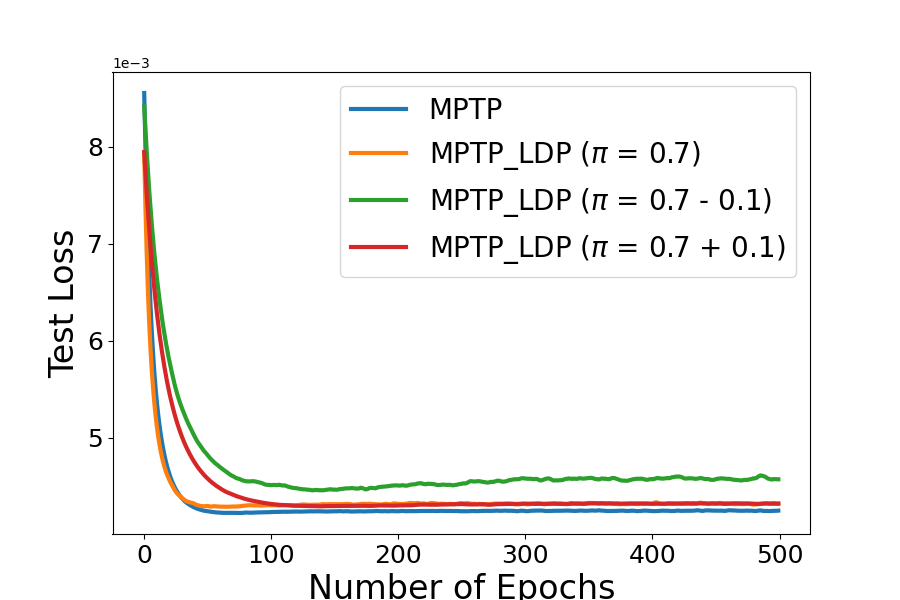} 
        \caption{Transformed attributes: $\pi=0.9, 0.8, 0.7$ from left to right}
    \end{subfigure}
    \caption{Effect of estimation error in noise rate ($\pm 10\%$) on test loss.}
    \label{fig:auto_mu_TX_X_A_loss_ne_err_r1}
    \end{figure} 
    
    \item Figure \ref{fig:auto_1.0_mu_X_TX_A_loss_ne_r1}: analogous to Figure \ref{fig:auto_1.0_mu_X_TX_A_loss_ne_r1}, comparison of test loss by noise rate for a given sample size, with estimated $\pi$.
    \begin{figure}[H]
    \centering
    \begin{subfigure}{0.49\textwidth}
        \centering
        \includegraphics[width=\textwidth]{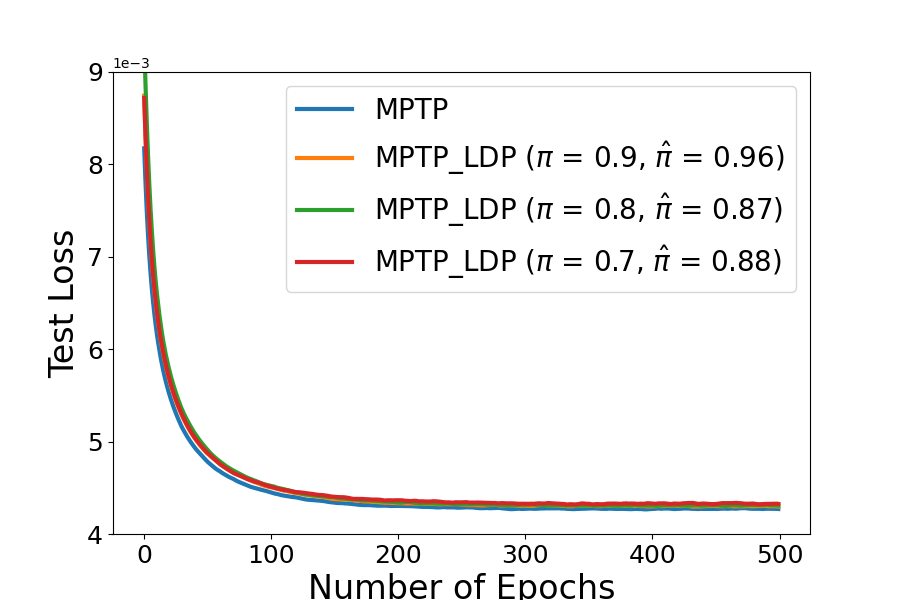} 
        \caption{Original attributes $X$}
        \label{fig:auto_1.0_mu_X_A_loss_ne_n14_r1}
    \end{subfigure}
    \begin{subfigure}{0.49\textwidth}
        \centering
        \includegraphics[width=\textwidth]{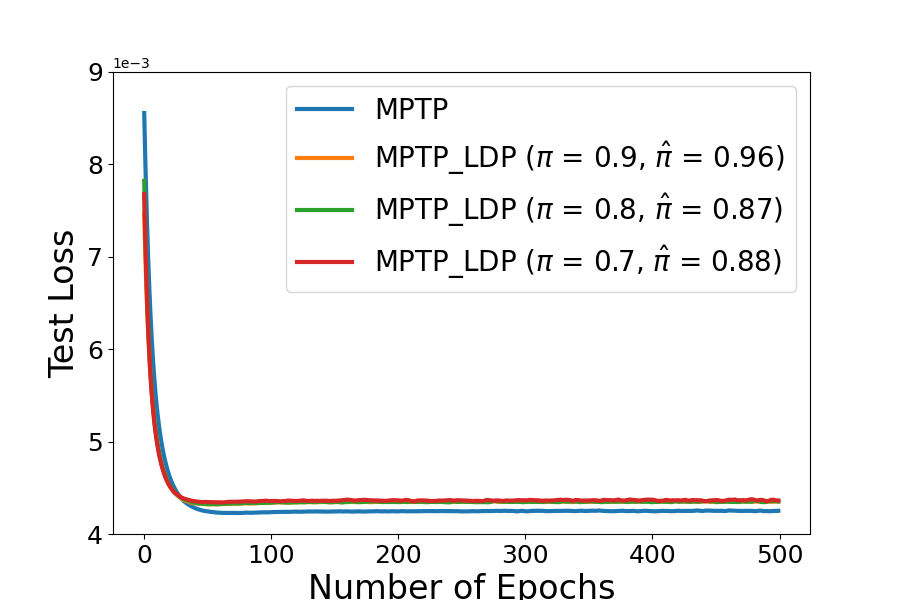} 
        \caption{Transformed attributes $\tilde{X}$}
        \label{fig:auto_1.0_mu_TX_A_loss_ne_n14_r1}
    \end{subfigure}
    \caption{Comparison of test loss by noise rate for a given sample size, with estimated $\pi$.}
    \label{fig:auto_1.0_mu_X_TX_A_loss_ne_r1}
    \end{figure}  
    
    \item Figure \ref{fig:auto_subset_mu_X_TX_A_loss_ne_r1}: analogous to Figure \ref{fig:health_subset_mu_X_TX_A_loss_ne_r1}, comparison of test loss by sample size for a given noise rate, with estimated $\pi$.
    \begin{figure}[H]
    \centering
    \begin{subfigure}{0.49\textwidth}
        \centering
        \includegraphics[width=\textwidth]{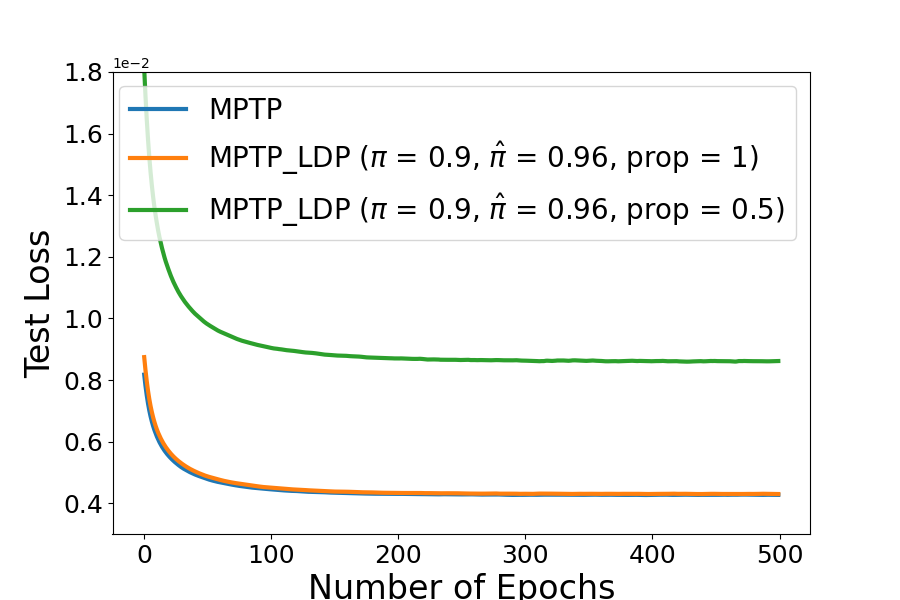} 
        \caption{Original attributes $X$}
        \label{fig:auto_subset_p0.35_mu_X_A_loss_ne_n14_r1}
    \end{subfigure}
    \begin{subfigure}{0.49\textwidth}
        \centering
        \includegraphics[width=\textwidth]{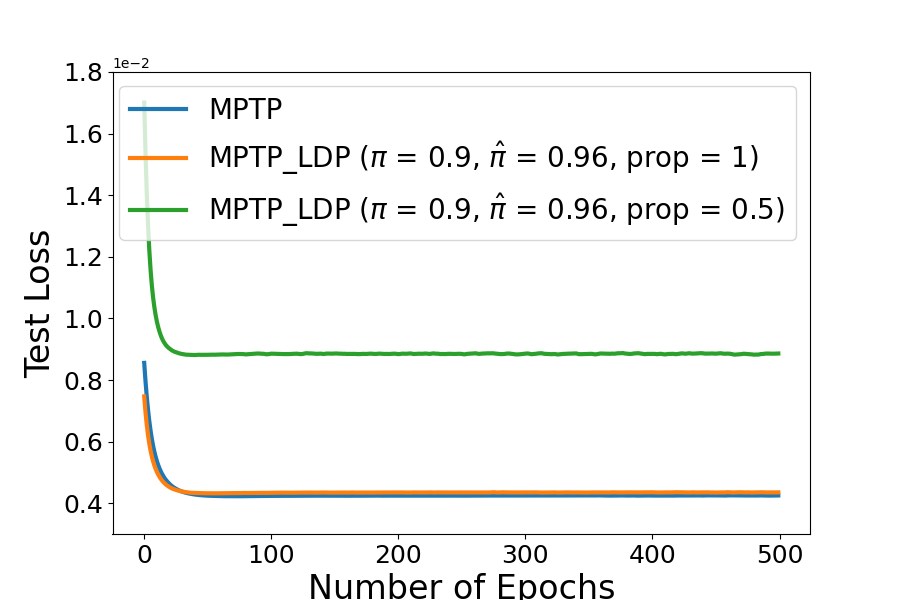} 
        \caption{Transformed attributes $\tilde{X}$}
        \label{fig:auto_subset_p0.35_mu_TX_A_loss_ne_n14_r1}
    \end{subfigure}
    \caption{Comparison of test loss by sample size for a given noise rate, with estimated $\pi$.}
    \label{fig:auto_subset_mu_X_TX_A_loss_ne_r1}
    \end{figure} 
    
    \item Figure \ref{fig:auto_1.0_mu_X_TX_A_loss_ne_n1_r1}: analogous to Figure \ref{fig:health_1.0_mu_X_TX_A_loss_ne_n1_r1}, effect of group size ($n_1= 2 ~\text{or}~4$) on test loss.
    \begin{figure}[H]
        \centering
    \begin{subfigure}{1.2\textwidth}
        \includegraphics[width=0.3\textwidth]{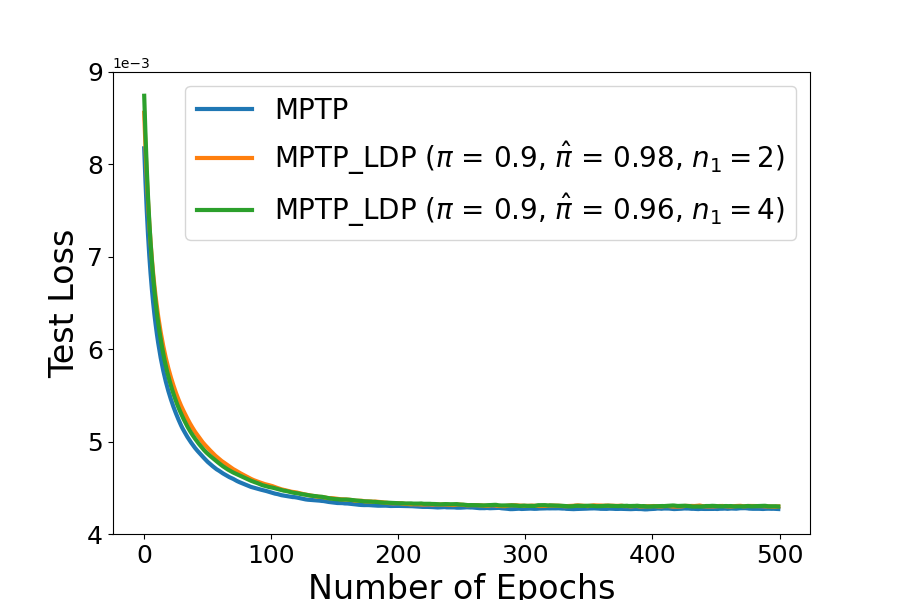}
        \includegraphics[width=0.3\textwidth]{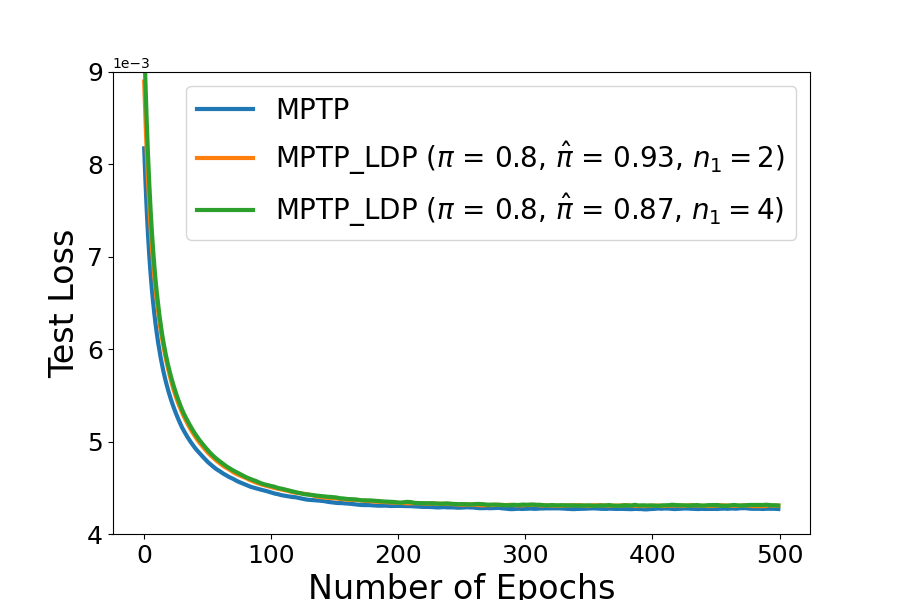} 
        \includegraphics[width=0.3\textwidth]{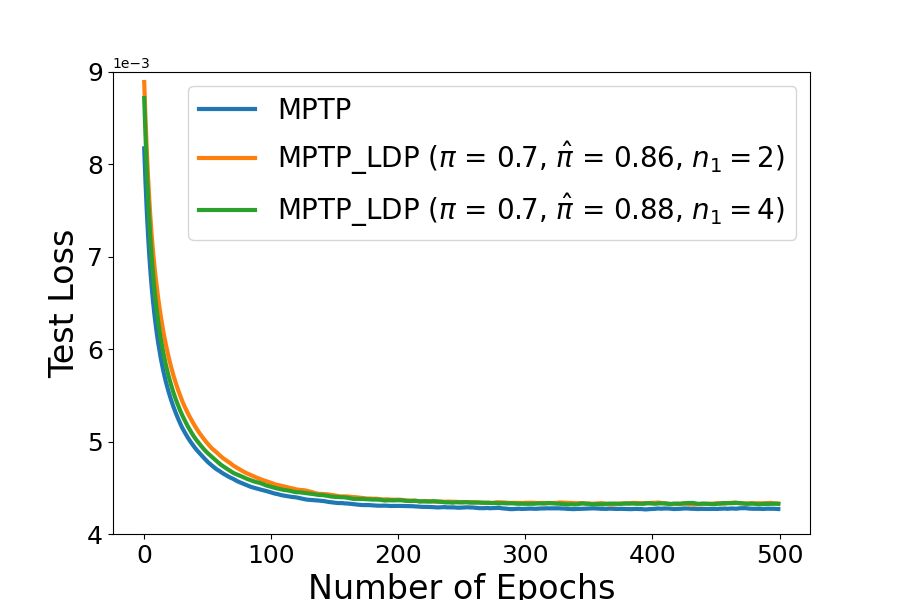} 
        \caption{Original attributes: $\pi=0.9, 0.8, 0.7$ from left to right}
    \end{subfigure}
    \begin{subfigure}{1.2\textwidth}
        \includegraphics[width=0.3\textwidth]{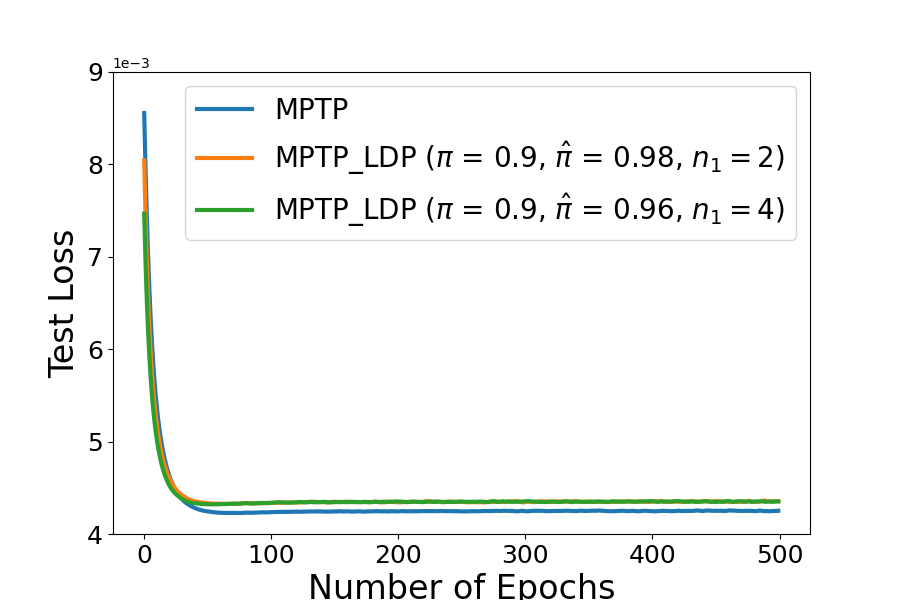}
        \includegraphics[width=0.3\textwidth]{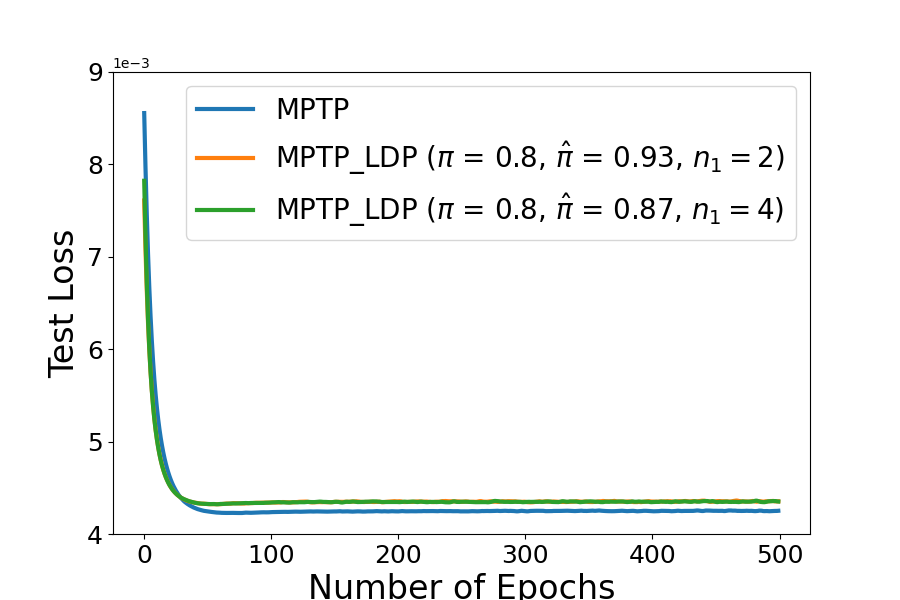} 
        \includegraphics[width=0.3\textwidth]{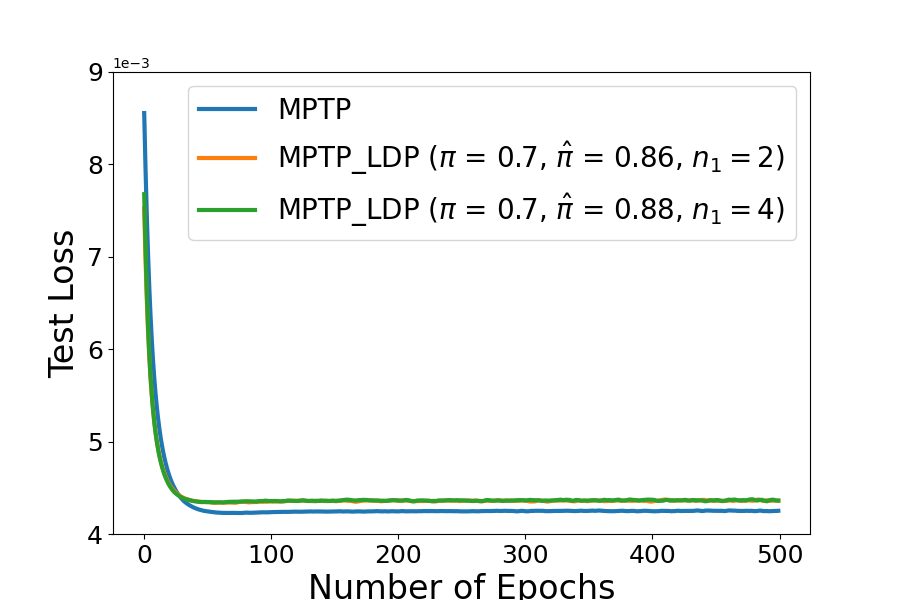} 
        \caption{Transformed attributes: $\pi=0.9, 0.8, 0.7$ from left to right}
    \end{subfigure}
    % \vspace{-8pt}
    \caption{Effect of group size ($n_1= 2 ~\text{or}~4$) on test loss.}
    \label{fig:auto_1.0_mu_X_TX_A_loss_ne_n1_r1}
\end{figure}  
\end{itemize}

%###################################################################################################
%###########################################################################################

\end{document}